\PassOptionsToPackage{numbers, compress}{natbib}
\documentclass[]{antgroup}
\usepackage{antgroup}

\usepackage{graphicx}
\usepackage{xspace}
\usepackage{tikz}
\usepackage{todonotes}
\usepackage{multirow}
\usepackage{amsmath}
\usepackage{cleveref}
\usepackage{subcaption}
\usepackage{multicol}
\usepackage{amssymb}
\usepackage{booktabs}
\usepackage{array}
\usepackage{bm}
\usepackage{enumitem}
\usepackage{CJKutf8}
\usepackage{algorithm}
\usepackage{algpseudocode}
\usepackage{tabularx}
\usepackage{bbm}
\usepackage{makecell}
\usepackage{threeparttable}
\usepackage{float}
\usepackage[normalem]{ulem}
\useunder{\uline}{\ul}{}

\usepackage{amsmath,amsfonts,bm}

\def\figref#1{figure~\ref{#1}}

\def\secref#1{section~\ref{#1}}

\def\eqref#1{equation~\ref{#1}}

\def\algref#1{algorithm~\ref{#1}}

\def\1{\bm{1}}

\DeclareMathAlphabet{\mathsfit}{\encodingdefault}{\sfdefault}{m}{sl}
\SetMathAlphabet{\mathsfit}{bold}{\encodingdefault}{\sfdefault}{bx}{n}

\newcommand*\justify{%
  \fontdimen2\font=0.4em%
  \fontdimen3\font=0.2em%
  \fontdimen4\font=0.1em%
  \fontdimen7\font=0.1em%
  \hyphenchar\font=`\-%
}

\renewcommand{\texttt}[1]{%
  \begingroup
  \ttfamily
  \begingroup\lccode`~=`/\lowercase{\endgroup\def~}{/\discretionary{}{}{}}%
  \begingroup\lccode`~=`[\lowercase{\endgroup\def~}{[\discretionary{}{}{}}%
  \begingroup\lccode`~=`.\lowercase{\endgroup\def~}{.\discretionary{}{}{}}%
  \catcode`/=\active\catcode`[=\active\catcode`.=\active
  \justify\scantokens{#1\noexpand}%
  \endgroup
}

\usepackage{amsmath}
\usepackage{amssymb}
\usepackage{amsfonts}                               %
\usepackage{amsthm}
\usepackage[mathcal]{eucal}
\usepackage{mathrsfs}
\usepackage{bm}                                     %
\usepackage{blkarray}                               %
\usepackage{nicefrac}                               %

\usepackage{wrapfig}
\usepackage{graphicx}                               %
\usepackage{caption}
\usepackage{cleveref}

\usepackage{tikz}                                          
\usepackage{circuitikz}
\usetikzlibrary{patterns}
\usetikzlibrary{positioning,calc,fit,decorations.pathmorphing,shapes.geometric, shapes.gates.logic.US, calc}
\usetikzlibrary{arrows,arrows.meta,decorations.markings,shapes,shapes.arrows}
\usetikzlibrary{decorations,decorations.pathreplacing}
\usetikzlibrary{backgrounds}
\usepackage{pgfplots}
\usepackage{pgfplotstable}
\usepgfplotslibrary{groupplots}
\usepackage{scalefnt}
\pgfplotsset{compat=newest}

\definecolor{firstcolor}{HTML}{C3423F}
\definecolor{secondcolor}{HTML}{2A4B8C}
\definecolor{myblue}{HTML}{D9E6FF}
\definecolor{statusgreen}{HTML}{26734D}
\definecolor{statusorange}{HTML}{CC7000}

\newcommand{\github}{%
  \raisebox{-1.5pt}{\includegraphics[height=1.05em]{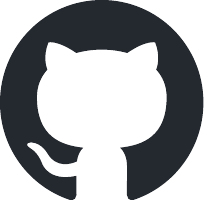}}\xspace}
\newcommand{\hf}{%
  \raisebox{-1.5pt}{\includegraphics[height=1.05em]{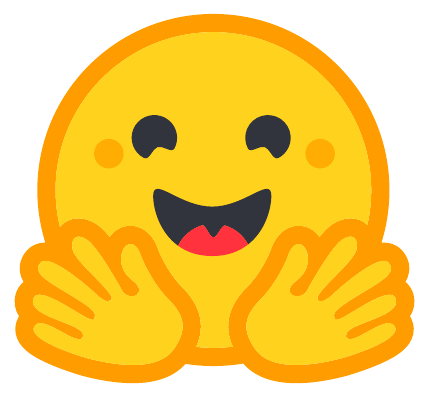}}\xspace}

\providecommand{\lladaimage}{{LLaDA-Image}\xspace}
\providecommand{\lladaimageturbo}{{LLaDA-Image Turbo}\xspace}

\usepackage{tocloft}
\newenvironment{talign*}
  {\csname align*\endcsname}
  {\endalign}

\definecolor{darkred}{RGB}{150,0,0}
\providecommand{\darkredtext}[1]{\textcolor{darkred}{#1}}

\renewcommand{\eqref}[1]{%
  \hyperref[#1]{\darkredtext{\textup{Eq.~(\ref*{#1})}}}}
\renewcommand{\secref}[1]{%
  \hyperref[#1]{\darkredtext{Sec.~\ref*{#1}}}}
\renewcommand{\figref}[1]{%
  \hyperref[#1]{\darkredtext{Fig.~\ref*{#1}}}}
\renewcommand{\algref}[1]{%
  \hyperref[#1]{\darkredtext{Alg.~\ref*{#1}}}}

\providecommand{\thmref}[1]{\hyperref[#1]{\darkredtext{Thm.~\ref*{#1}}}}
\providecommand{\defref}[1]{\hyperref[#1]{\darkredtext{Def.~\ref*{#1}}}}
\providecommand{\propref}[1]{\hyperref[#1]{\darkredtext{Prop.~\ref*{#1}}}}
\providecommand{\assumpref}[1]{\hyperref[#1]{\darkredtext{Assump.~\ref*{#1}}}}
\providecommand{\remarkref}[1]{\hyperref[#1]{\darkredtext{Rem.~\ref*{#1}}}}
\providecommand{\hypref}[1]{\hyperref[#1]{\darkredtext{Hyp.~\ref*{#1}}}}
\providecommand{\conjref}[1]{\hyperref[#1]{\darkredtext{Conj.~\ref*{#1}}}}
\providecommand{\lemref}[1]{\hyperref[#1]{\darkredtext{Lem.~\ref*{#1}}}}
\providecommand{\corref}[1]{\hyperref[#1]{\darkredtext{Cor.~\ref*{#1}}}}
\providecommand{\noteref}[1]{\hyperref[#1]{\darkredtext{Nota.~\ref*{#1}}}}
\providecommand{\claimref}[1]{\hyperref[#1]{\darkredtext{Clm.~\ref*{#1}}}}
\providecommand{\obsref}[1]{\hyperref[#1]{\darkredtext{Obs.~\ref*{#1}}}}
\providecommand{\tabref}[1]{%
  \hyperref[#1]{\darkredtext{Tab.~\ref*{#1}}}}
\providecommand{\appref}[1]{%
  \hyperref[#1]{\darkredtext{App.~\ref*{#1}}}}

\theoremstyle{plain}

\theoremstyle{definition}

\theoremstyle{remark}

\tcbset{
  llada theorem box/.style={
    enhanced,
    breakable,
    sharp corners,
    boxrule=0pt,
    leftrule=2.5pt,
    colback=title_blue!4!white,
    colframe=title_blue,
    left=6pt,
    right=6pt,
    top=3pt,
    bottom=3pt,
    before skip=7pt,
    after skip=7pt
  },
  llada remark box/.style={
    llada theorem box,
    colback=statusorange!5!white,
    colframe=statusorange
  }
}

\tcolorboxenvironment{theorem}{llada theorem box}
\tcolorboxenvironment{proposition}{llada theorem box}
\tcolorboxenvironment{lemma}{llada theorem box}
\tcolorboxenvironment{corollary}{llada theorem box}
\tcolorboxenvironment{conjecture}{llada theorem box}
\tcolorboxenvironment{definition}{llada theorem box}
\tcolorboxenvironment{assumption}{llada theorem box}
\tcolorboxenvironment{hypothesis}{llada theorem box}
\tcolorboxenvironment{notation}{llada theorem box}
\tcolorboxenvironment{claim}{llada theorem box}
\tcolorboxenvironment{problem}{llada theorem box}
\tcolorboxenvironment{observation}{llada theorem box}
\tcolorboxenvironment{remark}{llada remark box}

\title{LLaDA-Image: Building Strong Image Generators with \\ Fully Open Training Recipes}

\author{
\centering
Chuyan Chen,
Haoxing Chen$^{*}$,
Kun Chen$^{*}$,
Zhenglin Cheng$^{\blacklozenge}$,
Long Cui,
Ruishan Fang,
Zhangxuan Gu,
Zhicheng Huang$^{*}$,
Zhenzhong Lan$^{\dag}$,
Yuanting Lei,
Haoquan Li,
Jianguo Li$^{\dag}$,
Rongchuan Li,
Sidu Li,\\
Tao Lin$^{\dag}$,
Deyuan Liu,
Jiacheng Liu$^{\blacklozenge}$,
Lin Liu,
Yuxuan Lou,
Zhisheng Lu,
Yuxin Ma,
Shuheng Shen,\\
Peng Sun$^{\blacklozenge}$,
Chaoyang Wang,
Hongjun Wang,
Xiaomei Wang$^{*}$,
Yongxin Wang$^{*}$,
Chengzhang Wu,\\
Hongru Wu,
Jun Xie$^{\blacklozenge}$
}

\begingroup
\footnotetext{Authors are listed alphabetically by surname. $\blacklozenge$ indicates core contributors to the image generators. $*$ indicates core contributors to the understanding backbone. $\dag$ indicates tech leaders.}
\endgroup

\affiliation{AGI Research Center, Inclusion AI}

\begin{document}
\maketitle

\begin{abstract}
    We introduce \textbf{\lladaimage}, a unified framework that pairs a 6B Diffusion Transformer (DiT) trained from scratch with a frozen vision-language understanding module built on the LLaDA2.0-Mini diffusion language model backbone. Instead of relying heavily on paired image--text data from the beginning, we first build a strong visual generative prior through image-only pre-training and mid-training. The generation pipeline comprises \textbf{220M samples, 98\% of which are real images.} For efficient and scalable optimization, we use parameter-free RMSNorm throughout the DiT together with the Muon optimizer. The resulting unified model produces highly photorealistic images while accurately following fine-grained editing instructions. We further distill \lladaimage into \lladaimageturbo, enabling fast inference in 2--4 sampling steps. On Qwen-Image-Bench, \lladaimage achieves overall scores of 53.53 and 53.38 on the English and Chinese tracks, respectively, setting a new state-of-the-art among open-source models on both tracks. To support further research on capable and efficient generative models, we release our model weights, training code, and detailed recipes.
\end{abstract}

\vspace{-4mm}
\begin{center}
    \renewcommand{\arraystretch}{1.4}
    \begin{tabular}{rll}
        \github{} & \textbf{GitHub}                  & \url{https://github.com/inclusionAI/LLaDA-Image}                  \\
        \hf{}     & \textbf{HuggingFace} & \url{https://huggingface.co/collections/inclusionAI/llada-image}
    \end{tabular}
\end{center}

\paragraph{Scope.}
This report asks a practical systems question: what does it take to train a
strong image generator from scratch with a fully open recipe? We trace the
complete path from image-only visual-prior learning through language alignment
and joint generation--editing to TwinFlow distillation. For each stage, we
document the data construction, model interfaces, optimization choices,
training schedule, and evaluation used to produce the released checkpoints.
Our goal is to make the resulting design trade-offs inspectable and the recipe
reusable. We do not claim that every component is universally optimal, nor do
we attempt an exhaustive comparison of all architectures or data mixtures.

\begin{figure}[H]
    \centering
    \includegraphics[width=\linewidth]{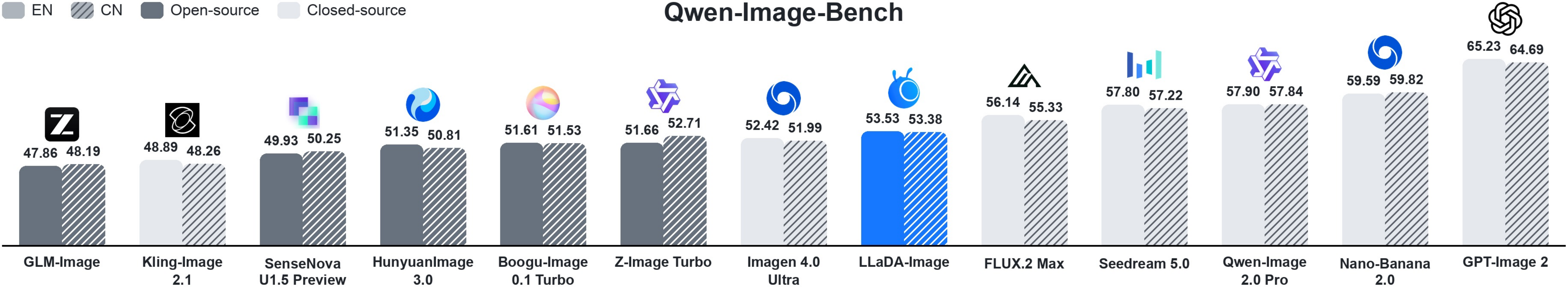}
    \caption{\textbf{Qwen-Image-Bench Results.} LLaDA-Image achieves open-source SOTA
        on both English and Chinese tracks; models are ordered by their two-track average.}
    \label{fig:qwen-image-bench-overall}
\end{figure}

\begingroup

\newcommand{\visualdemoimage}[2][0.80\textheight]{%
    \includegraphics[
        width=0.98\linewidth,
        height=#1,
        keepaspectratio
    ]{#2}%
}

\newcommand{\comparisonprompttext}[1]{%
    \begingroup
    \ifcase#1\relax
        \def\comparisonpromptid{0}%
    \or \def\comparisonpromptid{4}%
    \or \def\comparisonpromptid{3}%
    \or \def\comparisonpromptid{1}%
    \or \def\comparisonpromptid{9}%
    \or \def\comparisonpromptid{5}%
    \or \def\comparisonpromptid{6}%
    \or \def\comparisonpromptid{7}%
    \or \def\comparisonpromptid{8}%
    \else \def\comparisonpromptid{0}%
    \fi
    
\begin{CJK*}{UTF8}{gbsn}
\ifcase\comparisonpromptid\relax
  \textcolor{red}{[Missing comparison prompt identifier.]}
\or
  一尊汉代陶制侍女俑独立摆放在博物馆展台上，人物双手收于身前，长袍垂直落下，造型简洁克制，面部五官以古朴方式塑造，表面残留少量褪色的朱红、黑色和彩绘颜料，陶土呈暖灰褐色，可见细小裂纹、土沁与局部缺损。柔和自然光从左侧落下，背景为温暖的灰米色墙面，突出人物轮廓与历史痕迹，真实考古摄影，垂直构图
\or
  一只明代风格青花瓷梅瓶独立陈列，瓶身修长，肩部饱满，白色瓷胎略带温润象牙色，钴蓝色缠枝莲纹沿瓶身自然展开，青花颜色有明显浓淡变化与晕散效果，釉面带细微反光和真实开片质感。背景采用低饱和米灰色博物馆空间，底部是深色木质展台，一束柔和侧光勾勒瓶身轮廓，画面干净典雅，考古图录式摄影，垂直构图
\or
  一幅垂直构图的中国传统浅绛山水画轴，画面描绘春雨过后的江南山村。近景是一段湿润的石岸，几株柳树从画面左下方向上生长，细长柳枝自然下垂，嫩叶以极淡的青绿色点染，树下散落着灰褐色石块和刚刚冒出的青草。中景是一座横跨溪流的石拱桥，桥后分布数间白墙灰瓦的江南民居，屋檐和墙角被薄雾部分遮挡，溪水以大片留白与几道浅墨水纹表现。远处低矮山峦沿画面向上层层递进，以湿润淡墨晕染，山腰笼罩着细腻雾气。右上角用略带行书意味的楷书分两列竖排题写，右列从上至下写“沾衣欲湿杏花雨”，左列从上至下写“吹面不寒杨柳风”，两列字距疏朗，排列整齐，墨色自然。纸本呈柔和的暖米黄色，可见细微纸纹、墨色渗化与自然旧化痕迹，装裱采用低饱和青灰色绫边，整体色调清淡雅致，具有真实传统中国画的湿润空气感。
\or
  一张阴天海边小镇中的旅行女性人像摄影，采用偏远一些的中景构图。一位二十多岁的年轻白人女性沿着靠海的人行道行走，两只手插在风衣口袋里，人物并没有正对镜头，而是在走动过程中自然侧过脸。她留着浅棕色偏金的中长卷发，海风将大量发丝吹向一侧，发型明显凌乱，不呈现精心整理的波浪造型。脸型偏长椭圆，额头较宽，颧骨轻微明显，下颌柔和。眉毛为自然浅棕色，毛发并不浓密。眼睛呈灰蓝色，中等大小，视线看向道路前方。鼻梁较高，鼻尖具有真实体积。嘴唇偏薄，嘴角自然闭合。皮肤白皙没有雀斑，眼下存在淡淡青灰阴影，皮肤表面保持自然纹理，并非光滑陶瓷质感。她穿深蓝色棉质风衣、灰色针织衫和浅色牛仔裤。背景是灰蓝色海面、石墙、低矮住宅、停泊的小渔船和被风吹动的草丛，阴天天光平坦而柔和，人物与环境都有清晰空间关系，整体更接近真实旅行纪实摄影而不是人像写真。
\or
  一张冰川观景区的男性旅行游客照，拍摄于晴朗上午。主体是一位三十岁左右的东亚男性，戴着深灰色运动墨镜，站在木质观景平台中央，双手带着手套高高张开，头微微抬起，脸上露出自然放松的笑容。人物脸型偏方长，额头较宽，黑色短发被风吹得自然蓬松。鼻梁较直，鼻尖略宽，嘴唇厚度适中。皮肤呈自然偏暖的肤色，面颊有轻微日晒痕迹，整体保持偏哑光质感。他穿橙棕色户外夹克、黑色T恤、深灰色户外长裤和登山鞋，肩上背着一个小型登山包。背景是一条从巨大雪山之间缓慢向山谷延伸的冰川，冰川表面呈浅蓝白色，可以看到自然裂隙和灰色碎石带，两侧山体为黑灰色岩石与积雪交错。天空蓝得清透，几朵白云位于主峰附近，平台周围还有护栏和少量游客，整体规模感明确，同时保持普通游客旅行纪念照的自然感。
\or
  一张写实风格的欧洲山地小镇风光摄影作品，拍摄于清晨。画面主体是一座依山而建的地中海小镇，成排的浅黄色、米白色和暖橙色石砌房屋顺着山坡层层向上延伸，屋顶多为红褐色瓦片，窗户尺寸不一，部分窗台摆放着绿植和花盆。狭窄弯曲的石板街巷在房屋之间穿行，几段台阶沿着坡地向上连接不同高度的平台。画面下方是小镇边缘的露台和护墙，远处能看到蓝灰色群山和一小片平静水面。晨光从侧面照亮建筑立面，墙面有清楚但柔和的明暗变化，整体色彩自然克制，空气清新，像真实旅行中拍到的欧洲山城景象。
\or
  一张写实风格的江南水乡城镇风光摄影作品，画面以一条蜿蜒水道为主线，两侧分布着白墙灰瓦的传统民居和临水店铺，建筑高低错落，沿着微微起伏的地势延伸。前景是一座低矮石拱桥，桥下有一艘窄长木船缓慢划过，船上坐着游客。中景是临河石板路、木窗、黑色檐口、晾晒的衣物和零散绿植，几家店铺门前挂着简单招牌，其中一块写着 "茶馆"。远处可以看到更密集的屋顶、树木和薄雾中的小桥。整体光线柔和自然，画面既有水乡层次，也有真实生活气息。
\or
  一张写实风格的海岸悬崖广角风景摄影作品，采用高处俯瞰视角。前景是一片被海风侵蚀得起伏不平的深绿色草坡，草丛间裸露出灰褐色岩石，几条狭窄步道沿着悬崖边缘蜿蜒向远处。中景是数百米高的巨大海蚀悬崖，岩壁呈深灰与褐色，垂直切入深蓝色海面，局部能看到明显岩层和被海浪冲刷形成的凹槽。海水不断拍击崖脚，形成细长白色浪带和大片泡沫。远处连续数个岬角逐渐消失在蓝灰色空气透视中，天空开阔，大片白云与蓝色天空交错。画面不出现人物和建筑，突出海岸线的巨大尺度、风力感和开阔空间。
\or
  明1368-1644年掐丝珐琅海棠式双耳瓶，浅灰背景上，一件明代掐丝珐琅海棠式双耳瓶居中摆放。器物通体呈海棠四瓣花形，由口至足轮廓一致，形似四个铜瓶接连合成。瓶口呈四瓣花形，边缘鎏金。颈部两侧对称装饰两只立体鎏金螭龙耳，龙身卷曲，龙首探向瓶口。颈部与腹部之间有一圈凸起的变形莲瓣纹饰带。瓶身主体以蓝绿色为地，表面布满掐丝珐琅工艺烧制的垂拱式夔龙纹，纹饰线条细密，色彩包含蓝、绿、红、黄等，间以金色丝线勾勒，质感古朴。瓶底为四瓣花形圈足，足边亦见鎏金痕迹。整体器型端庄，工艺繁复，色彩斑斓。
\else
  \textcolor{red}{[Missing comparison prompt identifier.]}
\fi
\end{CJK*}
    \endgroup
}

\newcommand{\comparisonpromptfont}{\fontsize{5.0pt}{5.3pt}\selectfont}

\newcommand{\modeldemoimage}[2]{%
    \begin{minipage}[b][0.20\textheight][t]{0.245\linewidth}
        \centering
        \begin{minipage}[c][0.185\textheight][c]{\linewidth}
            \centering
            \IfFileExists{#2}{%
                \includegraphics[width=0.94\linewidth,height=0.185\textheight,keepaspectratio]{#2}%
            }{%
                \setlength{\fboxsep}{0pt}%
                \fbox{%
                    \begin{minipage}[c][0.185\textheight][c]{0.92\linewidth}
                        \centering
                        {\scriptsize\textcolor{gray}{Portrait Image}}
                    \end{minipage}%
                }%
            }%
        \end{minipage}%
        \par\vspace{0.1em}
        {\scriptsize\bfseries #1\par}
    \end{minipage}%
}

\newcommand{\modeldemoprompt}[2]{%
    \begin{minipage}[b][0.20\textheight][t]{0.245\linewidth}
        \begin{tcolorbox}[
            width=\linewidth,
            height=0.185\textheight,
            valign=center,
            enhanced,
            sharp corners,
            boxrule=0.6pt,
            colback=secondcolor!4!white,
            colframe=secondcolor!60!black,
            left=6pt,
            right=6pt,
            top=5pt,
            bottom=5pt
        ]
            {\comparisonpromptfont #2\par}
        \end{tcolorbox}
        \par\vspace{0.1em}
        {\scriptsize\bfseries Input Prompt\par}
    \end{minipage}%
}

\newcommand{\modelcomparisongrid}[2]{%
    \modeldemoprompt{#1}{\comparisonprompttext{#1}}
    \hfill
    \modeldemoimage{LLaDA-Image Turbo}{figs/demo-comparison/prompt#1-model1.jpg}
    \hfill
    \modeldemoimage{Z-Image Turbo}{figs/demo-comparison/prompt#1-model2.jpg}
    \hfill
    \modeldemoimage{Boogu-Image 0.1 Turbo}{figs/demo-comparison/prompt#1-model3.jpg}
    \par\vspace{0.2em}
    \modeldemoimage{SenseNova U1.5 Preview}{figs/demo-comparison/prompt#1-model4.jpg}
    \hfill
    \modeldemoimage{Qwen-Image 2512}{figs/demo-comparison/prompt#1-model5.jpg}
    \hfill
    \modeldemoimage{GPT-Image 2}{figs/demo-comparison/prompt#1-model6.jpg}
    \hfill
    \modeldemoimage{Gemini 3.1 Flash Image}{figs/demo-comparison/prompt#1-model7.jpg}

    \vspace{0.35em}
    {\color{gray!45}\rule{0.94\linewidth}{0.4pt}\par}
    \vspace{0.35em}

    \modeldemoprompt{#2}{\comparisonprompttext{#2}}
    \hfill
    \modeldemoimage{LLaDA-Image Turbo}{figs/demo-comparison/prompt#2-model1.jpg}
    \hfill
    \modeldemoimage{Z-Image Turbo}{figs/demo-comparison/prompt#2-model2.jpg}
    \hfill
    \modeldemoimage{Boogu-Image 0.1 Turbo}{figs/demo-comparison/prompt#2-model3.jpg}
    \par\vspace{0.2em}
    \modeldemoimage{SenseNova U1.5 Preview}{figs/demo-comparison/prompt#2-model4.jpg}
    \hfill
    \modeldemoimage{Qwen-Image 2512}{figs/demo-comparison/prompt#2-model5.jpg}
    \hfill
    \modeldemoimage{GPT-Image 2}{figs/demo-comparison/prompt#2-model6.jpg}
    \hfill
    \modeldemoimage{Gemini 3.1 Flash Image}{figs/demo-comparison/prompt#2-model7.jpg}%
}

\newcommand{\modelcomparisonpage}[3]{%
    \clearpage
    \nointerlineskip
    \thispagestyle{plain}
    \begin{center}
        {\LARGE\bfseries #1\par}
        \vspace{0.2em}
        \modelcomparisongrid{#2}{#3}
    \end{center}%
}

\newgeometry{top=0.9cm,bottom=1.9cm,left=2cm,right=2cm}
\nointerlineskip
\thispagestyle{plain}
\begin{center}
    {\LARGE\bfseries Real or Generated? A Reader Challenge}
    \vspace{0.8em}
    \begin{tcolorbox}[
            width=0.94\linewidth,
            enhanced,
            sharp corners,
            boxrule=0pt,
            leftrule=3pt,
            colback=secondcolor!5!white,
            colframe=secondcolor,
            left=8pt,
            right=8pt,
            top=6pt,
            bottom=6pt
        ]
        \centering
        Which images, if any, are real photographs? Inspect the details,
        make your selections, and lock in your answer before turning the
        page.
    \end{tcolorbox}
    \vspace{0.8em}
    \visualdemoimage{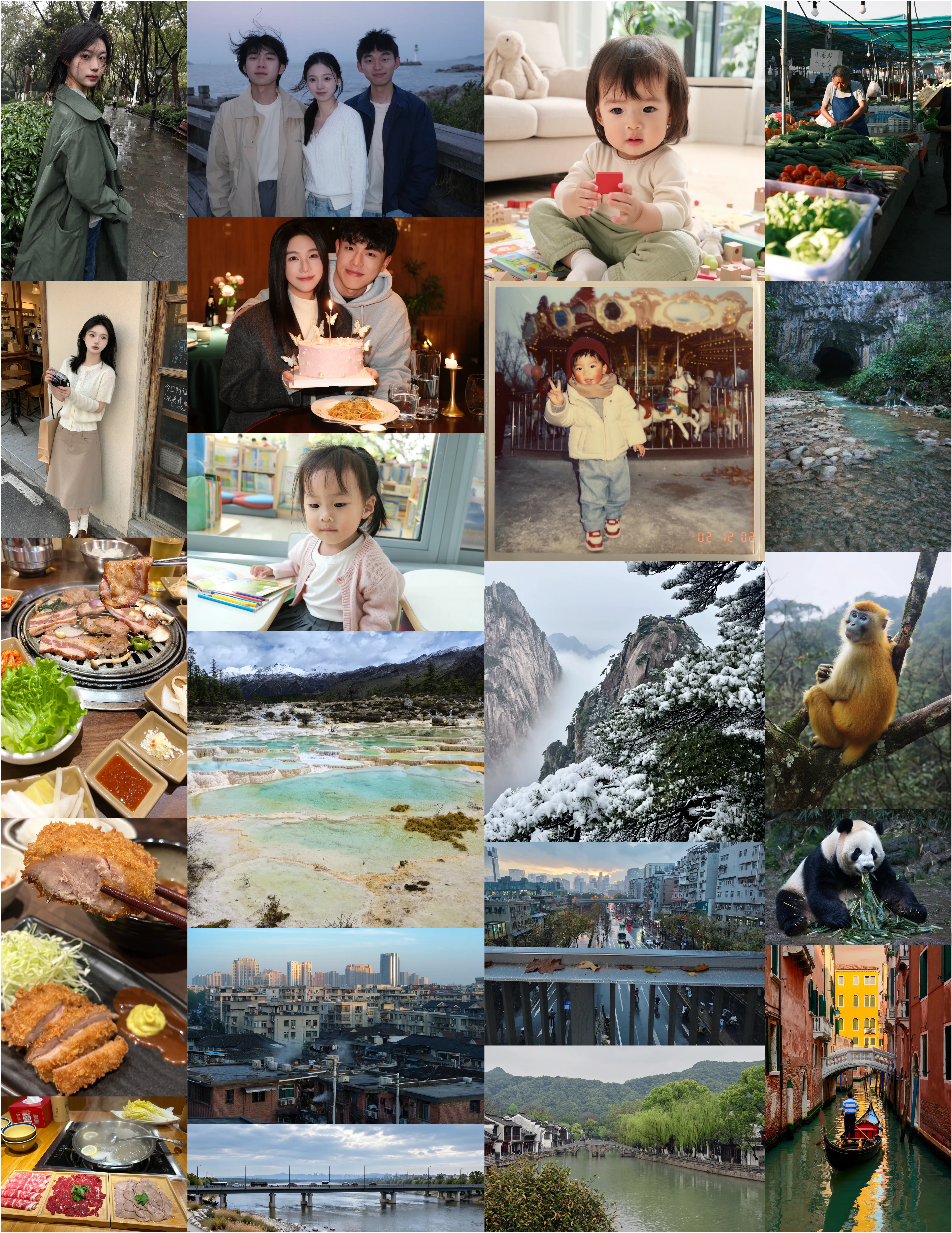}\par
    \vspace{0.7em}
    {\small\itshape
        Your task: identify every real image in the grid---if there are any.
        The answer appears on the next page.\par}
\end{center}

\clearpage
\nointerlineskip
\thispagestyle{plain}
\begin{center}
    {\LARGE\bfseries The Reveal: None of Them Are Real\par}
    \vspace{0.6em}
    {\small
        Every candidate in the challenge was produced by \lladaimage;
        no real images were included.\par}
    \vspace{0.8em}
    \begin{tcolorbox}[
            width=0.94\linewidth,
            enhanced,
            sharp corners,
            boxrule=0pt,
            leftrule=3pt,
            colback=firstcolor!5!white,
            colframe=firstcolor,
            left=8pt,
            right=8pt,
            top=6pt,
            bottom=6pt
        ]
        \centering
        \textbf{Answer: every candidate was generated.}
        The apparent photographs, rendered text, and stylized scenes were all
        synthesized by the same unified model.
    \end{tcolorbox}
    \vspace{0.8em}
    \visualdemoimage[0.77\textheight]{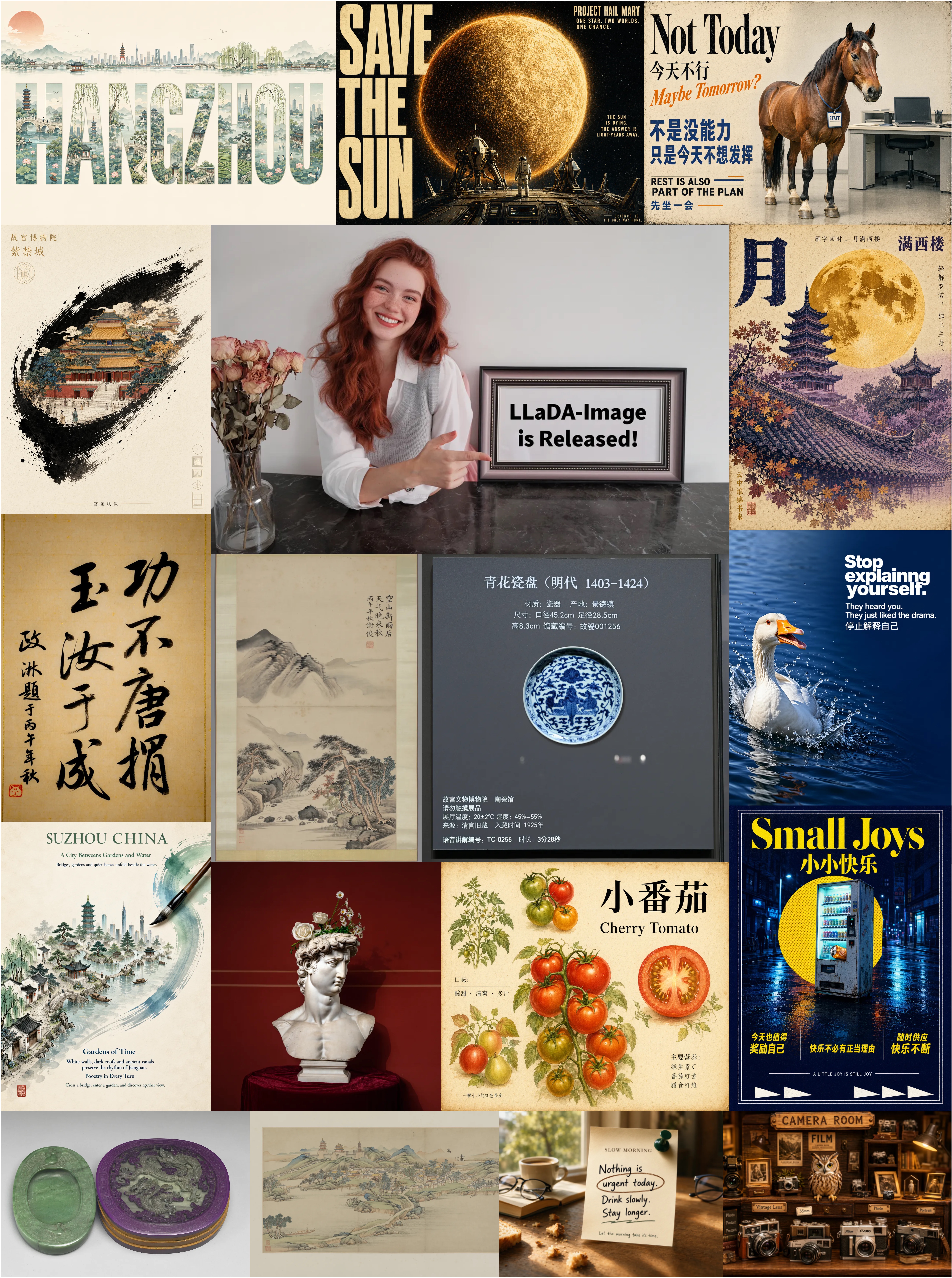}\par
    \vspace{0.7em}
    {\small
        \textbf{Disclosure.} All results displayed on these two pages are
        generated outputs, spanning text-to-image synthesis, bilingual text
        rendering. This informal reader challenge is a qualitative demonstration, not a
        controlled perceptual study.\par}
\end{center}

\restoregeometry
\modelcomparisonpage{Cross-Model Qualitative Comparison}{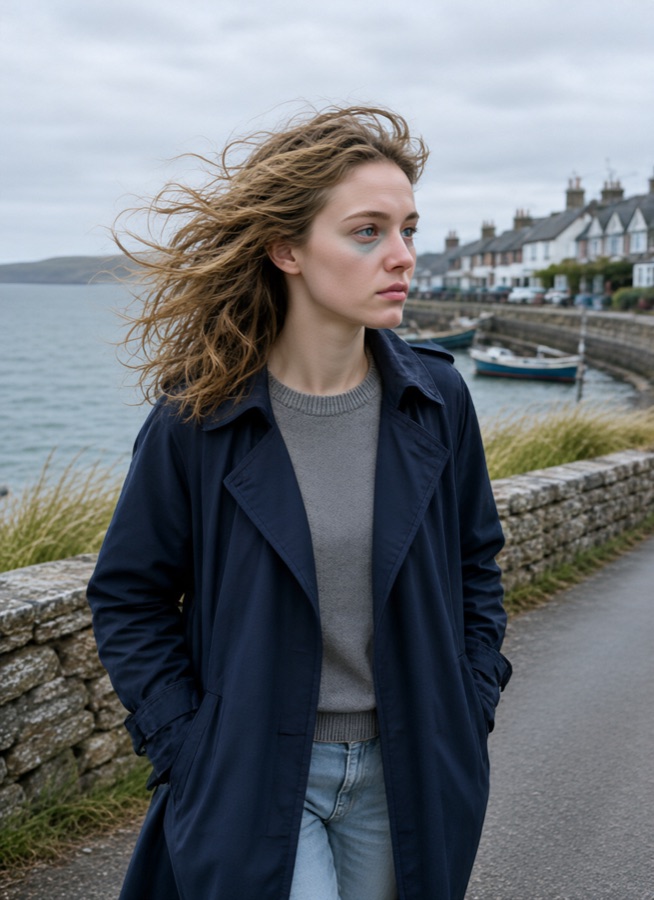}{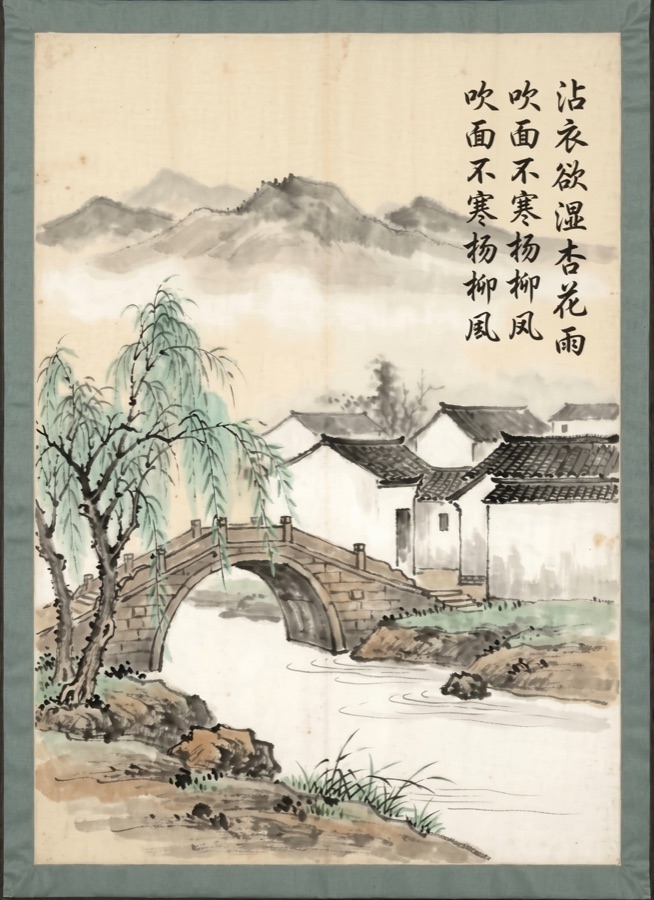}

\clearpage
\newgeometry{top=0.9cm,bottom=0.8cm,left=2cm,right=2cm}
\nointerlineskip
\thispagestyle{plain}

\newcommand{\editinggridimage}[1]{%
    \centering
    \IfFileExists{#1}{%
        \includegraphics[
            width=\linewidth,
            height=0.14\textheight,
            keepaspectratio
        ]{#1}%
    }{%
        \colorbox{black!2}{%
            \parbox[c][0.135\textheight][c]{\dimexpr\linewidth-2\fboxsep\relax}{%
                \centering\scriptsize\color{black!38}Image placeholder}}
    }%
}

\newcommand{\editinggridcase}[4]{%
    \begin{minipage}[t][0.205\textheight][t]{0.49\linewidth}
        \begin{tcolorbox}[
                width=\linewidth,
                height=0.042\textheight,
                valign=center,
                enhanced,
                sharp corners,
                boxrule=0pt,
                leftrule=2.5pt,
                colback=secondcolor!5!white,
                colframe=secondcolor,
                left=7pt,
                right=7pt,
                top=5pt,
                bottom=5pt
            ]
            \raggedright
            {\scriptsize\fontseries{bx}\selectfont Case \##1\quad #2\par}
        \end{tcolorbox}

        \vspace{0.55em}
        \begin{minipage}[c][0.14\textheight][c]{0.46\linewidth}
            \centering
            \editinggridimage{#3}
        \end{minipage}%
        \hfill
        \begin{minipage}[c][0.14\textheight][c]{0.045\linewidth}
            \centering
            {\color{secondcolor}\normalsize$\rightarrow$}
        \end{minipage}%
        \hfill
        \begin{minipage}[c][0.14\textheight][c]{0.46\linewidth}
            \centering
            \editinggridimage{#4}
        \end{minipage}%
    \end{minipage}%
}

\begin{center}
    {\LARGE\bfseries Instruction-Guided Image Editing\par}
    \vspace{0.25em}
    {\small
        \lladaimage follows diverse editing instructions while preserving
        unedited content.\par}

    \vspace{0.75em}
    \editinggridcase
    {1}
    {\begin{CJK*}{UTF8}{gbsn}把图片的风格改为吉卜力风格\end{CJK*}}
    {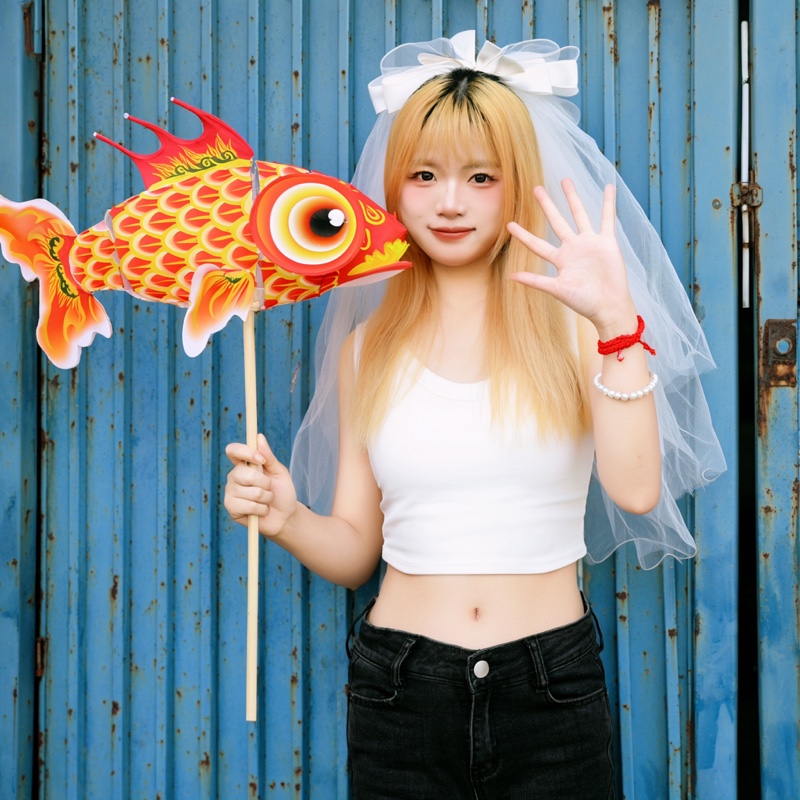}
    {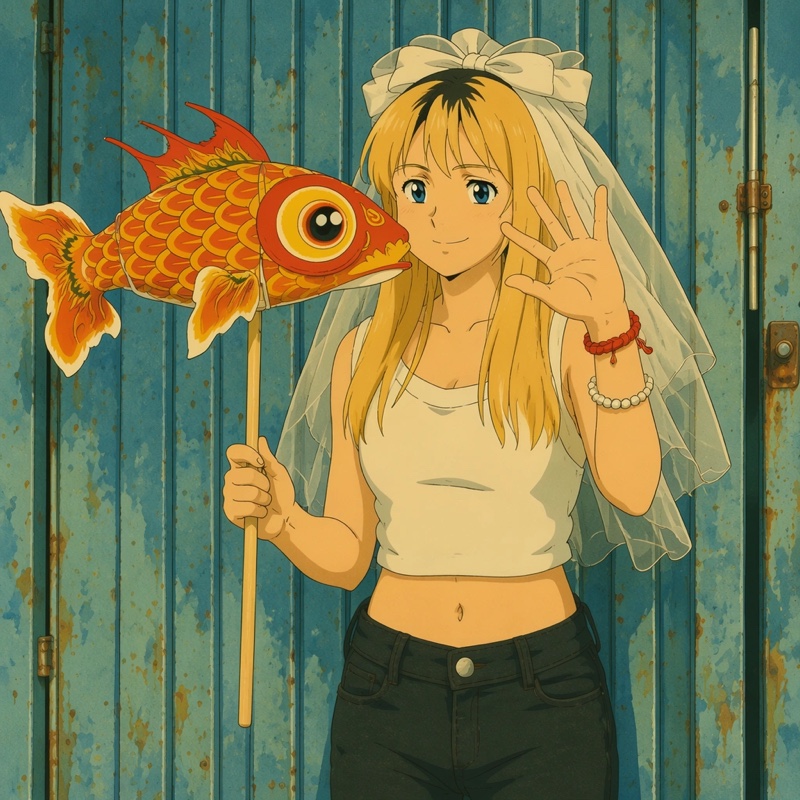}
    \hfill
    \editinggridcase
    {2}
    {\begin{CJK*}{UTF8}{gbsn}把 ``Stay / fresh'' 改成 ``Keep / fresh''\end{CJK*}}
    {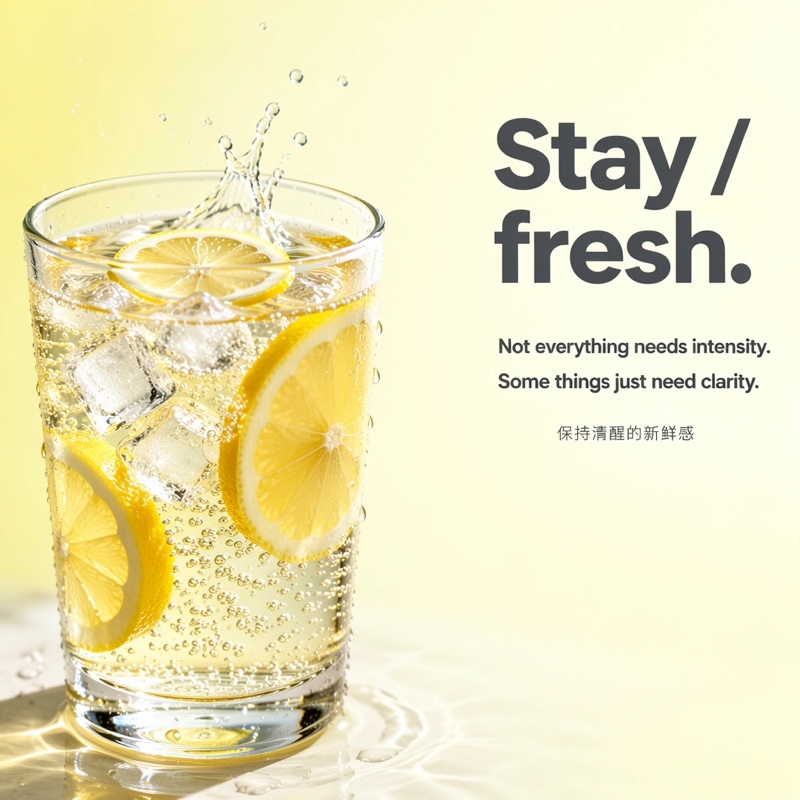}
    {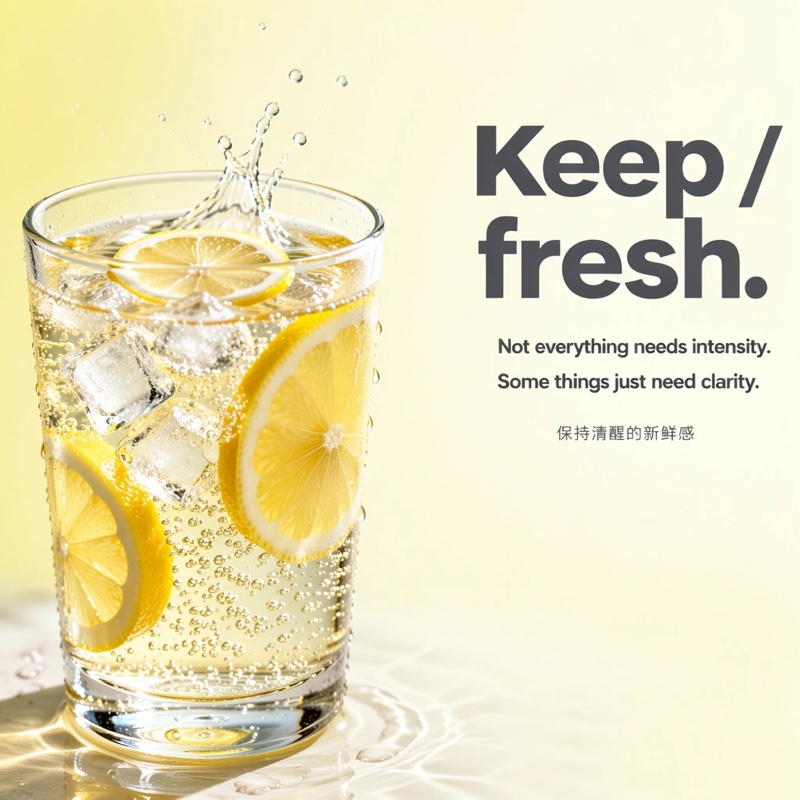}

    \vspace{0.55em}
    \editinggridcase
    {3}
    {\begin{CJK*}{UTF8}{gbsn}把图中女性身上红色的包去掉\end{CJK*}}
    {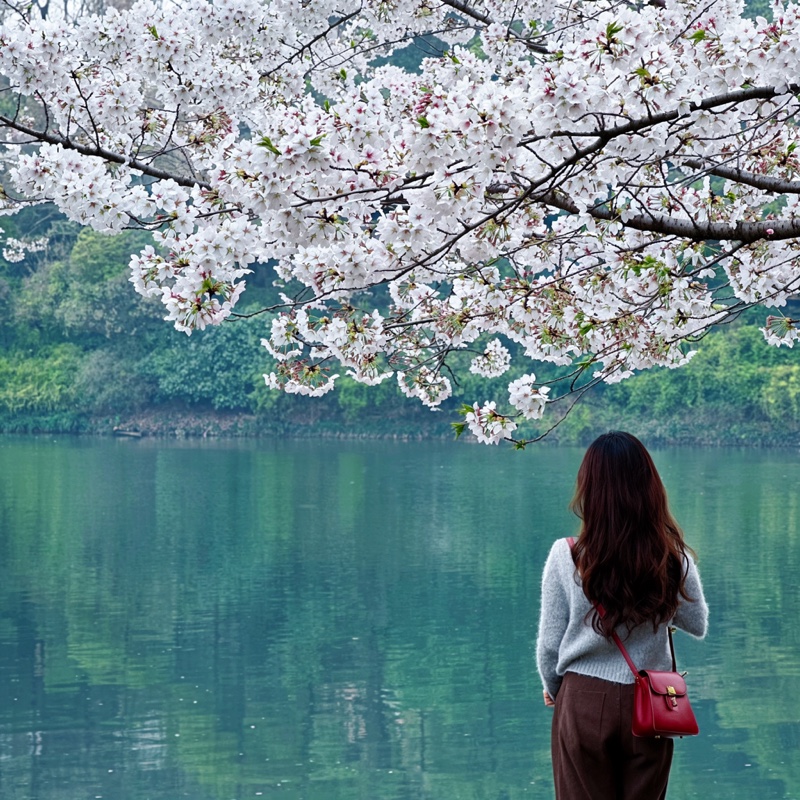}
    {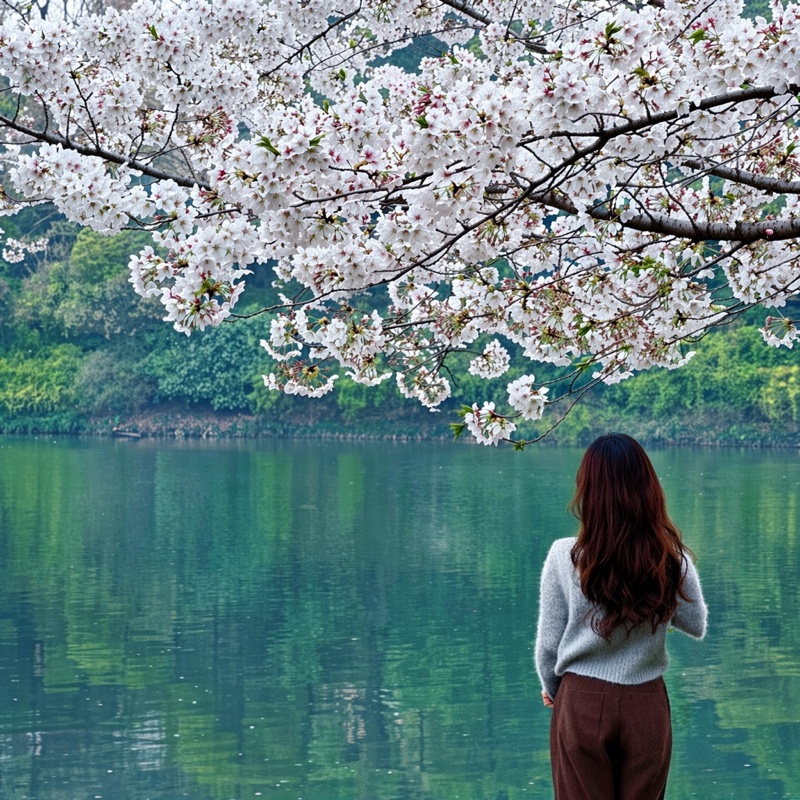}
    \hfill
    \editinggridcase
    {4}
    {\begin{CJK*}{UTF8}{gbsn}对这个手绘图进行上色，其中小女孩的肤色为肉色，头发改成黑色，小熊为棕色\end{CJK*}}
    {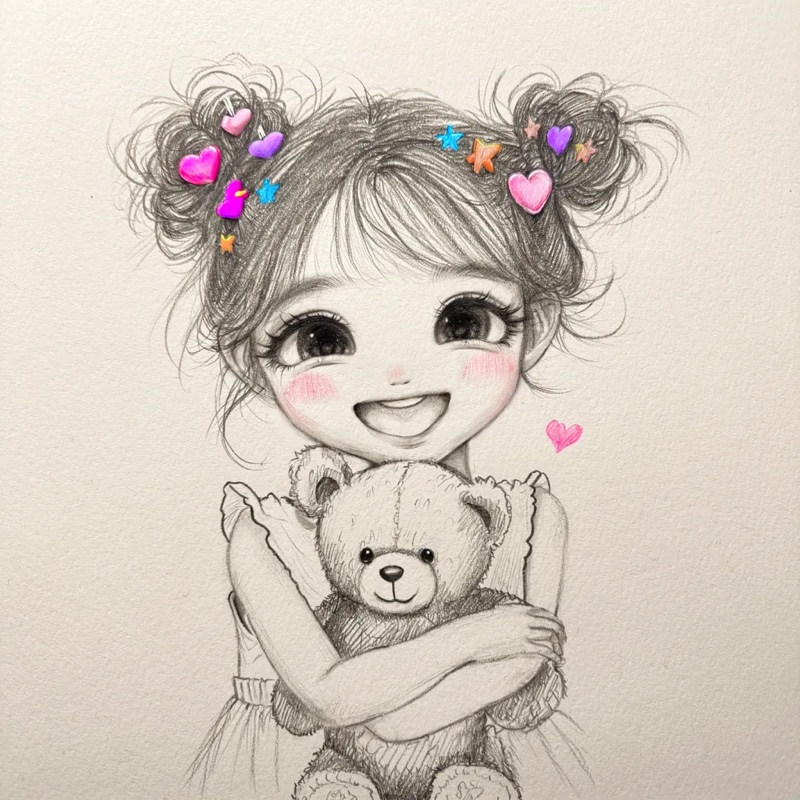}
    {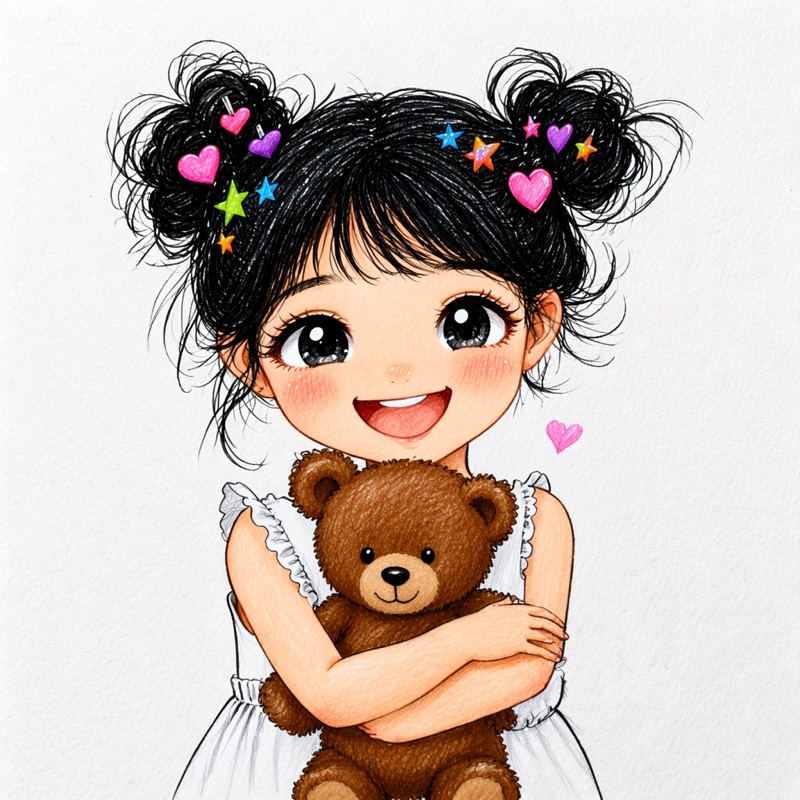}

    \vspace{0.55em}
    \editinggridcase
    {5}
    {\begin{CJK*}{UTF8}{gbsn}让她单手比一个“V”字手势，保持背景不变\end{CJK*}}
    {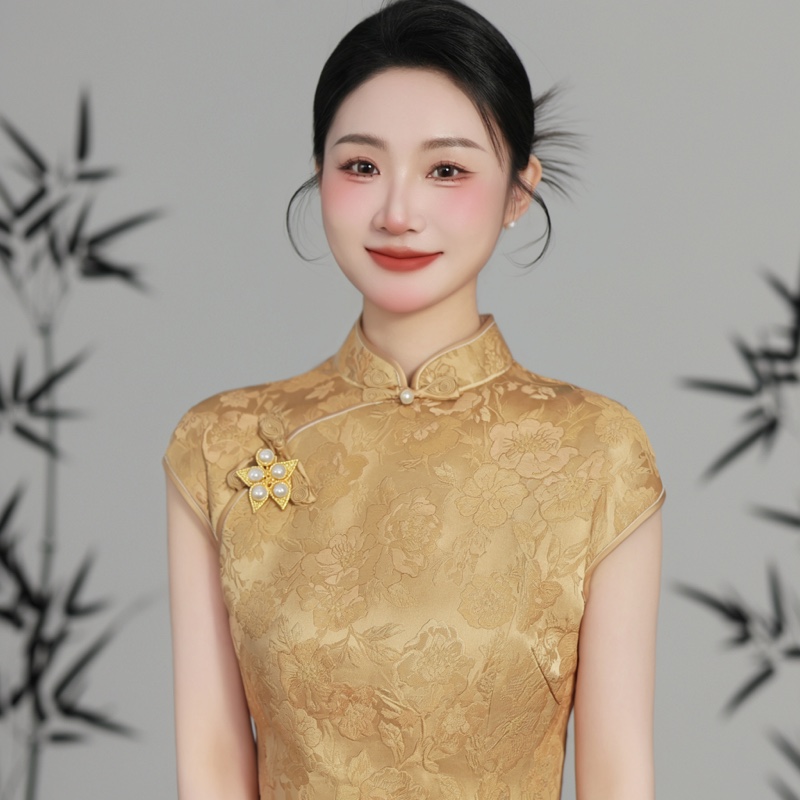}
    {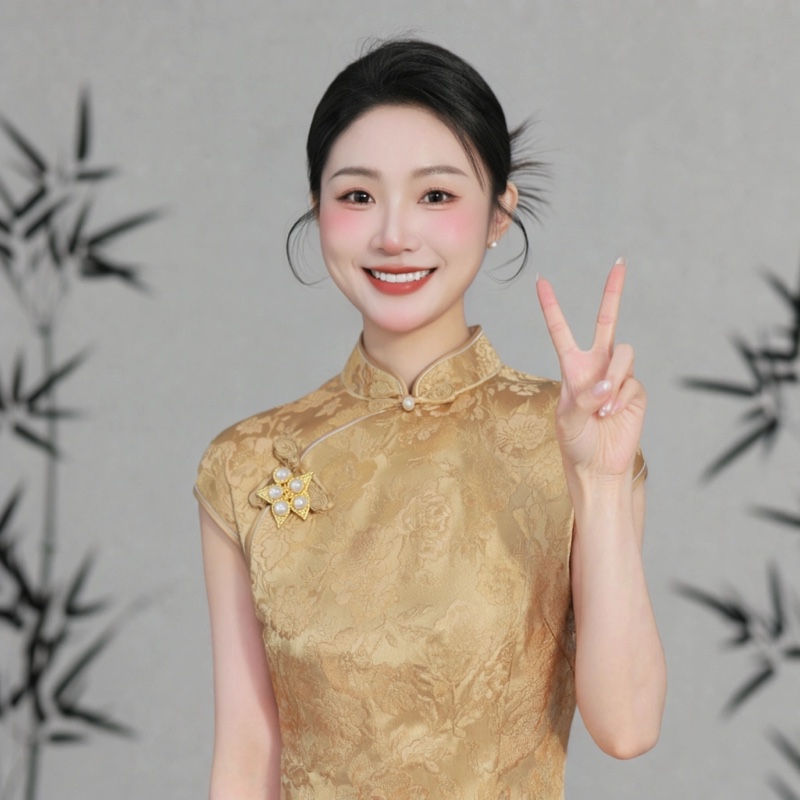}
    \hfill
    \editinggridcase
    {6}
    {\begin{CJK*}{UTF8}{gbsn}把这个背景换成绿色的大草原，草原上还有蒙古包\end{CJK*}}
    {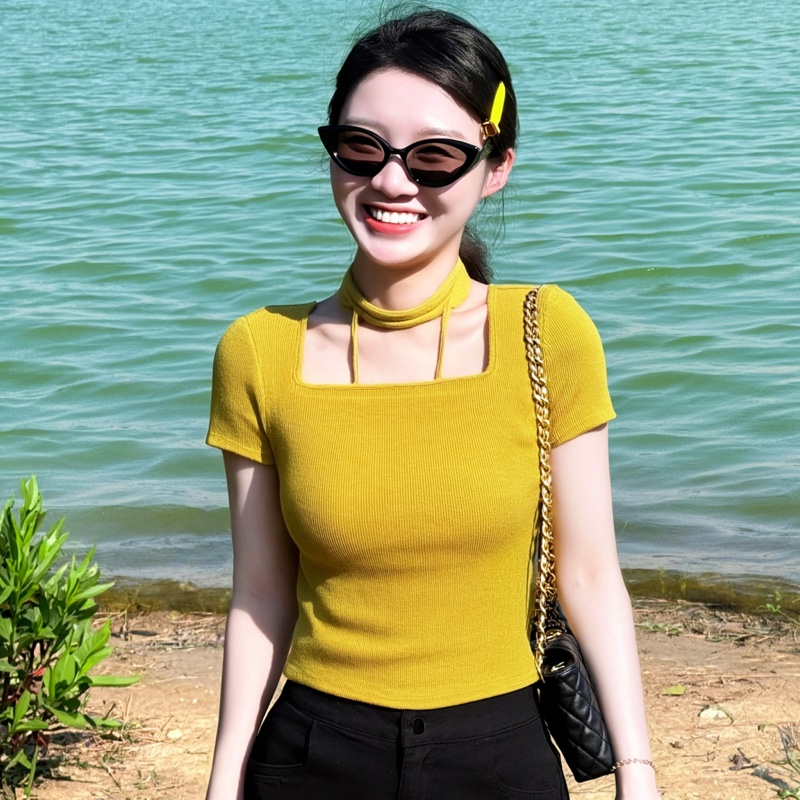}
    {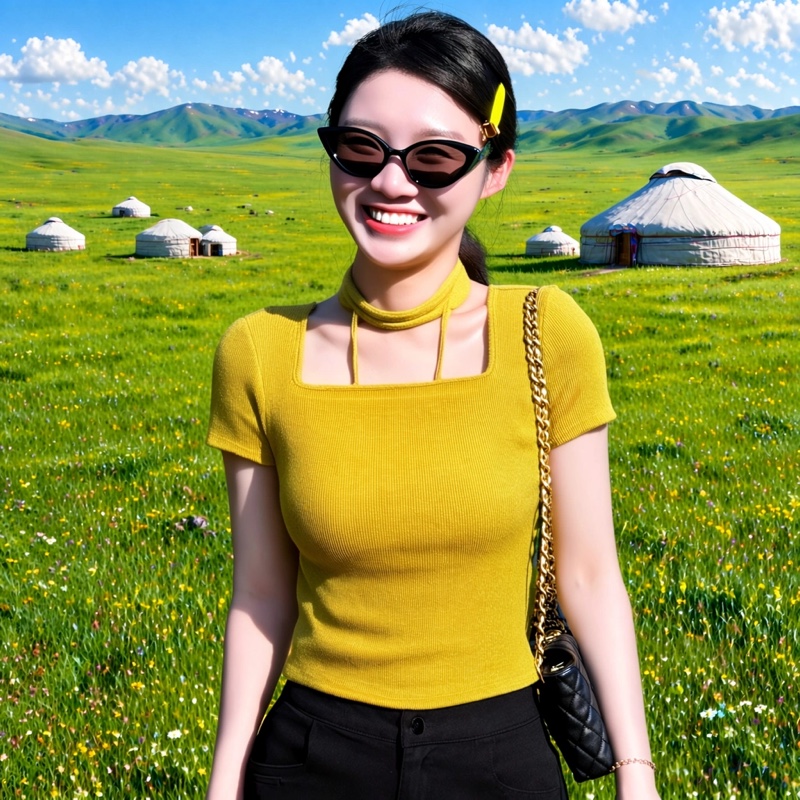}

    \vspace{0.55em}
    \editinggridcase
    {7}
    {\begin{CJK*}{UTF8}{gbsn}在这个地毯上去添加一个茶几\end{CJK*}}
    {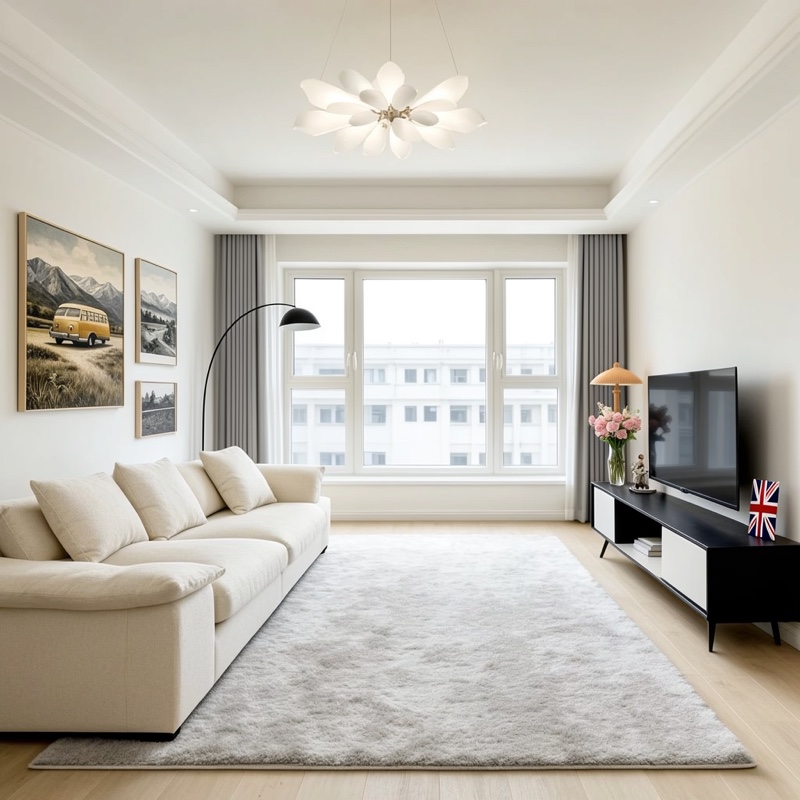}
    {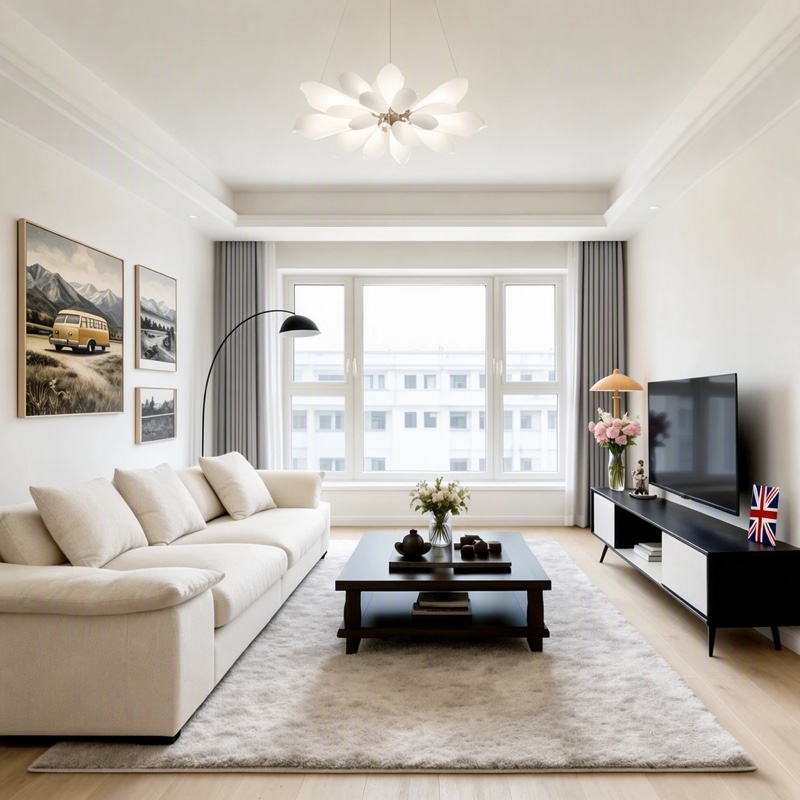}
    \hfill
    \editinggridcase
    {8}
    {\begin{CJK*}{UTF8}{gbsn}把这个蓝色的冰淇淋换成草莓味圣代\end{CJK*}}
    {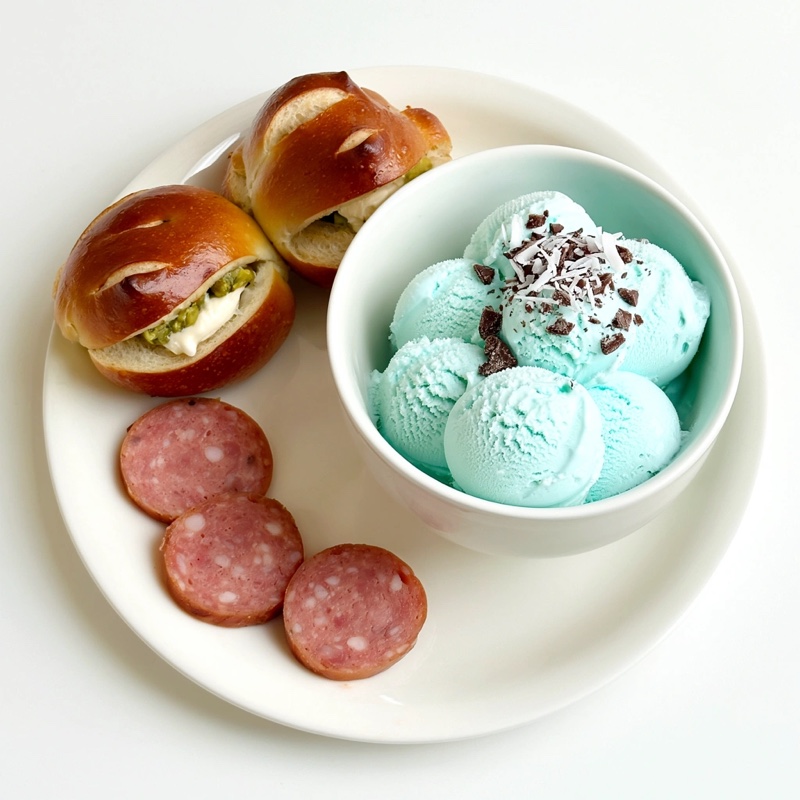}
    {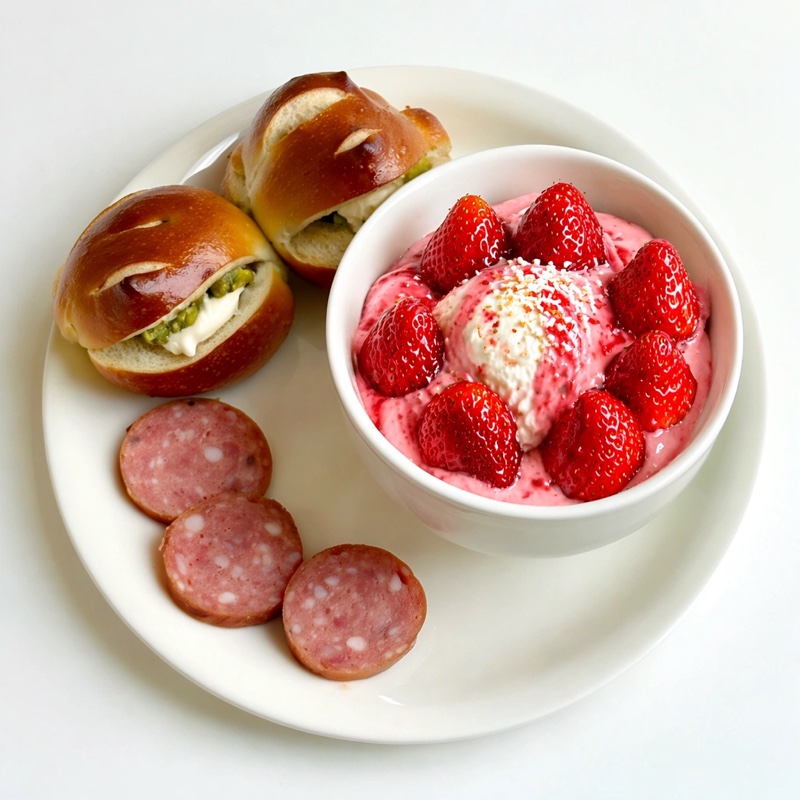}

\end{center}

\restoregeometry

\endgroup

\clearpage
\section*{Contents}
\makeatletter
\@starttoc{toc}
\makeatother

\clearpage
\section{Introduction}
\label{sec:introduction}

Image generation systems are evolving from specialized text-to-image models
into general-purpose visual creation systems~\citep{ai2026llada2,wu2025omnigen2,bagel}. They must understand complex and
multilingual instructions, synthesize photorealistic content, preserve
reference information during editing, render text accurately, and remain
usable under practical inference budgets. Although proprietary systems have
made impressive progress, their data, model, and training recipes remain
largely inaccessible. For open models, the goal is therefore not generation
quality alone: a useful system must combine broad capabilities with an
attainable data budget, stable large-scale training, and efficient deployment.

These requirements create linked bottlenecks across data, model design, and
deployment. Prevailing training paradigms~\citep{rombach2022latentdiffusion,esser2024scaling} couple visual-prior learning with
language alignment from the outset, requiring paired captions during the most
compute-intensive stages. Yet captions are expensive and lossy; at low
training resolutions, details mentioned in a caption may disappear after
downsampling. Synthetic image--text pairs can accelerate early convergence,
but synthetic-heavy mixtures may propagate artifacts and limit long-run
realism. Meanwhile, a unified model must connect multimodal understanding to
generation while preserving reference-image evidence for editing, and
deployment requires compressing long diffusion trajectories without quality
loss. Addressing these challenges together requires co-designing supervision,
architecture, the training pipeline, and distillation.

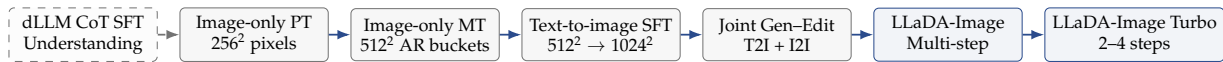
\begin{figure}[H]
    \centering
    \resizebox{0.97\linewidth}{!}{%
        \begin{tikzpicture}[
            node distance=3mm,
            stage/.style={
                    draw=black!55,
                    rounded corners=2pt,
                    align=center,
                    minimum width=2.05cm,
                    minimum height=0.72cm,
                    fill=black!3,
                    font=\scriptsize,
                    inner sep=3pt
                },
            release/.style={
                    stage,
                    draw=secondcolor,
                    fill=secondcolor!6!white
                },
            prerequisite/.style={
                    stage,
                    dashed,
                    fill=black!1
                },
            flow/.style={
            -{Latex[length=1.8mm]},
            line width=0.7pt,
            draw=secondcolor
            },
            prerequisite flow/.style={
                    flow,
                    dashed,
                    draw=black!55
                }
            ]
            \node[prerequisite] (cot) {dLLM CoT SFT\\Understanding};
            \node[stage, right=of cot] (pt) {Image-only PT\\$256^2$ pixels};
            \node[stage, right=of pt] (mt) {Image-only MT\\$512^2$ AR buckets};
            \node[stage, right=of mt] (t2i) {Text-to-image SFT\\$512^2 \rightarrow 1024^2$};
            \node[stage, right=of t2i] (joint) {Joint Gen--Edit\\T2I + I2I};
            \node[release, right=of joint] (base) {\lladaimage\\Multi-step};
            \node[release, right=of base] (turbo) {\lladaimageturbo\\2--4 steps};

            \draw[prerequisite flow] (cot) -- (pt);
            \draw[flow] (pt) -- (mt);
            \draw[flow] (mt) -- (t2i);
            \draw[flow] (t2i) -- (joint);
            \draw[flow] (joint) -- (base);
            \draw[flow] (base) -- (turbo);
        \end{tikzpicture}}
    \vspace{-0.5mm}
    \caption{\textbf{Training and release path.} CoT SFT prepares the frozen
        understanding backbone (dashed); the solid path trains the generator and
        yields the Base and TwinFlow-distilled Turbo checkpoints. PT and MT denote pre-training and mid-training, respectively.}
    \label{fig:training-release-path}
\end{figure}

We present \lladaimage, a 6B-parameter Diffusion Transformer trained from
scratch, together with the complete model and training stack behind it. The
system first learns its visual prior through image-only pre-training and
mid-training, then progressively introduces language alignment, refinement,
and joint generation--editing training. A dLLM-based understanding module and
a single-stream DiT support understanding, generation, and editing within one
framework, while TwinFlow produces the few-step \lladaimageturbo variants.
The principal characteristics of the system are summarized as follows:
\begin{itemize}[leftmargin=*,itemsep=4pt,topsep=4pt]
    \item \textbf{Data-efficient image-only pre-training and mid-training.}
          We decouple visual-prior learning from language alignment and avoid
          large-scale paired supervision in the early training stages. A frozen
          dLLM-based vision-language model derives the condition from the same image
          region that the DiT learns to generate, ensuring condition--target
          compatibility without an external caption and enabling informative crops
          of high-resolution images to be resized only mildly for $256^2$ training.
          The complete image-generation pipeline processes approximately 220M
          generation-training samples,
          with image-only training accounting for more than $90\%$ of the total, keeping
          the overall data requirement moderate.

    \item \textbf{Unified dLLM-based understanding, generation, and editing.}
          A dLLM-based VLM jointly interprets text, visual context, and structured
          reasoning traces. A Residual Query Adapter and lightweight Transformer
          connector expose generation-relevant hidden states to a pure single-stream
          DiT. Text-to-image prompts and editing instructions share this semantic
          pathway, while SigLIP-VQ features and the clean VAE latent provide
          complementary semantic and pixel-level reference signals for editing.
          This design brings multimodal understanding, text-to-image generation,
          and reference-preserving editing together in one checkpoint.

    \item \textbf{A stable, real-data-dominant progressive training pipeline.}
          We progress from $256^2$ image-only pre-training to
          aspect-ratio-bucketed mid-training, paired language alignment at $512^2$
          and $1024^2$, targeted refinement on text-rich and portrait data, and
          finally joint generation--editing training. The real-image share remains
          above $70\%$ throughout supervised fine-tuning. Although this
          real-data-dominant recipe can improve more slowly on early benchmarks than
          synthetic-heavy alternatives, it yields stronger realism at convergence
          and a higher long-horizon capability ceiling. Parameter-free RMSNorm
          throughout the DiT and the Muon optimizer further stabilize the full
          training pipeline.

    \item \textbf{Efficient 2--4-step inference with \lladaimageturbo.}
          Building on distribution-matching distillation \citep{yin2024dmd2} and
          self-adversarial flow training, TwinFlow~\citep{cheng2026twinflow}
          distills the original multi-step model into \lladaimageturbo. The
          resulting variants require only 2--4 sampling steps.

    \item \textbf{Open weights, code, and training recipes.}
          We release the \lladaimage and \lladaimageturbo checkpoints, training and
          inference code, and detailed recipes for data construction, progressive
          training, unified generation and editing, and TwinFlow distillation.
\end{itemize}

Taken together, these components yield two complementary deployment profiles
within a single model family. The original \lladaimage checkpoint serves as the
full multi-step model for high-fidelity generation and unified editing, whereas
\lladaimageturbo is its distilled 2--4-step deployment variant for
substantially lower inference cost. Evaluations on Qwen-Image-Bench,
LongText-Bench, CVTG-2K, and GEdit-Bench demonstrate broad visual quality and
prompt alignment, balanced Chinese--English text rendering, and competitive
instruction-guided editing without separate task-specific backbones. More
broadly, the results show that a capable visual creation system can be trained
with a moderate data budget dominated by image-only and real data, and then
efficiently adapted for deployment through few-step distillation. We release
four checkpoints in total: the standard Base and Turbo checkpoints, together
with an FP8 variant of each. We also release the training and inference code
and the progressive recipes to make this path reproducible.

\section{Model Design}
\label{sec:architecture}

\begin{figure}[!t]
    \centering
    \includegraphics[width=\linewidth]{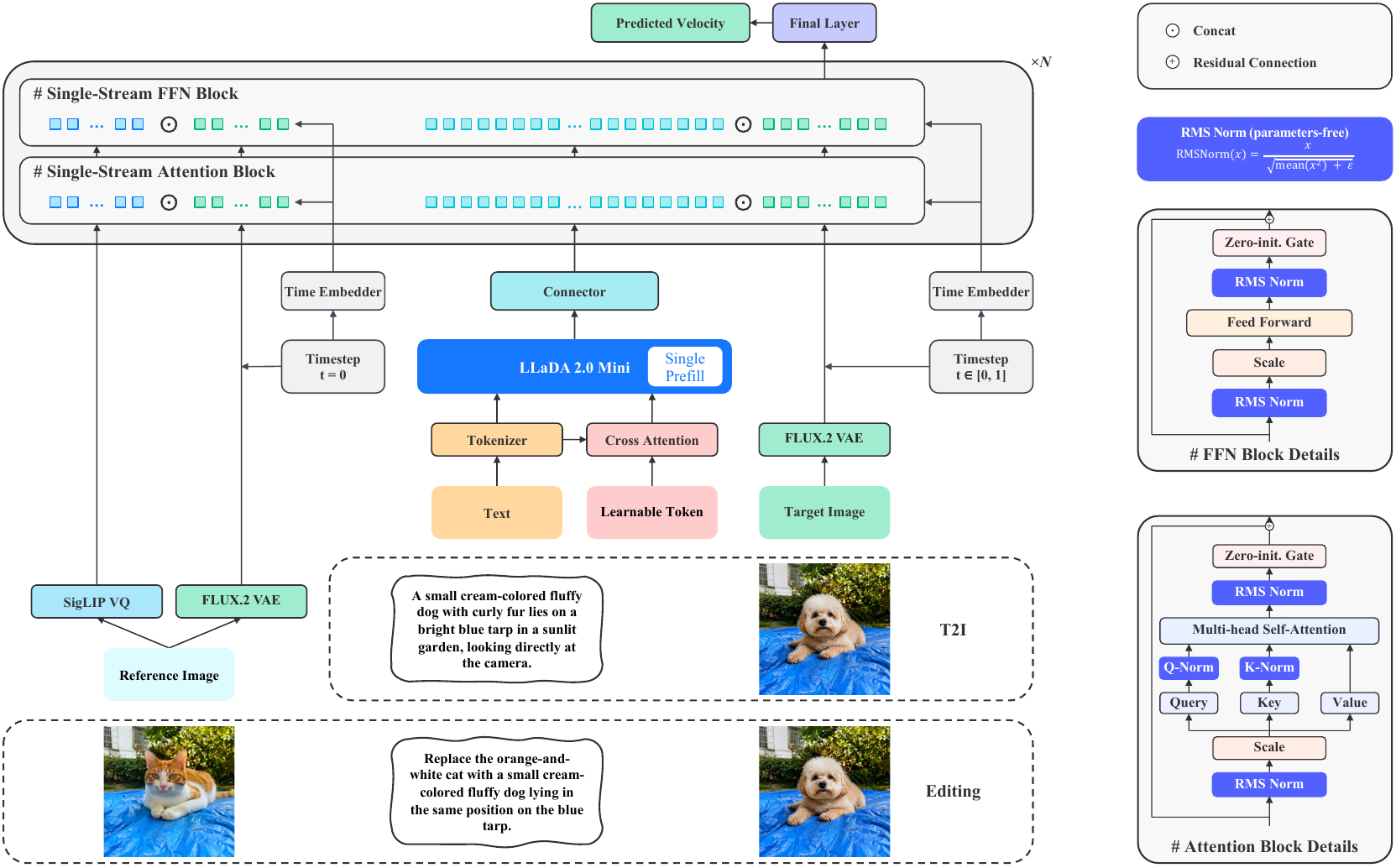}
    \caption{\textbf{Overview of the \lladaimage architecture.}
        Learnable query tokens aggregate information from the input tokens through
        cross-attention in the RQA. The resulting query representations are
        concatenated with the original inputs and encoded by LLaDA 2.0 Mini, after
        which the connector maps the hidden states into the DiT conditioning
        space. The FLUX.2 VAE supplies image latents, while editing additionally
        conditions the generator on SigLIP-VQ features and clean VAE latents from
        the reference image. All modal tokens are concatenated and processed by
        the single-stream DiT to predict the flow-matching velocity.}
    \label{fig:llada_image_architecture}
\end{figure}

As illustrated in \figref{fig:llada_image_architecture}, \lladaimage
consists of three principal components: a diffusion large language model
(dLLM)-based vision-language model (VLM) for multimodal understanding, a
connector that projects VLM representations into the generator's conditioning
space, and a diffusion transformer (DiT) for image synthesis~\citep{peebles2023dit}. This modular
architecture cleanly decouples high-level semantic interpretation from
pixel-space generation, assigning each to a dedicated component while using
the connector as an explicit bridge. Crucially, the unified architecture
seamlessly supports both text-to-image generation and image editing. For
editing, the framework reuses the shared text-conditioning pathway and DiT
backbone while engaging an additional reference-image pathway directly within
the DiT---critically bypassing the VLM entirely.

\subsection{dLLM-based Vision-Language Model}
\label{sec:model_vlm}

The understanding component is a VLM built upon the LLaDA 2.0 Mini dLLM
backbone, including its SigLIP-VQ vision encoder~\citep{ai2026llada2}. It
processes the text prompt or editing instruction to steer generation, while
also being capable of consuming visual tokens during image-only
self-conditioning. Unlike conventional text encoders that merely encode the
input prompt into static embeddings, the VLM provides a unified interface
capable of joint reasoning over multimodal contexts. Crucially, the reference
image used during editing is deliberately excluded from this comprehension
stage; instead, it is injected directly into the DiT via the dedicated pathway
detailed in \secref{sec:model_editing_pathway}.

\paragraph{Multimodal input construction.}
Let $\boldsymbol{t}$ denote the text tokenizer and embedding layers, $\boldsymbol{v}$ the SigLIP-VQ vision encoder, and $\boldsymbol{g}$ the VLM backbone. Given an input text sequence $\mathbf{s}_{\mathrm{text}}$ and an optional image $\mathbf{y}$, we obtain the corresponding text and image token representations as:
\begin{equation}
    \mathbf{c}_{\mathrm{text}} = \boldsymbol{t}(\mathbf{s}_{\mathrm{text}}),
    \qquad
    \mathbf{c}_{\mathrm{img}} = \boldsymbol{v}(\mathbf{y}).
    \label{eq:multimodal_tokenization}
\end{equation}
Depending on the task, the active token representations in
\eqref{eq:multimodal_tokenization} are concatenated in a predefined order to
form a unified sequence $\mathbf{c}$:
\begin{itemize}[leftmargin=*,noitemsep,topsep=0pt]
    \item \textbf{Text-to-Image Generation:} $\mathbf{c}$ contains solely the text condition $\mathbf{c}_{\mathrm{text}}$.
    \item \textbf{Image-Only Pre-training:} $\mathbf{c}$ comprises the auxiliary prompt alongside the sampled visual tokens $\mathbf{c}_{\mathrm{img}}$.
    \item \textbf{Image Editing:} $\mathbf{c}$ processes only the editing instruction, as the reference image completely bypasses the VLM.
\end{itemize}
By decoupling these pathways, the VLM remains a general-purpose multimodal understanding module. The bridge that adapts these representations for visual synthesis is the understanding-to-generation connector, described next.

\subsection{Understanding-to-Generation Connector}
\label{sec:model_connector}

The VLM and DiT operate in distinct representation spaces optimized for different objectives: the former emphasizes high-level semantic understanding, whereas the latter requires conditioning signals that directly guide visual denoising trajectories. To bridge this gap, we introduce an understanding-to-generation interface consisting of two generation-specific modules: a \textbf{Residual Query Adapter (RQA)} preceding the VLM prefill, and a \textbf{Transformer Connector} following it. Both modules function independently of the core VLM, serving specifically to extract and project VLM representations into the DiT conditioning space.

\paragraph{Residual query adaptation.}
The frozen VLM does not directly expose the fine-grained visual priors required
for high-fidelity image generation. To avoid the cost of fine-tuning the entire
VLM and preserve its pre-trained understanding capabilities, we adopt the
lightweight RQA introduced in IOMM~\citep{sun2026rethinking}. Let
$\mathbf{q}_{0}$ denote a set of learnable query tokens and
$\boldsymbol{q}_{\psi}$ represent the RQA parameterized by $\psi$. Prior to VLM
prefilling, these learnable queries cross-attend to the full multimodal input
sequence $\mathbf{c}$:
\begin{equation}
    \mathbf{q}_{\mathrm{res}} = \boldsymbol{q}_{\psi}(\mathbf{q}_{0}, \mathbf{c}).
    \label{eq:residual_query_adapter}
\end{equation}
Instead of replacing the original input, the residual queries produced by
\eqref{eq:residual_query_adapter} are appended to $\mathbf{c}$. This augmented
sequence is then processed by the frozen VLM backbone $\boldsymbol{g}$ in a
single prefill forward pass:
\begin{equation}
    \mathbf{h}_{\mathrm{vlm}} = \boldsymbol{g}\!\left(\operatorname{concat}\left(\mathbf{c}, \mathbf{q}_{\mathrm{res}}\right)\right).
    \label{eq:vlm_prefill}
\end{equation}
Together, \eqref{eq:residual_query_adapter} and \eqref{eq:vlm_prefill}
prompt the frozen VLM to extract and expose multimodal context that is
maximally beneficial for generative synthesis.

\paragraph{Condition-space alignment.}
The prefill hidden states $\mathbf{h}_{\mathrm{vlm}}$ remain within the VLM representation space, which typically differs from the DiT conditioning space in both feature dimension and geometric structure. We therefore process $\mathbf{h}_{\mathrm{vlm}}$ with a connector $\boldsymbol{c}_{\phi}$ composed of a shallow stack of Transformer blocks. This module projects the features and aligns them directly with the DiT condition space:
\begin{equation}
    \mathbf{h}_{\mathrm{cond}} = \boldsymbol{c}_{\phi}(\mathbf{h}_{\mathrm{vlm}}).
    \label{eq:connector_mapping}
\end{equation}
The connector mapping in \eqref{eq:connector_mapping} complements the RQA:
the former translates VLM features into a format optimized for DiT
consumption, while the latter elicits generation-centric information during
the VLM prefill. Together, they establish a unified conditioning pipeline
across text-to-image generation and image-only training. For image editing,
this shared pathway handles the instruction text, while an independent
reference-image stream feeds directly into the DiT.

\subsection{Single-stream Diffusion Transformer}
\label{sec:model_dit}

\begin{tcolorbox}[
        enhanced,
        sharp corners,
        boxrule=0pt,
        leftrule=2.5pt,
        colback=title_blue!4!white,
        colframe=title_blue,
        left=7pt,
        right=7pt,
        top=5pt,
        bottom=5pt,
        before skip=7pt,
        after skip=7pt
    ]
    \textbf{Recipe \#1.}
    We replace every normalization layer in the DiT with parameter-free
    RMSNorm~\citep{zhang2019rmsnorm}. This simple design choice substantially improves optimization
    stability over long training horizons.
\end{tcolorbox}

The generation component is a pure single-stream Transformer-based DiT. Given
a noised image latent $\mathbf{x}_t$, timestep $t$, and condition encoding
$\mathbf{h}_{\mathrm{cond}}$, we embed the image and condition tokens into a
shared sequence and process them jointly through the same stack of Transformer
blocks. In contrast to architectures that maintain separate condition and
image streams, every block in our DiT directly models interactions between
semantic conditions and evolving visual tokens through joint self-attention.
The predicted flow field is
\begin{equation}
    \widehat{\mathbf{v}}_t
    = \boldsymbol{F}_{\theta}
    \!\left(\mathbf{x}_t,t,\mathbf{h}_{\mathrm{cond}}\right)
    = \boldsymbol{F}_{\theta}
    \!\left(\mathbf{x}_t,t,
    \boldsymbol{c}_{\phi}\!\left(
        \boldsymbol{g}\!\left(
            \operatorname{concat}\!\left(
                \mathbf{c},
                \boldsymbol{q}_{\psi}(\mathbf{q}_{0},\mathbf{c})
                \right)
            \right)
        \right)\right),
    \label{eq:model_end_to_end}
\end{equation}
In \eqref{eq:model_end_to_end}, $\boldsymbol{F}_{\theta}$ denotes the DiT.
This single-stream formulation
provides a direct path for multimodal semantics to influence image synthesis
at every layer, while retaining a simple and homogeneous Transformer
backbone.

\paragraph{Reference-image conditioning for editing.}
\label{sec:model_editing_pathway}
As shown in the lower editing pathway of
\figref{fig:llada_image_architecture}, image editing introduces a
reference-image stream in parallel with, rather than inside, the common
VLM--connector pathway. Let
$\mathbf{y}_{\mathrm{ref}}$ denote the clean reference image. We first extract
its semantic representation with SigLIP-VQ, and then process the resulting
features using a DiT-specific reference branch $\boldsymbol{b}_{\omega}$:
\begin{equation}
    \mathbf{f}_{\mathrm{ref}}
    = \boldsymbol{v}(\mathbf{y}_{\mathrm{ref}}),
    \qquad
    \mathbf{h}_{\mathrm{ref}}
    = \boldsymbol{b}_{\omega}(\mathbf{f}_{\mathrm{ref}}).
    \label{eq:editing_semantic_condition}
\end{equation}
The branch $\boldsymbol{b}_{\omega}$ in
\eqref{eq:editing_semantic_condition} consists of a dedicated feature embedder
followed by two Transformer layers. It adapts the SigLIP-VQ representation to the
DiT token space without injecting the reference image into the VLM. The
resulting reference tokens are unified with the text condition produced by the
connector,
\begin{equation}
    \mathbf{h}_{\mathrm{joint}}
    = \operatorname{concat}\!\left(
    \mathbf{h}_{\mathrm{cond}},
    \mathbf{h}_{\mathrm{ref}}
    \right),
    \label{eq:editing_joint_condition}
\end{equation}
The joint tokens in \eqref{eq:editing_joint_condition} then participate with
the visual tokens in all subsequent single-stream Transformer layers of the
DiT.

Semantic features alone cannot fully preserve the low-level appearance of
regions that should remain unchanged. We therefore additionally encode the
clean reference image with the FLUX.2 VAE~\citep{blackforestlabs2026flux2} and concatenate its latent with the
noised target latent $\mathbf{x}_t$ before the DiT input embedder:
\begin{equation}
    \mathbf{x}_{\mathrm{ref}}
    = \operatorname{VAE}(\mathbf{y}_{\mathrm{ref}}),
    \qquad
    \mathbf{x}^{\mathrm{edit}}_t
    = \operatorname{concat}\!\left(
    \mathbf{x}_t,
    \mathbf{x}_{\mathrm{ref}}
    \right),
    \qquad
    \widehat{\mathbf{v}}^{\mathrm{edit}}_t
    = \boldsymbol{F}_{\theta}\!\left(
    \mathbf{x}^{\mathrm{edit}}_t,t,\mathbf{h}_{\mathrm{joint}}
    \right).
    \label{eq:editing_pixel_condition}
\end{equation}
The semantic pathway in \eqref{eq:editing_semantic_condition} and the pixel
pathway in \eqref{eq:editing_pixel_condition} are complementary: the former
supplies high-level guidance about the reference content, whereas the clean
VAE latent provides native pixel-level evidence for regions that do not need
to be modified. Their combination improves instruction following while
substantially strengthening identity, structure, texture, and background
consistency with the reference image, all without introducing a separate
editing backbone.

\section{Data Collection and Preparation}
\label{sec:data}

We collect a large-scale image corpus from web sources and curated image
collections. Most samples are real images, while synthetic images form a
smaller part of the supervised fine-tuning (SFT) data and are used mainly for
text rendering. The corpus provides image-only data for pre-training and
mid-training, and paired image--text data for SFT.

\begin{tcolorbox}[
    enhanced,
    sharp corners,
    boxrule=0pt,
    leftrule=2.5pt,
    colback=title_blue!4!white,
    colframe=title_blue,
    left=7pt,
    right=7pt,
    top=5pt,
    bottom=5pt,
    before skip=7pt,
    after skip=7pt
  ]
  \textbf{Recipe \#2.}
  Real image-only data can improve final generation quality and support
  visual-prior learning at scale without paired captions.
\end{tcolorbox}

\subsection{Data Composition}

\begin{figure}[t]
  \centering
  \includegraphics[width=\linewidth]{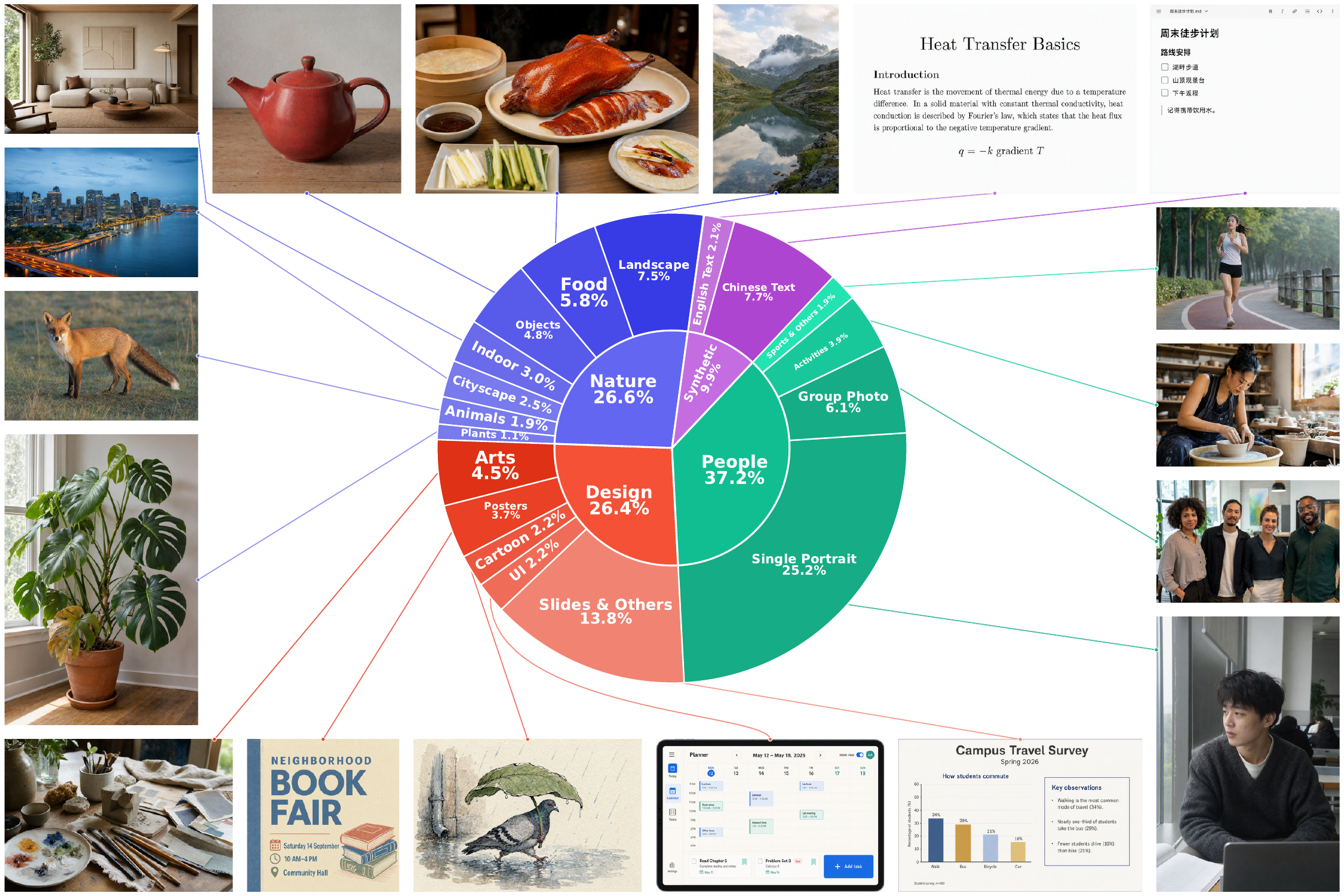}
  \caption{Estimated SFT-weighted content distribution, with examples
    illustrating each content group.}
  \label{fig:sft-weighted-data-distribution}
\end{figure}

We organize the SFT data into four broad content groups: nature, design,
people, and synthetic content. Of the 220M samples used for generation
training, $98\%$ are real images, while image-only samples account for more
than $90\%$ of the total. Pre-training and mid-training use only real images,
while the real-image share of the paired image--text data used for SFT remains
above $70\%$. Using real image-only data during pre-training
and mid-training provides a scalable way to expose the model to diverse visual
content.
\figref{fig:sft-weighted-data-distribution} summarizes the SFT-weighted data
composition.

\subsection{Data Processing}

We filter the collected images in three stages: metadata, aesthetics, and
quality filtering. Metadata filtering keeps only valid images with a total
pixel count greater than $1024^2$ and a file-size-to-pixel ratio of at least
0.15 bytes per pixel. Aesthetics filtering removes images with
ArtiMuse~\citep{cao2025artimusefinegrainedimageaesthetics} scores below 60,
and quality filtering removes images with DeQA-Score~\citep{deqa_score} below
4.0. After filtering,
pre-training uses random square crops resized to $256^2$, whereas
mid-training and subsequent SFT stages use the aspect-ratio buckets described
in \secref{sec:midtraining}.

We use both Qwen3.6-35B-A3B~\citep{qwen2026qwen36} and
Qwen3-VL-235B-A22B-Instruct~\citep{bai2025qwen3vl} to caption the
filtered images from image content alone. The captioning prompt requires a
faithful and complete description of visible subjects, attributes, actions,
spatial relations, scenes, and visual styles. The models also transcribe
visible text while preserving its original language and spelling.

We keep the resulting image--text pairs only when image quality and caption
agreement are sufficient. Qwen3.6-35B-A3B compares each caption with its source
image and removes captions that hallucinate objects, attributes, relations, or
text, or that inaccurately transcribe visible text. We further discard
malformed outputs, refusal responses, degenerate repetitions, private
information, and watermark signals. The remaining captions provide image--text
supervision for SFT. Additional details are provided in
\appref{app:caption-generation-filtering}.

For image-editing data, we apply image-quality filtering similar to that used
for the SFT data, followed by an edit-instruction consistency check. We keep
only source--target pairs whose instructions describe the visible changes in
the correct direction and contain no unsupported details.

\section{Training Recipe}
\label{sec:model-training}

The overall training and release path is summarized in
\figref{fig:training-release-path}. Inspired by recent
studies~\cite{ai2026llada2, han2026towards} demonstrating that strong
comprehension capabilities can facilitate generation, we first conduct
Chain-of-Thought (CoT) supervised fine-tuning to enhance the backbone's
performance. The details of our training recipe are provided below.
\tabref{tab:understanding-training-configurations} and
\tabref{tab:generation-training-configurations} summarize the independent
training pipelines for understanding and generation, respectively. Across all
generation stages, we use the Muon optimizer~\citep{liu2025muon}.
\begingroup
\newcommand{\configtodo}{\textcolor{red}{[TODO]}}

\begin{table}[!htbp]
    \centering
    \small
    \setlength{\tabcolsep}{6pt}
    \renewcommand{\arraystretch}{1.25}
    \caption{Detailed configuration for understanding training. CoT SFT denotes
        CoT Supervised Fine-tuning.}
    \label{tab:understanding-training-configurations}
    \begin{tabular}{lc}
        \toprule
        \textbf{Configuration} & \textbf{CoT SFT}                                 \\
        \midrule
        \multicolumn{2}{l}{\textbf{Training Process}}                             \\
        \midrule
        Training Samples       & $\sim$2.6M packed seq. ($16{,}384$ tok.)         \\
        Training Epochs        & $4$                                              \\
        Resolution             & $512^2$                                          \\
        Global Batch Size      & $512$                                            \\
        \midrule
        \multicolumn{2}{l}{\textbf{Data Distribution}}                            \\
        \midrule
        Supervision            & Masked answer tokens (block diffusion, $b{=}32$) \\
        Data Composition       & Gen : Und : Text $=9{:}9{:}2$                    \\
        \midrule
        \multicolumn{2}{l}{\textbf{Hyperparameters}}                              \\
        \midrule
        Optimizer              & AdamW ($\beta_1{=}0.9$, $\beta_2{=}0.95$)        \\
        Learning Rate          & $1\times10^{-5}\rightarrow1\times10^{-6}$        \\
        LR Schedule            & Cosine decay                                     \\
        LR Warmup              & Linear, $1\%$                                    \\
        Weight Decay           & 0.1                                              \\
        Grad. Norm Clip        & 1.0                                              \\
        \bottomrule
    \end{tabular}
\end{table}

\endgroup

\subsection{CoT Supervised Fine-tuning}
\label{sec:cot_sft}

\paragraph{Data Process and Construction.}
We tokenize all corpora offline and greedily bin-pack them into sequences of
exactly $L=16{,}384$ tokens, and we VQ-encode each image once at a $512^2$ target
resolution under variable-aspect-ratio cropping. A block-diagonal attention mask
built from the cumulative lengths keeps packed samples mutually invisible, so
packing raises utilization without cross-sample leakage. We mix generation,
understanding and text-only data at a $9{:}9{:}2$ ratio.

\paragraph{Training Objective.}
We keep the block-diffusion formulation~\citep{ai2026llada2} with a block size
of $b=32$. For each
supervised region we draw a masking ratio $\rho=\cos(r\pi/2)$ with
$r\sim\mathcal{U}(0,1)$ and replace that fraction of the response tokens by the
mask token. The system prompt, the question and all visual tokens are never masked
and never supervised, so the loss is a cross-entropy over the masked positions of
the assistant turn: the \texttt{<think>} trace together with the answer. One mask
covers only part of the response, so we consume each batch twice, under the
sampled mask and under its complement, as two optimizer steps. Every response
token is then supervised within the pair, while each forward pass still reads a
partially clean context. We normalize the loss per block: within each block we
divide the loss of every supervised token by the number of supervised tokens in
that block, then average over the blocks that carry supervision and over the
packed sequences in the batch.

\paragraph{Optimization.}
We train with AdamW~\citep{loshchilov2019adamw} under FSDP2 full sharding with mixed precision and gradient
checkpointing. The learning rate warms up linearly and then decays by cosine to
its floor; since the complementary pass takes its own step, the schedule spans
twice the number of data iterations.

\subsection{Image-only Pre-training}
\label{sec:pretraining}

The most compute-intensive stage of visual generation training conventionally
relies on large-scale image--text pairs. Following Image-Only Training for UMMs
(IOMM)~\citep{sun2026rethinking}, we instead bootstrap the visual generative
component from unlabeled images alone. The key observation is that an image
already contains the high-level semantics required to supervise its generation:
a frozen vision-language model (VLM) can extract these semantics and
use them as a self-conditioning signal. This stage therefore learns a strong
visual prior before any explicit image--text alignment, while keeping the
pre-trained VLM intact.

\begin{tcolorbox}[
    enhanced,
    sharp corners,
    boxrule=0pt,
    leftrule=2.5pt,
    colback=title_blue!4!white,
    colframe=title_blue,
    left=7pt,
    right=7pt,
    top=5pt,
    bottom=5pt,
    before skip=7pt,
    after skip=7pt
  ]
  \textbf{Recipe \#3.}
  Large-scale image--text pairs are not required to pre-train \lladaimage: images
  alone provide sufficient supervision for learning a strong visual generative
  prior.
\end{tcolorbox}

\paragraph{Resolution-aware image sampling.}
We conduct image-only pre-training at a resolution of $256^2$. When
training with conventional image--text pairs, the caption usually describes
the complete image; consequently, the entire image must be resized to the
training resolution to preserve image--caption alignment. For a
high-resolution source image, aggressive downsampling to $256^2$ can
remove fine-grained content referenced by the caption, introduce visual
distortions, and ultimately create a mismatch between the image and its text
condition. Image-only pre-training removes this constraint because its
condition is derived from the sampled image itself. We randomly crop a region
from each source image whose native crop resolution is either $256^2$
or moderately larger, and then resize the crop to $256^2$ with only a
small downsampling ratio. The VLM extracts the semantic condition directly
from this processed crop, ensuring that the condition supplied to the
diffusion transformer remains faithful to the actual training target. This
sampling strategy exposes the model to substantially more local detail while
avoiding information loss and distortion caused by aggressively compressing
an entire high-resolution image.

\paragraph{Masked self-conditioning.}
Given a training image $\mathbf{y}$, let $\mathbf{x}$ denote its clean latent
in the generator's representation space. Following the notation introduced in
\secref{sec:model_vlm}, $\boldsymbol{v}$ denotes the frozen SigLIP-VQ
vision encoder and $\boldsymbol{g}$ denotes the frozen VLM. We first encode the
image into a sequence of patch features
$\mathbf{c}_{\mathrm{img}}=\boldsymbol{v}(\mathbf{y})\in\mathbb{R}^{N\times D}$,
where $N$ is the number of image tokens and $D$ is their feature dimension.
These features are paired with
$\mathbf{c}_{\mathrm{aux}}=\boldsymbol{t}(\mathbf{s}_{\mathrm{aux}})$, where
$\mathbf{s}_{\mathrm{aux}}$ is a fixed auxiliary prompt such as
\texttt{Generate an image identical to the reference image.} Directly
conditioning on all image features would reveal an almost complete description
of the reconstruction target and encourage a trivial identity mapping. We
therefore define an image-token masking ratio $\rho_{\mathrm{img}}$ and sample
a binary token mask $\mathbf{m}\in\{0,1\}^{N}$ with keep probability
$1-\rho_{\mathrm{img}}$:
\begin{equation}
  \widetilde{\mathbf{c}}_{\mathrm{img}}
  = \mathbf{c}_{\mathrm{img}} \odot \mathbf{m},
  \qquad
  \mathbf{c}
  = \operatorname{concat}\!\left(
  \mathbf{c}_{\mathrm{aux}},
  \widetilde{\mathbf{c}}_{\mathrm{img}}
  \right),
  \label{eq:image_only_condition}
\end{equation}
where $\mathbf{m}$ is broadcast along the feature dimension. The construction
in \eqref{eq:image_only_condition} converts pre-training from dense
reconstruction into a sparse-to-dense prediction task:
the generator must recover missing content from the visible patches and thus
learn contextual and compositional visual structure. This stage changes only
how the multimodal input sequence $\mathbf{c}$ is constructed; its subsequent
transformation follows the common model architecture described in
\secref{sec:model_vlm} and \secref{sec:model_connector}. Specifically, the RQA,
frozen VLM, and connector apply \eqref{eq:residual_query_adapter},
\eqref{eq:vlm_prefill}, and \eqref{eq:connector_mapping}, respectively, to
produce $\mathbf{h}_{\mathrm{cond}}$, which conditions the DiT through
\eqref{eq:model_end_to_end}. Thus, image-only training reuses exactly the same
RQA--VLM--connector pathway as downstream conditional generation.

\paragraph{Flow-matching objective.}
We optimize the visual generator with flow matching~\citep{lipman2023flowmatching}. For Gaussian noise
$\mathbf{z}\sim\mathcal{N}(\mathbf{0},\mathbf{I})$ and
$t\sim\mathcal{U}(0,1)$, we construct the interpolation
$\mathbf{x}_t=(1-t)\mathbf{x}+t\mathbf{z}$. The generator
$\boldsymbol{F}_{\theta}$ predicts the corresponding constant velocity field,
and is trained with
\begin{equation}
  \mathcal{L}_{\mathrm{FM}}(\theta,\phi,\psi)
  = \mathbb{E}_{\mathbf{y},\mathbf{x},\mathbf{z},t,\mathbf{m}}
  \left[
    \left\|
    \boldsymbol{F}_{\theta}(\mathbf{x}_t,t,\mathbf{h}_{\mathrm{cond}})
    -(\mathbf{z}-\mathbf{x})
    \right\|_2^2
    \right].
  \label{eq:image_only_flow_matching}
\end{equation}
When optimizing \eqref{eq:image_only_flow_matching}, only the visual generator
$\boldsymbol{F}_{\theta}$, residual query adapter
$\boldsymbol{q}_{\psi}$, and connector $\boldsymbol{c}_{\phi}$ are updated;
$\boldsymbol{v}$ and $\boldsymbol{g}$ remain frozen throughout this stage.
Combining masked reconstruction with
parameter-efficient self-conditioning allows the model to absorb large-scale
visual knowledge from inexpensive image-only data and provides a robust
initialization for the subsequent alignment stages.

\begingroup

\begin{table}[!htbp]
    \centering
    \footnotesize
    \setlength{\tabcolsep}{1.5pt}
    \renewcommand{\arraystretch}{1.18}
    \newcolumntype{N}{>{\hsize=0.90\hsize\linewidth=\hsize\centering\arraybackslash}X}
    \newcolumntype{R}{>{\hsize=0.75\hsize\linewidth=\hsize\centering\arraybackslash}X}
    \newcolumntype{J}{>{\hsize=1.15\hsize\linewidth=\hsize\centering\arraybackslash}X}
    \newcolumntype{Z}{>{\hsize=1.50\hsize\linewidth=\hsize\centering\arraybackslash}X}
    \caption{Detailed configurations for generation training. PT, MT, and SFT
        denote image-only pre-training, image-only mid-training, and supervised
        fine-tuning, respectively. SFT comprises a $512^2$ text-alignment stage
        and a $512^2\!\rightarrow\!1024^2$ resolution-transition stage, followed by
        refinement and editing; the final column reports few-step distillation.
        Centered entries apply to all stages they span. The 220M samples are
        counted cumulatively across the image-generation pipeline and refer
        exclusively to generation training.}
    \label{tab:generation-training-configurations}
    \begin{tabularx}{\linewidth}{
        >{\raggedright\arraybackslash}p{0.15\linewidth}
        *{2}{N}|*{2}{N}RJ|Z}
        \toprule
        \multirow{2}{*}{\textbf{Configuration}} & \multirow{2}{*}{\textbf{PT}}    & \multirow{2}{*}{\textbf{MT}} & \multicolumn{4}{c|}{\textbf{SFT}}    & \multirow{2}{*}{\textbf{Distillation}}                                             \\
        \cmidrule(lr){4-7}
        {}                                      & {}                              & {}                           & \textbf{$512^2$ Align.} & \mbox{\textbf{$512^2\!\rightarrow\!1024^2$}} & \textbf{Refine.} & \textbf{Editing} & {}  \\
        \midrule
        \multicolumn{8}{l}{\textbf{Training Process}}                                                                                                                                                                                        \\
        \midrule
        Training Samples                        & \multicolumn{7}{c}{220M total}                                                                                                                                                             \\
        Resolution                              & $256^2$                         & \makecell{$512^2$\\(bucketed)} & \makecell{$512^2$                                                                                                         \\(bucketed)} & \multicolumn{3}{|c|}{$1024^2$ (bucketed)} & \makecell{$1024^2$\\(bucketed)} \\
        Global Batch Size                       & 24,576                          & 6,400                        & 4,608                                & 2,048                                  & 2,880            & 2,688            & 256 \\
        Sampling Steps                          & \multicolumn{2}{c|}{--}         & \multicolumn{4}{c|}{50}      & 2--4                                                                                                                      \\
        \midrule
        \multicolumn{8}{l}{\textbf{Data Distribution}}                                                                                                                                                                                       \\
        \midrule
        Supervision                             & \multicolumn{2}{c|}{Image-only} & \multicolumn{4}{c|}{Image--Text Pairs} & Image--Text Pairs                                                                                                           \\
        Data Ratio                              & \multicolumn{2}{c|}{Image-only} & \multicolumn{3}{c|}{T2I only} & T2I:I2I $=1{:}1$ & T2I:I2I $=1{:}1$                                                                                 \\
        \midrule
        \multicolumn{8}{l}{\textbf{Hyperparameters}}                                                                                                                                                                                         \\
        \midrule
        Optimizer                               & \multicolumn{2}{c|}{Muon}       & \multicolumn{4}{c|}{Muon}    & Muon                                                                                                                      \\
        Learning Rate                           & $4\times10^{-4}$                & $2\times10^{-4}$             & \multicolumn{2}{c|}{$5\times10^{-5}$} & \multicolumn{2}{c|}{$3\times10^{-5}$} & $5\times10^{-6}$                           \\
        Weight Decay                            & \multicolumn{2}{c|}{0.0}        & \multicolumn{4}{c|}{0.0}     & 0.0                                                                                                                       \\
        Grad. Norm Clip                         & \multicolumn{2}{c|}{1.0}        & \multicolumn{4}{c|}{1.0}     & 1.0                                                                                                                       \\
        EMA Ratio                               & \multicolumn{2}{c|}{--}         & --                           & \multicolumn{3}{|c|}{0.9995}         & 0.995                                                                              \\
        \bottomrule
    \end{tabularx}
\end{table}

\endgroup

\FloatBarrier

\subsection{Image-only Mid-training}
\label{sec:midtraining}

We continue to use the same image-only training formulation as in
pre-training, including image-derived self-conditioning, image-token masking,
and the flow-matching objective, while increasing the nominal pixel budget
from $256^2$ to $512^2$. The condition therefore remains inherently aligned
with the processed training image, allowing us to expose the DiT to finer
spatial structure without introducing caption mismatch. We also begin using
the aspect-ratio-bucketed sampling strategy detailed below, so $512^2$ denotes
the target pixel budget rather than requiring every image to be resized to a
fixed square.

This stage primarily serves as a bridge between image-only pre-training and
paired text-to-image alignment. As discussed above, captions generally
describe the complete source image, requiring the full image to be
downsampled when image--text pairs are used. Excessive compression can remove
fine-grained details referenced by the caption and thereby weaken
image--text compatibility. More importantly, directly switching from
$256^2$ image-only training to high-resolution image--text pair training
would change both the spatial distribution and the source of supervision at
the same time, producing an unnecessarily abrupt shift in the training
regime. Image-only mid-training decouples these two changes: the model first
adapts to the $512^2$ pixel budget under the familiar image-derived condition,
and only then transitions to textual supervision during SFT. This staged
schedule provides a smoother transition in both resolution and conditioning
modality while preserving the visual prior learned during pre-training.

\paragraph{Aspect-ratio-bucketed variable-resolution training.}
We introduce variable-resolution training during image-only mid-training and
retain it throughout all subsequent SFT stages. The data loader maintains a
predefined collection of resolution buckets with different aspect ratios but
approximately the same pixel budget. When the dataset yields an image, the
data loader assigns it to the bucket whose aspect ratio is closest to that of
the source image. It then downsamples the image while preserving its original
aspect ratio. Because the resized image must cover the selected bucket, one
side can remain slightly larger than the target; we simply crop this small
overflow. Since the selected bucket already closely matches the source aspect
ratio, only a few boundary pixels are typically removed and the semantic
content is essentially unaffected. The detailed assignment rule, preprocessing
procedure, and complete bucket configurations are provided in
\appref{app:aspect-ratio-buckets}.

This strategy offers three practical benefits. First, although different data
parallel ranks may receive different spatial shapes, the nearly constant
per-image pixel budget keeps their memory consumption and computation closely
matched, avoiding rank-specific memory peaks and stragglers. Second, preserving
the source aspect ratio prevents geometric distortion from forcibly resizing
every image to a square, while the minimal crop avoids the substantial content
loss caused by a square center crop. Finally, exposing the model to multiple
spatial shapes directly equips it to generate images at diverse aspect ratios.

\subsection{Supervised Fine-tuning}
\label{sec:posttraining}

\paragraph{Logit-normal timestep sampling.}
We adopt logit-normal timestep sampling~\citep{esser2024scaling} instead of a
uniform distribution, which allocates the same number of training examples to
all noise levels, although their importance during inference is not uniform.
Under our interpolation convention
$\mathbf{x}_t=(1-t)\mathbf{x}+t\mathbf{z}$, larger $t$ corresponds to a
higher-noise state. Predictions in this region are more sensitive and less
stable: the image structure has not yet been established, and errors made
early in the denoising trajectory can propagate through all subsequent steps.
At small $t$, by contrast, most visual content has already stabilized and the
remaining denoising correction is comparatively limited. We therefore devote
a larger fraction of the SFT optimization budget to high-noise states instead
of sampling the high- and low-noise regions equally.

Specifically, we draw one standard Gaussian variable for each training sample
and transform it with a sigmoid function,
\begin{equation}
  \epsilon \sim \mathcal{N}(0,1),
  \qquad
  t = \operatorname{sigmoid}\!\left(
  P_{\mathrm{mean}} + P_{\mathrm{std}}\epsilon
  \right),
  \label{eq:logit_normal_timestep}
\end{equation}
In \eqref{eq:logit_normal_timestep}, $P_{\mathrm{mean}}$ controls the center of
the distribution in logit space and $P_{\mathrm{std}}$ controls its spread. We
set $P_{\mathrm{mean}}=0.8$ and $P_{\mathrm{std}}=0.8$ to shift more probability
mass toward larger $t$ while retaining support across
the complete denoising trajectory. This logit-normal sampling strategy
concentrates supervision on the more challenging high-noise regime, improving
the stability of the early denoising process without repeatedly oversampling
low-noise states that are already comparatively well resolved.

\subsubsection{Text-to-Image Training}
\label{sec:t2i_sft}

The preceding image-only pre-training and mid-training stages equip the
diffusion transformer (DiT) with a strong visual prior and basic generation
capability without requiring paired captions. However, the final text-to-image
system must generate images from textual instructions at inference time. We
therefore introduce image--text pairs during supervised fine-tuning to align
the semantic representations extracted from text with the visual generation
space learned in the image-only stages. This transition brings the training
condition closer to the model's actual usage while preserving the generative
prior acquired from substantially larger image-only corpora.

\begin{tcolorbox}[
    enhanced,
    sharp corners,
    boxrule=0pt,
    leftrule=2.5pt,
    colback=title_blue!4!white,
    colframe=title_blue,
    left=7pt,
    right=7pt,
    top=5pt,
    bottom=5pt,
    before skip=7pt,
    after skip=7pt
  ]
  \textbf{Recipe \#4.}
  Compared with training predominantly on synthetic data, using a large
  proportion of real images leads to slower early convergence on standard
  benchmarks, but substantially improves the realism of the model's final
  generated images after sufficient training.
\end{tcolorbox}

\paragraph{$512^2$-pixel alignment.}
We begin with an alignment stage using aspect-ratio buckets containing
approximately $512^2$ pixels per image. Here, $512^2$ is only the square
bucket rather than a fixed training shape; non-square images are assigned to
buckets with comparable pixel counts while retaining their aspect ratios.
Because image-only mid-training has already adapted the model to this pixel
budget, the resolution distribution remains approximately fixed while the
supervision changes from image-derived conditions to paired text. The primary
purpose of this stage is therefore not to relearn image synthesis from scratch,
but to associate textual concepts, attributes, and spatial relations with the
corresponding visual content while preserving the well-formed visual prior of
the DiT.

\paragraph{Scaling to $1024^2$ pixels.}
After establishing image--text alignment with the $512^2$-pixel buckets, we
continue training with a second set of aspect-ratio buckets containing
approximately $1024^2$ pixels per image. The progressive pixel-budget schedule
avoids introducing high-resolution generation and text alignment
simultaneously, yielding a smoother transition to the final operating
resolution. At this stage, the model learns to preserve semantic alignment
while producing finer textures, sharper boundaries, and more spatial detail.
The exact bucket resolutions used at both pixel budgets are listed in
\appref{app:aspect-ratio-buckets}.

\paragraph{Targeted refinement and data composition.}
In the refinement stage, we increase the proportions of text-intensive and
portrait-centric examples to strengthen two particularly demanding
capabilities: accurate text rendering and high-fidelity human generation. At
the same time, we deliberately limit the amount of synthetic data throughout
all SFT stages, with the real-image share consistently remaining above $70\%$
of the training mixture. Although this real-data-dominant recipe can
converge more slowly than a synthetic-heavy mixture early in training, we find
that its advantage becomes increasingly apparent over longer training: the
generated images exhibit substantially stronger realism and the model reaches
a higher overall capability ceiling.

\subsubsection{Image Editing Training}
\label{sec:editing_sft}

Many open-source systems provide text-to-image generation and image editing
through separate model variants~\citep{wu2025qwen,qwen2025imageedit2511,cai2025z,team2025longcat}.
In contrast, we aim to support both capabilities within a single unified model,
avoiding task-specific checkpoints and allowing generation and editing to
benefit from a shared visual prior.

\paragraph{Mixed-task continued training.}
During the editing stage, we jointly train on text-to-image (T2I) and
image-to-image editing (I2I) examples with a $1{:}1$ sampling ratio. Editing
data alone has a narrower distribution than the large-scale T2I corpus, and
continued training exclusively on it can cause the model to overfit the
editing objective, weaken its open-ended generation quality, and forget part
of the visual prior acquired in earlier stages. Mixing T2I examples into every
stage of editing training acts as capability replay: it continuously
regularizes the model toward broad semantic coverage, visual quality, and
text-to-image alignment. At the same time, the I2I examples teach precise edit
instruction following and reference-content preservation. The balanced
mixture therefore reduces task interference and allows one checkpoint to
acquire editing ability without sacrificing its original text-to-image
capability. More importantly, the two tasks share the same generative
backbone, so improvements in visual priors and instruction understanding can
transfer between generation and editing instead of being isolated in two
model branches.

\paragraph{Reference-conditioned editing.}
The editing data flow augments the standard T2I path with a parallel reference
image stream. The editing instruction follows the common
RQA--VLM--connector pathway, whereas the reference image does not enter the
VLM. Its semantic features are extracted by SigLIP-VQ and processed by a
DiT-specific embedder followed by two Transformer layers. These reference
tokens are then unified with the other condition and visual tokens before the
main single-stream DiT layers. In parallel, the clean reference image is
encoded into a VAE latent and concatenated with the noised target latent
$\mathbf{x}_t$. This second signal directly exposes native pixel information
from regions that should remain unchanged, improving structural and visual
consistency between the edited result and its reference. The complete model
pathway is detailed in \secref{sec:model_editing_pathway}; its semantic, joint,
and pixel-level conditions are defined in
\eqref{eq:editing_semantic_condition}, \eqref{eq:editing_joint_condition}, and
\eqref{eq:editing_pixel_condition}, respectively.

\subsubsection{Checkpoint Merging}
\label{sec:checkpoint_merging}

Although our data loader controls the sampling ratios among different data
sources at the global level, the empirical data composition seen by the model
can still vary across individual training steps. Such local variation induces
small fluctuations in the learned parameters. We observe the same phenomenon
when monitoring generation benchmarks throughout training: once optimization
approaches a performance plateau, scores from nearby checkpoints continue to
oscillate, even though their overall capabilities are comparable. Selecting a
single checkpoint may therefore retain step-specific bias and make the final
result unnecessarily sensitive to short-term sampling noise.

Motivated by weight averaging~\citep{izmailov2018swa}, we apply checkpoint
merging to the converged checkpoints. Aggregating model parameters from multiple nearby training states
suppresses transient, step-dependent deviations while preserving the shared
capabilities learned across them. In practice, this substantially smooths the
variation observed among individual checkpoints and produces a more stable and
robust final model across different evaluation benchmarks.

\subsection{Few-step Distillation}
\label{sec:distill}

\subsubsection{TwinFlow for Few-step Distillation}
We distill the multi-step \lladaimage into \lladaimageturbo, a few-step
generator. We use TwinFlow~\citep{cheng2026twinflow}, which builds on
distribution-matching distillation~\citep{yin2024dmd2,liu2026decoupleddmd}.
DMD2 minimizes
the reverse KL divergence between the student distribution and the target
distribution using a frozen real-score model and an online fake-score model.
The latter must track the non-stationary distribution produced by the student,
so the fake-score and generator objectives are optimized alternately, with the
fake-score estimator assigned the faster time scale. Following TwinFlow's
signed-time construction, we represent both roles with
one shared DiT backbone instead of introducing a separate fake-score network.
Positive times $+t\in[0,1]$ are reserved for the generator update, whereas
the corresponding negative times $-t\in[-1,0]$ train the fake-score estimator.
The sign is a role indicator: $+t$ and $-t$ denote different learning roles.

Let $\boldsymbol{F}_{\mathrm{base}}$ be a frozen copy of the original
multi-step model and $\boldsymbol{F}_{\theta}$ the model being distilled.
Starting from $\mathbf{z}\sim\mathcal{N}(\mathbf{0},\mathbf{I})$, we derive
the objective using the direct one-step mapping from $\mathbf{z}$ to a clean
sample:
\begin{equation}
  \widehat{\mathbf{x}}
  =\mathbf{z}-\boldsymbol{F}_{\theta}
  (\mathbf{z},+1,\mathbf{h}_{\mathrm{cond}}).
  \label{eq:distill_fake_sample}
\end{equation}
Under the flow-matching convention used in \eqref{eq:distill_fake_sample},
$\boldsymbol{F}_{\theta}$ predicts
$\mathbf{z}-\mathbf{x}$, so this update directly maps noise to data. To learn
the fake score, we detach the sample $\widehat{\mathbf{x}}$ from
\eqref{eq:distill_fake_sample} and sample
$t\sim\mathcal{U}(0,1)$ and fresh noise
$\mathbf{z}'\sim\mathcal{N}(\mathbf{0},\mathbf{I})$ to form
\begin{equation}
  \widetilde{\mathbf{x}}_t
  =(1-t)\operatorname{sg}(\widehat{\mathbf{x}})
  +t\mathbf{z}'.
  \label{eq:distill_fake_noising}
\end{equation}
Using the noised fake sample in \eqref{eq:distill_fake_noising}, the
negative-time branch is optimized with
\begin{equation}
  \mathcal{L}_{\mathrm{fake}}(\theta)
  =\mathbb{E}\!\left[
    \left\|
    \boldsymbol{F}_{\theta}
    (\widetilde{\mathbf{x}}_t,-t,\mathbf{h}_{\mathrm{cond}})
    -\left(
    \mathbf{z}'
    -\operatorname{sg}(\widehat{\mathbf{x}})
    \right)
    \right\|_2^2
    \right],
  \label{eq:distill_fake_score_loss}
\end{equation}
In \eqref{eq:distill_fake_score_loss}, $\operatorname{sg}(\cdot)$ denotes
stop-gradient. This loss fits the
$-t$ branch to the evolving student distribution, with gradients flowing only
through the negative-time prediction.

For distribution matching, we retain gradients through the sample in
\eqref{eq:distill_fake_sample} and re-noise it as
\begin{equation}
  \widehat{\mathbf{x}}_t
  =(1-t)\widehat{\mathbf{x}}+t\mathbf{z}'.
  \label{eq:distill_dmd_noising}
\end{equation}
A velocity prediction on the noised sample in \eqref{eq:distill_dmd_noising}
yields the score
$\mathbf{s}_t=-[\widehat{\mathbf{x}}_t+(1-t)\widehat{\mathbf{v}}_t]/t$.
Both models are evaluated on $\widehat{\mathbf{x}}_t$ with
$\mathbf{h}_{\mathrm{cond}}$: applying this conversion to the frozen
$\boldsymbol{F}_{\mathrm{base}}$ at $+t$ gives $\mathbf{s}_{\mathrm{real}}$,
whereas applying it to $\boldsymbol{F}_{\theta}$ at $-t$ gives
$\mathbf{s}_{\mathrm{fake}}$. No separate auxiliary score network is
introduced. The DMD objective is
\begin{equation}
  \mathcal{L}_{\mathrm{DMD}}(\theta)
  =\mathbb{E}\!\left[
    w(t)
    \left\langle
    \operatorname{sg}\!\left(
    \mathbf{s}_{\mathrm{fake}}
    -\mathbf{s}_{\mathrm{real}}
    \right),
    \widehat{\mathbf{x}}_t
    \right\rangle
    \right],
  \label{eq:distill_dmd_loss}
\end{equation}
In \eqref{eq:distill_dmd_loss}, $w(t)$ is the timestep weight. With both scores
detached, \eqref{eq:distill_dmd_loss} updates only the positive-time generator,
whereas \eqref{eq:distill_fake_score_loss} updates the negative-time fake-score
branch.
Alternating the two realizes DMD optimization within a shared model.

In practice, we adopt three implementation choices.
First, motivated by the ``one-input, dual-output'' design of Duality Models
(DuMo)~\citep{sun2026duality}, we attach two output heads to the shared DiT
backbone. The fake-score head produces the negative-time prediction used to
compute $\mathbf{s}_{\mathrm{fake}}$, whereas the DMD head parameterizes the
positive-time generator optimized by \eqref{eq:distill_dmd_loss}. Thus, calls to
$\boldsymbol{F}_{\theta}$ at $-t$ and $+t$ in the notation above select the
fake-score and DMD heads, respectively. At inference, we retain only the DMD
head and discard the fake-score head, so the auxiliary branch introduces no
inference-time overhead.
Second, we use four-step backward simulation to align the training inputs with
the student's inference trajectory.
Third, we update the generator and fake-score branch at a $1{:}2$ schedule.
Compared with the $1{:}5$ schedule in DMD2~\citep{yin2024dmd2}, this reduces
fake-score updates and improves training efficiency.
After distillation, we further refine the model with TBSM~\citep{sun2026tbsm}-based
reinforcement tuning.

\section{Experiments}
\label{sec:experiments}

\subsection{Text-to-Image Generation}
\label{sec:text-to-image-generation}

\subsubsection{General Image Generation Performance}
\label{sec:general-evaluation}

Unless otherwise specified, the baseline results in this section are the raw
scores reported by the corresponding models. For consistent evaluation, all
benchmark results for \lladaimageturbo are obtained with four sampling steps.
We exclude results obtained with prompt enhancement, explicit thinking or
reasoning, or other test-time prompt rewriting pipelines, so that the comparison
reflects the generation model under its standard inference setting.

\paragraph{Compared systems and sources.}
The proprietary comparison set comprises GPT-Image 2~\citep{openai2026gptimage2},
GPT-Image 1.5~\citep{openai2025gptimage15}, GPT-Image 1 (including its High
setting)~\citep{openai2025gptimage1}, Nano-Banana 2.0~\citep{google2026nanobanana2},
Nano-Banana Pro~\citep{google2025nanobananapro}, Qwen-Image 2.0
Pro~\citep{zhao2026qwenimage2}, Seedream 5.0~\citep{bytedance2026seedream5},
Seedream 4.5~\citep{bytedance2025seedream45}, Seedream
4.0~\citep{bytedance2025seedream4}, FLUX.2 Max and Pro~\citep{blackforestlabs2026flux2},
Imagen 4.0 and Imagen 4.0 Ultra~\citep{google2025imagen4}, and Kling-Image
2.1~\citep{kuaishou2025klingimage21}.
The remaining baselines include LLaDA 2.0 Uni~\citep{ai2026llada2}, SenseNova U1.5
Preview~\citep{sensenova2026u15}, Lumina-DiMOO~\citep{dimoo}, the Qwen-Image
family~\citep{wu2025qwen,qwen2025image2512}, the LongCat-Image
family~\citep{team2025longcat}, the Boogu-Image 0.1 Base and Turbo
variants~\citep{chen2026boogu}, InternVL-U~\citep{tian2026internvl}, the Z-Image
family~\citep{cai2025z}, Seedream 3.0~\citep{gao2025seedream},
GLM-Image~\citep{zai2026glmimage}, Lumina-Image 2.0~\citep{qin2025lumina},
LLaDA-o~\citep{you2026lladao}, HunyuanImage 3.0~\citep{cao2025hunyuanimage},
NextFlow~\citep{zhang2026nextflow}, Janus Pro~\citep{chen2025janus},
OmniGen2~\citep{wu2025omnigen2}, Emu3-Gen~\citep{wang2024emu3},
MMaDA~\citep{Mmada}, Lumina-mGPT 2.0~\citep{xin2025lumina},
X-Omni~\citep{xomni}, BAGEL~\citep{bagel}, FLUX.1 [Dev]~\citep{blackforestlabs2024flux1},
Kolors 2.0~\citep{kuaishou2025kolors2}, HiDream-I1 Full~\citep{cai2025hidream},
and BLIP3-o~\citep{chen2025blip3}.

\paragraph{Qwen-Image-Bench.}
Qwen-Image-Bench~\citep{li2026qwenimagebench} is a creator-oriented benchmark that uses 1,000 stratified
prompts to assess five complementary aspects of practical image generation:
visual quality, aesthetics, prompt alignment, real-world fidelity, and creative
generation. Its English and Chinese tracks also allow us to examine whether a
model maintains the same breadth of capabilities across languages rather than
specializing in a single prompt distribution. \figref{fig:qwen-image-bench-overall}
summarizes the overall comparison, while
\tabref{table:qwen-image-bench-en} and \tabref{table:qwen-image-bench-cn}
provide the dimension-level breakdowns.
\lladaimage obtains overall scores of 53.53 on the English track and 53.38 on
the Chinese track. These results exceed the next-best open-source baseline,
Z-Image Turbo, by 1.87 and 0.67 points, respectively, establishing a new
open-source state of the art on both tracks without prompt enhancement or a
test-time thinking pipeline. The gains are broad: \lladaimage ranks first
among the compared open-source models in Quality, Aesthetics, and Alignment
in both languages, as well as in Creative Generation on the English track.

\begin{table}[H]
    \centering
    \renewcommand{\arraystretch}{1.28}
    \setlength\tabcolsep{2pt}
    \small
    \newcolumntype{C}{>{\centering\arraybackslash}X}

    \caption{Comparison of Text-to-Image Generation Performance on Qwen-Image-Bench (EN).}
    \begin{tabularx}{\linewidth}{c l *{6}{C}}
        \toprule
        \textbf{Type}
         & \textbf{Model}
         & \textbf{Quality}
         & \textbf{Aesthetics}
         & \textbf{Alignment}
         & \textbf{\makecell{Real-world                                                 \\Fidelity}}
         & \textbf{\makecell{Creative                                                   \\Generation}}
         & \textbf{Overall}                                                             \\
        \midrule
        \multirow{14}{*}{\rotatebox[origin=c]{90}{Closed-source}}
         & GPT-Image 2                  & 59.09 & 68.48 & 65.78 & 59.40 & 75.34 & 65.23 \\
         & GPT-Image 1.5                & 55.78 & 62.87 & 61.39 & 55.86 & 67.06 & 60.42 \\
         & Nano-Banana 2.0              & 54.86 & 62.63 & 61.11 & 54.66 & 64.49 & 59.59 \\
         & Nano-Banana Pro              & 55.30 & 61.38 & 60.30 & 55.91 & 64.54 & 59.33 \\
         & Qwen-Image 2.0 Pro           & 55.16 & 60.36 & 57.86 & 53.06 & 63.59 & 57.90 \\
         & Seedream 5.0                 & 54.01 & 59.96 & 58.63 & 53.86 & 63.64 & 57.80 \\
         & Seedream 4.5                 & 54.05 & 60.11 & 57.44 & 51.55 & 59.82 & 56.95 \\
         & Seedream 4.0                 & 54.40 & 59.26 & 56.75 & 52.68 & 57.64 & 56.48 \\
         & FLUX.2 Max                   & 53.99 & 58.77 & 57.31 & 50.69 & 57.79 & 56.14 \\
         & FLUX.2 Pro                   & 52.88 & 58.71 & 57.59 & 48.45 & 57.36 & 55.61 \\
         & GPT-Image 1                  & 52.50 & 55.77 & 55.35 & 48.25 & 57.29 & 54.24 \\
         & Imagen 4.0 Ultra             & 51.16 & 55.64 & 53.75 & 46.00 & 51.32 & 52.42 \\
         & Imagen 4.0                   & 50.63 & 53.93 & 52.56 & 45.13 & 48.56 & 51.08 \\
         & Kling-Image 2.1              & 49.04 & 50.94 & 50.47 & 44.29 & 46.23 & 48.89 \\
        \midrule
        \rowcolor{myblue}
        \multirow{11}{*}{\rotatebox[origin=c]{90}{Open-source}}
         & \lladaimage                  & 53.22 & 58.22 & 54.77 & 43.90 & 51.09 & 53.53 \\
         & Z-Image Turbo                & 51.25 & 54.87 & 53.61 & 44.08 & 49.20 & 51.66 \\
         & Boogu-Image 0.1 Turbo        & 51.30 & 53.91 & 53.37 & 46.52 & 48.90 & 51.61 \\
         & HunyuanImage 3.0             & 50.76 & 54.66 & 53.16 & 45.33 & 48.33 & 51.35 \\
         & Qwen-Image 2512              & 51.84 & 54.40 & 51.44 & 47.80 & 47.75 & 51.32 \\
         & Boogu-Image 0.1 Base         & 50.38 & 53.90 & 52.62 & 46.54 & 47.52 & 51.00 \\
         & \lladaimageturbo             & 51.89 & 55.22 & 52.49 & 42.08 & 45.75 & 50.98 \\
         & Z-Image                      & 49.42 & 53.88 & 53.41 & 43.44 & 46.93 & 50.52 \\
         & SenseNova U1.5 Preview       & 49.14 & 52.02 & 52.46 & 45.22 & 47.23 & 49.93 \\
         & Qwen-Image                   & 48.45 & 51.18 & 50.04 & 43.45 & 45.37 & 48.48 \\
         & GLM-Image                    & 49.86 & 49.98 & 47.49 & 44.25 & 44.67 & 47.86 \\
        \bottomrule
    \end{tabularx}
    \label{table:qwen-image-bench-en}
\end{table}

\begin{table}[!t]
    \centering
    \renewcommand{\arraystretch}{1.28}
    \setlength\tabcolsep{2pt}
    \small
    \newcolumntype{C}{>{\centering\arraybackslash}X}

    \caption{Comparison of Text-to-Image Generation Performance on Qwen-Image-Bench (CN).}
    \begin{tabularx}{\linewidth}{c l *{6}{C}}
        \toprule
        \textbf{Type}
         & \textbf{Model}
         & \textbf{Quality}
         & \textbf{Aesthetics}
         & \textbf{Alignment}
         & \textbf{\makecell{Real-world                                                 \\Fidelity}}
         & \textbf{\makecell{Creative                                                   \\Generation}}
         & \textbf{Overall}                                                             \\
        \midrule
        \multirow{14}{*}{\rotatebox[origin=c]{90}{Closed-source}}
         & GPT-Image 2                  & 58.65 & 67.53 & 65.85 & 57.38 & 75.23 & 64.69 \\
         & Nano-Banana 2.0              & 54.77 & 61.08 & 62.40 & 54.28 & 67.05 & 59.82 \\
         & GPT-Image 1.5                & 55.14 & 60.88 & 61.72 & 53.95 & 66.35 & 59.65 \\
         & Nano-Banana Pro              & 55.67 & 60.26 & 61.25 & 54.07 & 66.23 & 59.45 \\
         & Qwen-Image 2.0 Pro           & 54.39 & 58.67 & 59.28 & 51.83 & 64.94 & 57.84 \\
         & Seedream 5.0                 & 52.55 & 58.40 & 58.90 & 51.92 & 65.29 & 57.22 \\
         & Seedream 4.5                 & 54.41 & 58.72 & 57.31 & 51.69 & 60.64 & 56.78 \\
         & Seedream 4.0                 & 54.01 & 58.81 & 56.64 & 51.05 & 58.15 & 56.21 \\
         & FLUX.2 Max                   & 53.64 & 56.85 & 57.35 & 49.35 & 56.50 & 55.33 \\
         & FLUX.2 Pro                   & 52.30 & 56.94 & 57.01 & 47.29 & 56.18 & 54.57 \\
         & GPT-Image 1                  & 52.34 & 55.09 & 56.28 & 48.14 & 55.78 & 54.07 \\
         & Imagen 4.0 Ultra             & 50.90 & 54.25 & 54.02 & 45.59 & 51.14 & 51.99 \\
         & Imagen 4.0                   & 50.16 & 52.68 & 51.64 & 44.84 & 47.94 & 50.29 \\
         & Kling-Image 2.1              & 49.11 & 50.15 & 49.18 & 44.74 & 44.67 & 48.26 \\
        \midrule
        \rowcolor{myblue}
        \multirow{11}{*}{\rotatebox[origin=c]{90}{Open-source}}
         & \lladaimage                  & 52.92 & 56.92 & 54.87 & 44.86 & 52.10 & 53.38 \\
         & Z-Image Turbo                & 52.30 & 55.24 & 53.76 & 45.38 & 52.32 & 52.71 \\
         & Qwen-Image 2512              & 51.76 & 54.74 & 52.72 & 47.00 & 50.19 & 52.06 \\
         & Z-Image                      & 50.53 & 54.47 & 54.45 & 44.70 & 49.92 & 51.78 \\
         & Boogu-Image 0.1 Turbo        & 51.24 & 53.50 & 53.54 & 46.11 & 48.91 & 51.53 \\
         & Boogu-Image 0.1 Base         & 50.41 & 53.14 & 52.91 & 45.42 & 48.62 & 50.96 \\
         & HunyuanImage 3.0             & 50.35 & 53.57 & 52.00 & 44.31 & 49.12 & 50.81 \\
         & \lladaimageturbo             & 51.29 & 52.02 & 52.32 & 42.91 & 47.14 & 50.27 \\
         & SenseNova U1.5 Preview       & 49.19 & 51.87 & 52.61 & 45.11 & 48.80 & 50.25 \\
         & Qwen-Image                   & 48.44 & 52.25 & 50.72 & 43.16 & 47.30 & 49.23 \\
         & GLM-Image                    & 49.26 & 50.64 & 47.90 & 44.69 & 45.23 & 48.19 \\
        \bottomrule
    \end{tabularx}
    \label{table:qwen-image-bench-cn}
\end{table}

\paragraph{LongText-Bench.}
LongText-Bench~\citep{xomni} evaluates whether a model can render long text strings legibly
and correctly in both English and Chinese. Compared with short-word or
single-region tests, it is more sensitive to omitted, repeated, substituted,
and malformed characters as the amount of requested text increases. As
reported in \tabref{table:longtext-bench}, \lladaimage achieves
0.923 and 0.913 on the English and Chinese subsets, respectively. The small
cross-language gap indicates balanced bilingual behavior, although the scores
remain below the strongest dedicated text-rendering models. Because this
benchmark primarily emphasizes transcription correctness, we use it together
with CVTG-2K rather than treating it as a standalone measure of the overall
visual quality of text-centric images.

\paragraph{CVTG-2K.}
CVTG-2K~\citep{tai2025cvtg} focuses on complex visual text generation in practical scenarios such
as advertisements, posters, memes, and street views. Each prompt contains two
to five text regions, making the benchmark useful for measuring both rendering
accuracy and robustness as the number of textual elements grows. We report
word accuracy for each region count and its average, normalized edit distance
(NED) for transcription fidelity, and CLIPScore~\citep{hessel2021clipscore} for image--prompt alignment.
\tabref{table:cvtg2k_rendering} shows that \lladaimage obtains an average
word accuracy of 0.875, the second-highest among all compared models, together
with an NED of 0.945 and a CLIPScore of 0.818. Its word accuracy remains stable
as the number of text regions increases, decreasing from 0.892 on the
two-region subset to 0.857 on the five-region subset.

\begin{table}[H]
    \centering
    \renewcommand{\arraystretch}{1.2}
    \setlength\tabcolsep{6pt}
    \small
    \newcolumntype{C}{>{\centering\arraybackslash}X}

    \caption{Comparison of Long-Text Rendering Ability on LongText-Bench.}
    \begin{tabularx}{\linewidth}{lCC}
        \toprule
        \textbf{Model}
                              & \textbf{LongText-Bench-EN~$\uparrow$}
                              & \textbf{LongText-Bench-ZH~$\uparrow$}         \\
        \midrule
        Qwen-Image 2512       & 0.956                                 & 0.965 \\
        Boogu-Image 0.1 Base  & 0.952                                 & 0.969 \\
        Boogu-Image 0.1 Turbo & 0.944                                 & 0.977 \\
        Qwen-Image            & 0.943                                 & 0.946 \\
        Z-Image               & 0.935                                 & 0.936 \\
        Z-Image Turbo         & 0.917                                 & 0.926 \\
        \rowcolor{myblue}
        \lladaimage           & 0.923                                 & 0.913 \\
        \rowcolor{myblue}
        \lladaimageturbo      & 0.899                                 & 0.919 \\
        Seedream 3.0          & 0.896                                 & 0.878 \\
        X-Omni                & 0.900                                 & 0.814 \\
        GPT-Image 1 [High]    & 0.956                                 & 0.619 \\
        BAGEL                 & 0.373                                 & 0.310 \\
        OmniGen2              & 0.561                                 & 0.059 \\
        FLUX.1 [Dev]          & 0.607                                 & 0.005 \\
        Kolors 2.0            & 0.258                                 & 0.329 \\
        HiDream-I1 Full       & 0.543                                 & 0.024 \\
        BLIP3-o               & 0.021                                 & 0.018 \\
        Janus Pro             & 0.019                                 & 0.006 \\
        \bottomrule
    \end{tabularx}
    \label{table:longtext-bench}
\end{table}

\begin{table}[H]
    \centering
    \renewcommand{\arraystretch}{1.5}
    \setlength\tabcolsep{1.5pt}
    \small
    \newcolumntype{C}{>{\centering\arraybackslash}X}

    \caption{Comparison of Text Rendering Ability on CVTG-2K Benchmark.}
    \begin{tabularx}{\linewidth}{l *{5}{C} CC}
        \toprule
        \multirow{2}{*}{\textbf{Model}} & \multicolumn{5}{c}{\textbf{Word Accuracy~$\uparrow$}} & \multirow{2}{*}{\textbf{NED~$\uparrow$}} & \multirow{2}{*}{\textbf{CLIPScore~$\uparrow$}}                                       \\
        \cmidrule(lr){2-6}
                                        & 2 regions                                              & 3 regions                                & 4 regions                                      & 5 regions & average &       &       \\
        \midrule
        SenseNova U1.5 Preview          & 0.890                                                  & 0.887                                    & 0.902                                          & 0.869     & 0.887   & 0.943 & 0.809 \\        \rowcolor{myblue}
        \lladaimage                     & 0.892                                                  & 0.878                                    & 0.882                                          & 0.857     & 0.875   & 0.945 & 0.818 \\
        Z-Image                         & 0.901                                                  & 0.872                                    & 0.865                                          & 0.851     & 0.867   & 0.937 & 0.797 \\
        LongCat-Image                   & 0.912                                                  & 0.873                                    & 0.855                                          & 0.831     & 0.865   & 0.936 & 0.785 \\
        Boogu-Image 0.1 Base            & 0.852                                                  & 0.863                                    & 0.866                                          & 0.863     & 0.863   & 0.927 & 0.801 \\
        Z-Image Turbo                   & 0.887                                                  & 0.866                                    & 0.862                                          & 0.834     & 0.858   & 0.928 & 0.804 \\
        \rowcolor{myblue}
        \lladaimageturbo                & 0.846                                                  & 0.843                                    & 0.835                                          & 0.840     & 0.840   & 0.938 & 0.818 \\
        Qwen-Image 2512                 & 0.841                                                  & 0.843                                    & 0.842                                          & 0.827     & 0.838   & 0.918 & 0.788 \\        Qwen-Image                      & 0.837                                                  & 0.836                                    & 0.831                                          & 0.816     & 0.829   & 0.912 & 0.802 \\
        Boogu-Image 0.1 Turbo           & 0.792                                                  & 0.814                                    & 0.832                                          & 0.822     & 0.819   & 0.910 & 0.791 \\
        InternVL-U                      & 0.729                                                  & 0.660                                    & 0.618                                          & 0.549     & 0.623   & 0.804 & 0.816 \\
        Seedream 3.0                    & 0.628                                                  & 0.596                                    & 0.604                                          & 0.561     & 0.592   & 0.853 & 0.782 \\
        Lumina-DiMOO                    & 0.723                                                  & 0.646                                    & 0.571                                          & 0.505     & 0.590   & 0.805 & 0.831 \\
        FLUX.1 [Dev]                    & 0.608                                                  & 0.553                                    & 0.466                                          & 0.431     & 0.496   & 0.687 & 0.740 \\
        BAGEL                           & 0.498                                                  & 0.391                                    & 0.332                                          & 0.291     & 0.356   & 0.657 & 0.779 \\
        \bottomrule
    \end{tabularx}
    \label{table:cvtg2k_rendering}
\end{table}

\paragraph{Legacy diagnostic benchmarks.}
As discussed in the Boogu-Image 0.1 technical report~\citep{chen2026boogu},
GenEval and DPG-Bench
do not reliably track the actual capabilities of modern text-to-image models:
their rankings can diverge substantially from human-preference rankings, and
their increasingly saturated scores make small differences difficult to
interpret. We therefore do not treat either benchmark as primary evidence of
overall generation quality. Nevertheless, both benchmarks have been widely
reported in previous work. For historical continuity and comparison with the
existing literature, we include their results below as supplementary
references.

\paragraph{GenEval.}
GenEval~\citep{ghosh2023geneval} provides an object-centric diagnostic of compositional text-to-image
generation. It decomposes prompt following into single-object generation,
two-object composition, counting, color, spatial position, and attribute
binding, thereby exposing failures that may be hidden by a single holistic
alignment score. As shown in \tabref{table:geneval}, \lladaimage reaches an
overall score of 0.85. It attains 1.00 for single-object generation, 0.98 for
two-object composition, and a tied-highest Attribute Binding score of 0.84
among the compared models. Its lower Counting score of 0.53 limits the aggregate
result and exposes a specific compositional weakness. We report these category
scores as a fine-grained reference under this traditional evaluation protocol,
rather than interpreting the overall ranking as a faithful measure of general
generation quality.

\begin{table}[H]
    \centering
    \renewcommand{\arraystretch}{1.5}
    \setlength\tabcolsep{1.5pt}
    \small
    \newcolumntype{C}{>{\centering\arraybackslash}X}

    \caption{Comparison of Text-to-Image Generation Ability on GenEval Benchmark.}
    \begin{tabularx}{\linewidth}{l *{6}{C} c}
        \toprule
        \textbf{Model}        & \textbf{\makecell{Single                                           \\Object}} & \textbf{\makecell{Two\\Object}} & \textbf{Counting} & \textbf{Colors} & \textbf{Position} & \textbf{\makecell{Attribute\\Binding}}& \textbf{~Overall$\uparrow$} \\
        \midrule
        LLaDA 2.0 Uni         & 1.00                     & 0.98 & 0.73 & 0.92 & 0.90 & 0.84 & 0.89 \\
        SenseNova U1.5 Preview & 1.00                    & 0.97 & 0.85 & 0.90 & 0.84 & 0.78 & 0.89 \\
        Lumina-DiMOO          & 1.00                     & 0.94 & 0.85 & 0.89 & 0.85 & 0.76 & 0.88 \\
        Qwen-Image            & 0.99                     & 0.92 & 0.89 & 0.88 & 0.76 & 0.77 & 0.87 \\
        LongCat-Image         & 0.99                     & 0.98 & 0.86 & 0.86 & 0.75 & 0.73 & 0.87 \\
        Boogu-Image 0.1 Base  & 0.99                     & 0.95 & 0.80 & 0.84 & 0.85 & 0.68 & 0.85 \\
        InternVL-U            & 0.99                     & 0.94 & 0.74 & 0.91 & 0.77 & 0.74 & 0.85 \\
        \rowcolor{myblue}
        \lladaimage           & 1.00                     & 0.98 & 0.53 & 0.93 & 0.78 & 0.84 & 0.85 \\
        Seedream 3.0          & 0.99                     & 0.96 & 0.91 & 0.93 & 0.47 & 0.80 & 0.84 \\
        Boogu-Image 0.1 Turbo & 1.00                     & 0.97 & 0.85 & 0.76 & 0.86 & 0.60 & 0.84 \\
        Z-Image               & 1.00                     & 0.94 & 0.78 & 0.93 & 0.62 & 0.77 & 0.84 \\
        NextFlow              & 0.98                     & 0.92 & 0.73 & 0.90 & 0.77 & 0.69 & 0.83 \\
        Z-Image Turbo         & 1.00                     & 0.95 & 0.77 & 0.89 & 0.65 & 0.68 & 0.82 \\
        \rowcolor{myblue}
        \lladaimageturbo      & 1.00                     & 0.99 & 0.42 & 0.93 & 0.78 & 0.80 & 0.82 \\
        BAGEL                 & 0.99                     & 0.94 & 0.81 & 0.88 & 0.64 & 0.63 & 0.82 \\
        Janus Pro             & 0.99                     & 0.89 & 0.59 & 0.90 & 0.79 & 0.66 & 0.80 \\
        OmniGen2              & 1.00                     & 0.95 & 0.64 & 0.88 & 0.55 & 0.76 & 0.80 \\
        Qwen-Image 2512       & 1.00                     & 0.92 & 0.57 & 0.90 & 0.47 & 0.66 & 0.75 \\
        HunyuanImage 3.0      & 1.00                     & 0.92 & 0.48 & 0.82 & 0.42 & 0.63 & 0.72 \\
        Lumina-mGPT 2.0       & 0.99                     & 0.87 & 0.44 & 0.85 & 0.44 & 0.54 & 0.69 \\
        FLUX.1 [Dev]          & 0.98                     & 0.81 & 0.74 & 0.79 & 0.22 & 0.45 & 0.66 \\
        MMaDA                 & 0.99                     & 0.76 & 0.61 & 0.84 & 0.20 & 0.37 & 0.63 \\
        Emu3-Gen              & 0.98                     & 0.71 & 0.34 & 0.81 & 0.17 & 0.21 & 0.54 \\
        \bottomrule
    \end{tabularx}
    \label{table:geneval}
\end{table}

\paragraph{DPG-Bench.}
DPG-Bench~\citep{hu2024ella} evaluates dense prompt following using 1,065 long, structurally
complex descriptions. Its fine-grained evaluation decomposes a prompt into
global, entity, attribute, relation, and other constraints, which complements
GenEval's concise object-centric templates with a more realistic test of
verbose instruction interpretation. \tabref{table:dpg_benchmark} reports
an overall score of 87.48 for \lladaimage. The model remains consistently
strong across the Entity, Attribute, Relation, and Other dimensions, with
scores of 92.31, 92.37, 92.55, and 91.96, respectively. This balanced profile
places it ahead of several strong baselines, including LLaDA-o, LongCat-Image,
and HunyuanImage 3.0. We nevertheless include these scores as a reference for
comparison with prior work, rather than as a definitive measure of the
model's practical generation capability.

\begin{table}[H]
    \centering
    \renewcommand{\arraystretch}{1.2}
    \setlength\tabcolsep{1pt}
    \small
    \newcolumntype{C}{>{\centering\arraybackslash}X}

    \caption{Comparison of Text-to-Image Generation Ability on DPG Benchmark.}
    \begin{tabularx}{\linewidth}{l *{5}{C} c}
        \toprule
        \textbf{Model}        & \textbf{Global} & \textbf{Entity} & \textbf{ Attribute} & \textbf{Relation} & \textbf{Other} & \textbf{~Overall$\uparrow$} \\
        \midrule
        \rowcolor{myblue}
        \lladaimageturbo      & 92.23           & 91.86           & 93.35               & 91.01             & 91.53          & 88.55                       \\
        Boogu-Image 0.1 Turbo & 88.54           & 91.67           & 92.19               & 93.20             & 93.83          & 88.35                       \\
        Qwen-Image            & 91.32           & 91.56           & 92.02               & 94.31             & 92.73          & 88.32                       \\
        Seedream 3.0          & 94.31           & 92.65           & 91.36               & 92.78             & 88.24          & 88.27                       \\
        Z-Image               & 93.39           & 91.22           & 93.16               & 92.22             & 91.52          & 88.14                       \\
        LLaDA 2.0 Uni         & 91.14           & 93.55           & 91.98               & 92.17             & 93.18          & 87.76                       \\
        \rowcolor{myblue}
        \lladaimage           & 90.86           & 92.31           & 92.37               & 92.55             & 91.96          & 87.48                       \\
        SenseNova U1.5 Preview & 85.81           & 91.97           & 90.95               & 93.48             & 92.53          & 87.26                       \\
        Qwen-Image 2512       & 89.04           & 91.91           & 92.39               & 90.85             & 93.07          & 87.20                       \\
        Lumina-Image 2.0      & 86.63           & 91.97           & 90.20               & 94.85             & 84.80          & 87.20                       \\
        Boogu-Image 0.1 Base  & 89.33           & 90.64           & 92.22               & 93.72             & 93.32          & 87.13                       \\
        LLaDA-o               & 92.91           & 93.30           & 90.40               & 91.75             & 92.79          & 87.04                       \\
        LongCat-Image         & 89.10           & 92.54           & 92.00               & 93.28             & 87.50          & 86.80                       \\
        HunyuanImage 3.0      & 92.12           & 92.53           & 89.13               & 92.13             & 91.92          & 86.10                       \\
        Lumina-DiMOO          & 81.46           & 92.08           & 88.98               & 94.31             & 82.00          & 86.04                       \\
        NextFlow              & 92.40           & 90.05           & 90.51               & 92.72             & 91.14          & 86.00                       \\
        InternVL-U            & 90.39           & 90.78           & 90.68               & 90.29             & 88.77          & 85.18                       \\
        BAGEL                 & 88.94           & 90.37           & 91.29               & 90.82             & 88.67          & 85.07                       \\
        Z-Image Turbo         & 91.29           & 89.59           & 90.14               & 92.16             & 88.68          & 84.86                       \\
        Janus Pro             & 86.90           & 88.90           & 89.40               & 89.32             & 89.48          & 84.19                       \\
        FLUX.1 [Dev]          & 74.35           & 90.00           & 88.96               & 90.87             & 88.33          & 83.84                       \\
        OmniGen2              & 88.81           & 88.83           & 90.18               & 89.37             & 90.27          & 83.57                       \\
        Emu3-Gen              & 85.21           & 86.68           & 86.84               & 90.22             & 83.15          & 80.60                       \\
        MMaDA                 & 77.81           & 78.48           & 81.74               & 84.79             & 63.20          & 69.97                       \\
        \bottomrule
    \end{tabularx}
    \label{table:dpg_benchmark}
\end{table}

\subsection{Image Editing Generation}
\label{sec:image-editing-generation}

\paragraph{Compared systems and sources.}
The editing comparison covers SenseNova U1.5 Preview~\citep{sensenova2026u15},
FireRed-Image-Edit~\citep{firered2026edit}, Qwen-Image-Edit 2511 and
Qwen-Image-Edit 2509~\citep{qwen2025imageedit2511,qwen2025imageedit2509},
Seedream 4.5 and 4.0~\citep{bytedance2025seedream45,bytedance2025seedream4},
Nano-Banana Pro and Nano-Banana~\citep{google2025nanobananapro,google2025nanobanana},
LongCat-Image-Edit~\citep{team2025longcat}, Z-Image-Edit~\citep{cai2025z},
Step1X-Edit v1.2~\citep{liu2025step1xedit}, FLUX.2 [Dev]~\citep{blackforestlabs2026flux2},
and LLaDA 2.0 Uni~\citep{ai2026llada2}.

\paragraph{GEdit-Bench.}
We evaluate general instruction-based image editing on the English and Chinese
tracks of GEdit-Bench~\citep{liu2025step1xedit}, which contains 606 real-world editing cases spanning 11
task categories. The benchmark separately measures semantic consistency and
instruction following ($G_{\mathrm{SC}}$), perceptual quality and naturalness
($G_{\mathrm{PQ}}$), and their overall assessment ($G_{\mathrm{O}}$). This
decomposition is particularly relevant to a unified generation-and-editing
model: a successful output must execute the requested change while retaining
the visual plausibility and unedited content of the reference image. As
reported in \tabref{table:gedit-bench}, \lladaimage achieves overall scores
of 7.336 and 7.294 on GEdit-Bench-EN and GEdit-Bench-CN, respectively. These
results confirm that a single unified checkpoint can support bilingual image
editing while retaining its text-to-image capability. The component scores
also reveal that semantic consistency is more competitive than perceptual
quality, particularly on the English track ($8.043$ versus $7.182$). Closing
this perceptual-quality gap relative to specialized editing models remains an
important direction for future refinement.

\begin{table}[H]
    \centering
    \renewcommand{\arraystretch}{1.2}
    \setlength\tabcolsep{2pt}
    \small
    \newcolumntype{C}{>{\centering\arraybackslash}X}

    \caption{Comparison of Image Editing Performance on GEdit-Bench.}
    \begin{tabularx}{\linewidth}{l *{6}{C}}
        \toprule
        \multirow{2}{*}{\textbf{Model}}
                               & \multicolumn{3}{c}{\textbf{GEdit-Bench-EN}}
                               & \multicolumn{3}{c}{\textbf{GEdit-Bench-CN}}                                         \\
        \cmidrule(lr){2-4}\cmidrule(lr){5-7}
                               & \textbf{$G_{\mathrm{SC}}\uparrow$}
                               & \textbf{$G_{\mathrm{PQ}}\uparrow$}
                               & \textbf{$G_{\mathrm{O}}\uparrow$}
                               & \textbf{$G_{\mathrm{SC}}\uparrow$}
                               & \textbf{$G_{\mathrm{PQ}}\uparrow$}
                               & \textbf{$G_{\mathrm{O}}\uparrow$}                                                   \\
        \midrule
        SenseNova U1.5 Preview & --                                          & --    & 8.172 & --    & --    & 8.051 \\
        FireRed-Image-Edit     & 8.363                                       & 8.245 & 7.943 & 8.287 & 8.227 & 7.887 \\
        Qwen-Image-Edit 2511   & 8.297                                       & 8.202 & 7.877 & 8.252 & 8.134 & 7.819 \\
        Seedream 4.5           & 8.268                                       & 8.167 & 7.820 & 8.254 & 8.167 & 7.800 \\
        Nano-Banana Pro        & 8.102                                       & 8.344 & 7.738 & 8.135 & 8.306 & 7.799 \\
        LongCat-Image-Edit     & 8.128                                       & 8.177 & 7.748 & 8.141 & 8.117 & 7.731 \\
        Seedream 4.0           & 8.143                                       & 8.124 & 7.701 & 8.159 & 8.074 & 7.692 \\
        Z-Image-Edit           & 8.11                                        & 7.72  & 7.57  & 8.03  & 7.80  & 7.54  \\
        Qwen-Image-Edit 2509   & 7.974                                       & 7.714 & 7.480 & 7.988 & 7.679 & 7.467 \\
        FLUX.2 [Dev]           & 7.835                                       & 8.064 & 7.413 & 7.697 & 8.046 & 7.278 \\
        Nano-Banana            & 7.396                                       & 8.454 & 7.291 & 7.540 & 8.424 & 7.399 \\
        \rowcolor{myblue}
        \lladaimage            & 8.043                                       & 7.182 & 7.336 & 7.707 & 7.594 & 7.294 \\
        \rowcolor{myblue}
        \lladaimageturbo       & 7.533                                       & 7.200 & 7.024 & 7.252 & 7.292 & 6.898 \\
        LLaDA 2.0 Uni          & 6.68                                        & 7.52  & 6.61  & 6.63  & 7.67  & 6.66  \\
        \bottomrule
    \end{tabularx}
    \label{table:gedit-bench}
\end{table}

\FloatBarrier

\section{Conclusion and Future Directions}
\label{sec:conclusion}

This report presents \lladaimage, a 6B Diffusion Transformer trained from
scratch for unified text-to-image generation and instruction-guided editing.
Its open recipe combines image-only pre-training and mid-training, a dLLM-based
VLM with RQA--connector conditioning, a real-data-dominant progressive
curriculum, and TwinFlow distillation. Of 220M generation samples, $98\%$ are
real and more than $90\%$ support image-only training; real images remain above
$70\%$ in paired SFT. SigLIP-VQ and clean VAE latents preserve reference
information, while parameter-free RMSNorm and Muon stabilize training.
\lladaimage sets the open-source state of the art on both Qwen-Image-Bench
tracks, with strong bilingual text rendering and editing; \lladaimageturbo
reduces inference to 2--4 steps. We release the
weights, code, and recipes to make this path inspectable and reusable.

Remaining challenges include broader world knowledge and more reliable long
and multi-region text rendering. Broader knowledge coverage would improve the
faithful depiction of specialized subjects, culturally specific concepts, and
prompts whose visual realization depends on factual context. Text rendering can
also become less stable as the number of regions, the length of the requested
content, and the complexity of the layout increase. Training currently reaches
a nominal $1024^2$ pixel budget, so native 2K generation remains an extension
rather than an evaluated capability. Future work will strengthen grounding and
typography, broaden the coverage of knowledge-intensive concepts, extend the
curriculum to native 2K, and explore agentic generation through closed-loop
planning, evaluation, and editing.

\clearpage
\bibliographystyle{antgroup}\bibliography{ref/reference}

\clearpage
\appendix
\section{Aspect-Ratio Bucket Configurations}
\label{app:aspect-ratio-buckets}

We introduce aspect-ratio buckets during image-only mid-training and retain
them throughout supervised fine-tuning at two target pixel budgets.
The terms ``$512^2$-pixel'' and ``$1024^2$-pixel'' refer to the approximate
number of pixels in an image, not to a fixed square training resolution.
\tabref{tab:aspect-ratio-buckets} lists the exact configurations. The aspect
ratio is defined as width divided by height, and each spatial configuration is
reported as a $(W,H)$ tuple in pixels.

Let an input image have width $W$, height $H$, and aspect ratio
$a=W/H$. The data loader maintains the set of $K$ resolution buckets
$\mathcal{B}=\{(W_k,H_k)\}_{k=1}^{K}$ and assigns the image to the bucket with
the smallest absolute difference in aspect ratio:
\begin{equation}
  k^{*}
  = \underset{1\leq k\leq K}{\operatorname{arg\,min}}
  \left|
  \frac{W}{H}
  - \frac{W_k}{H_k}
  \right|.
  \label{eq:aspect-ratio-bucket-assignment}
\end{equation}
After selecting $k^{*}$ via \eqref{eq:aspect-ratio-bucket-assignment}, we
resize the image without changing its aspect ratio. For the selected bucket
$(W_{k^{*}},H_{k^{*}})$, the resize-to-fill scale is
\begin{equation}
  s
  = \max\!\left(
  \frac{W_{k^{*}}}{W},
  \frac{H_{k^{*}}}{H}
  \right),
  \qquad
  (W',H')=(sW,sH).
  \label{eq:bucket-resize-scale}
\end{equation}
In practice, the transformation in \eqref{eq:bucket-resize-scale} is a
downsampling step because training images are sufficiently large. One resized
dimension matches its bucket dimension,
whereas the other can be slightly larger. We crop only this excess region to
obtain the exact target shape. Because $k^{*}$ has nearly the same aspect ratio
as the source image, the overflow is typically only a few pixels; consequently,
the crop removes little content and does not materially alter the image.

The buckets within each stage have approximately equal areas
$W_kH_k$. Consequently, different data-parallel ranks can process different
aspect ratios while receiving similar numbers of image tokens. This prevents a
single rank from producing a disproportionate memory peak or becoming a
training-time straggler. Compared with forcing every sample into a square, the
procedure also avoids geometric distortion and aggressive square cropping,
while teaching the model to generate a broad range of output aspect ratios.

\begin{table}[H]
  \centering
  \small
  \setlength{\tabcolsep}{14pt}
  \renewcommand{\arraystretch}{1.05}
  \caption{Aspect-ratio bucket configurations at the $512^2$-pixel and
    $1024^2$-pixel budgets. Each tuple reports $(W,H)$ in pixels.}
  \label{tab:aspect-ratio-buckets}
  \begin{tabular}{ccc}
    \toprule
    \textbf{Aspect Ratio ($W/H$)} & \textbf{$512^2$-pixel $(W,H)$} & \textbf{$1024^2$-pixel $(W,H)$} \\
    \midrule
    \multicolumn{3}{l}{\textbf{Portrait ($W/H<1$)}}                                                  \\
    \midrule
    0.25                          & $(256,1024)$                   & $(512,2048)$                    \\
    0.26                          & $(256,992)$                    & $(512,1984)$                    \\
    0.27                          & $(256,960)$                    & $(512,1920)$                    \\
    0.28                          & $(256,928)$                    & $(512,1856)$                    \\
    0.32                          & $(288,896)$                    & $(576,1792)$                    \\
    0.33                          & $(288,864)$                    & $(576,1728)$                    \\
    0.35                          & $(288,832)$                    & $(576,1664)$                    \\
    0.40                          & $(320,800)$                    & $(640,1600)$                    \\
    0.42                          & $(320,768)$                    & $(640,1536)$                    \\
    0.48                          & $(352,736)$                    & $(704,1472)$                    \\
    0.50                          & $(352,704)$                    & $(704,1408)$                    \\
    0.52                          & $(352,672)$                    & $(704,1344)$                    \\
    0.57                          & $(384,672)$                    & $(768,1344)$                    \\
    0.60                          & $(384,640)$                    & $(768,1280)$                    \\
    0.68                          & $(416,608)$                    & $(832,1216)$                    \\
    0.72                          & $(416,576)$                    & $(832,1152)$                    \\
    0.78                          & $(448,576)$                    & $(896,1152)$                    \\
    0.82                          & $(448,544)$                    & $(896,1088)$                    \\
    0.88                          & $(480,544)$                    & $(960,1088)$                    \\
    0.94                          & $(480,512)$                    & $(960,1024)$                    \\
    \midrule
    \multicolumn{3}{l}{\textbf{Square ($W/H=1$)}}                                                    \\
    \midrule
    1.00                          & $(512,512)$                    & $(1024,1024)$                   \\
    \midrule
    \multicolumn{3}{l}{\textbf{Landscape ($W/H>1$)}}                                                 \\
    \midrule
    1.07                          & $(512,480)$                    & $(1024,960)$                    \\
    1.13                          & $(544,480)$                    & $(1088,960)$                    \\
    1.21                          & $(544,448)$                    & $(1088,896)$                    \\
    1.29                          & $(576,448)$                    & $(1152,896)$                    \\
    1.38                          & $(576,416)$                    & $(1152,832)$                    \\
    1.46                          & $(608,416)$                    & $(1216,832)$                    \\
    1.67                          & $(640,384)$                    & $(1280,768)$                    \\
    1.75                          & $(672,384)$                    & $(1344,768)$                    \\
    2.00                          & $(704,352)$                    & $(1408,704)$                    \\
    2.09                          & $(736,352)$                    & $(1472,704)$                    \\
    2.40                          & $(768,320)$                    & $(1536,640)$                    \\
    2.50                          & $(800,320)$                    & $(1600,640)$                    \\
    2.89                          & $(832,288)$                    & $(1664,576)$                    \\
    3.00                          & $(864,288)$                    & $(1728,576)$                    \\
    3.11                          & $(896,288)$                    & $(1792,576)$                    \\
    3.62                          & $(928,256)$                    & $(1856,512)$                    \\
    3.75                          & $(960,256)$                    & $(1920,512)$                    \\
    3.88                          & $(992,256)$                    & $(1984,512)$                    \\
    4.00                          & $(1024,256)$                   & $(2048,512)$                    \\
    \bottomrule
  \end{tabular}
\end{table}

\section{CoT Supervised Fine-tuning and Reinforcement Learning}
\label{app:cot-sft-understanding}

CoT supervised fine-tuning (\secref{sec:cot_sft}) is the only stage that
updates the understanding backbone, so we verify its effect separately before
the generation stages build on it.

\paragraph{Benchmarks.}
We evaluate across 18 multimodal understanding benchmarks, focusing on three
core capabilities: general VQA, reasoning, and OCR/document understanding:
\begin{itemize}[leftmargin=16pt]
  \item \textbf{General Tasks:} MMStar~\citep{mmstar},
        MMBench~\citep{Mmbench} (English and Chinese \texttt{dev} splits),
        HallusionBench~\citep{Hallusionbench},
        RealWorldQA~\citep{RealworldQA} and SimpleVQA~\citep{Simplevqa}.
  \item \textbf{Reasoning Tasks:} MMMU~\citep{yue2024mmmu} (\texttt{val}),
        MathVista~\citep{mathvista} (\texttt{mini}),
        We-Math~\citep{We-math}, and
        MathVision~\citep{wang2024measuring} (\texttt{mini}).
  \item \textbf{OCR\&Chart Tasks:} ChartQA~\citep{Chartqa},
        DocVQA~\citep{Docvqa} (\texttt{val}), InfoVQA~\citep{InfoVQA},
        OCRBench~\citep{Ocrbench}, and AI2D~\citep{AI2D} (\texttt{test}).
  \item \textbf{Other Multimodal Tasks:} CountBenchQA~\citep{countbench},
        VL-RewardBench~\citep{VLRewardBench}, and V$^{*}$~\citep{vstar}.
\end{itemize}

\begingroup
\begin{table}[H]
\centering
\setlength{\tabcolsep}{18pt}
\renewcommand{\arraystretch}{1.12}

\caption{Multimodal understanding results before and after CoT supervised
fine-tuning. The LLaDA 2.0 Uni column reproduces the scores reported in
its technical report~\citep{ai2026llada2}; the CoT SFT column is evaluated
by us under the protocol described in
\appref{app:cot-sft-understanding}.}
\label{tab:cot-sft-understanding}

\begin{tabular}{@{}lcc@{}}
\toprule
\textbf{Benchmark}
& \multicolumn{1}{c}{\textbf{LLaDA 2.0 Uni}}
& \multicolumn{1}{c}{\textbf{+ CoT SFT}} \\
\midrule

\multicolumn{3}{@{}l}{\textbf{General Tasks}} \\
\addlinespace[2pt]
MMStar                           & 64.1 & 64.1 \\
MMBench$_{\text{dev-EN}}$        & 81.5 & 86.1 \\
MMBench$_{\text{dev-CN}}$        & 81.2 & 86.2 \\
HallusionBench                   & 50.2 & 51.3 \\
RealWorldQA                      & 66.7 & 64.1 \\
SimpleVQA                        & 44.0 & 42.1 \\

\addlinespace[5pt]
\midrule
\multicolumn{3}{@{}l}{\textbf{Reasoning Tasks}} \\
\addlinespace[2pt]
MMMU$_{\text{val}}$              & 50.1 & 51.3 \\
MathVista$_{\text{mini}}$        & 68.1 & 68.4 \\
MathVision$_{\text{mini}}$       & 26.7 & 24.3 \\
We-Math                          & 29.3 & 30.9 \\

\addlinespace[5pt]
\midrule
\multicolumn{3}{@{}l}{\textbf{OCR \& Chart Tasks}} \\
\addlinespace[2pt]
ChartQA                          & 80.1 & 82.8 \\
DocVQA$_{\text{val}}$            & 89.5 & 88.2 \\
InfoVQA                          & 70.1 & 70.4 \\
OCRBench                         & 75.7 & 77.6 \\
AI2D$_{\text{test}}$             & 82.0 & 80.4 \\

\addlinespace[5pt]
\midrule
\multicolumn{3}{@{}l}{\textbf{Other Tasks}} \\
\addlinespace[2pt]
CountBenchQA                     & 86.0 & 84.2 \\
VL-RewardBench                   & 47.8 & 55.7 \\
V$^*$                            & 61.8 & 61.8 \\

\bottomrule
\end{tabular}
\end{table}
\endgroup

\begingroup

\clearpage
\thispagestyle{plain}
\section{Additional Qualitative Comparison}
\label{app:additional-qualitative-comparison}
\begin{center}
  \modelcomparisongrid{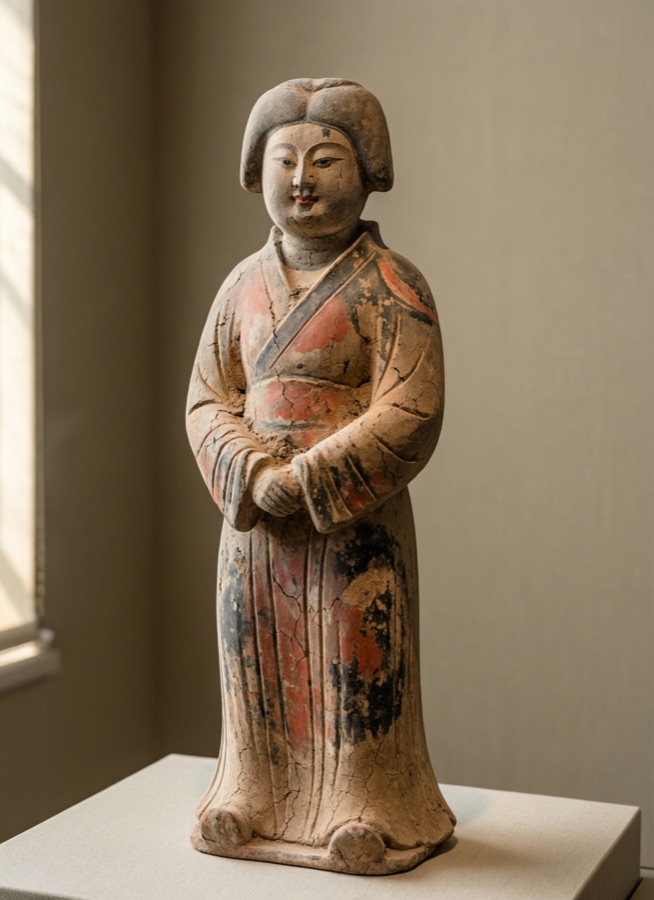}{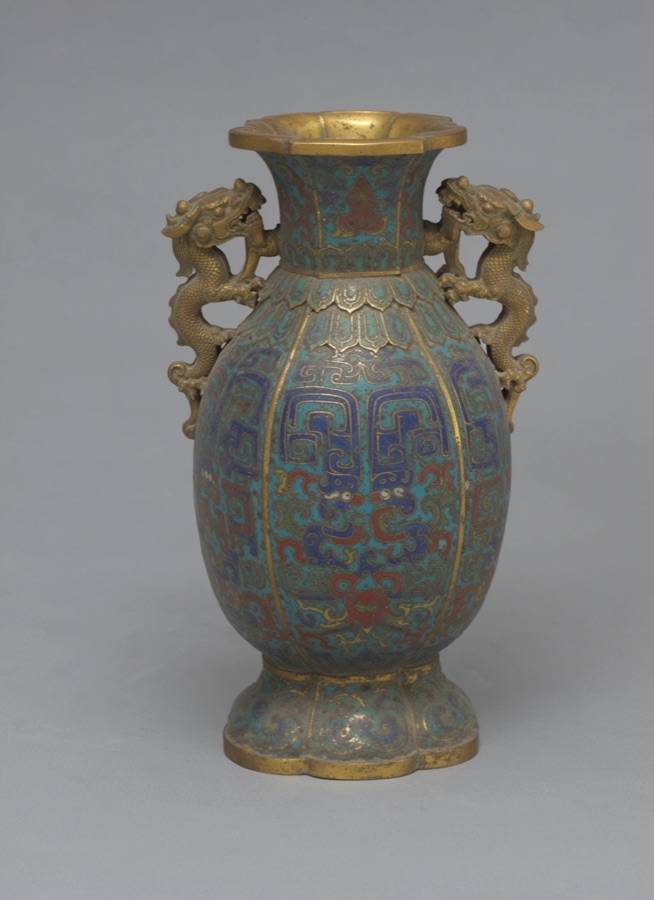}
\end{center}
\modelcomparisonpage{Additional Qualitative Comparison (Continued)}{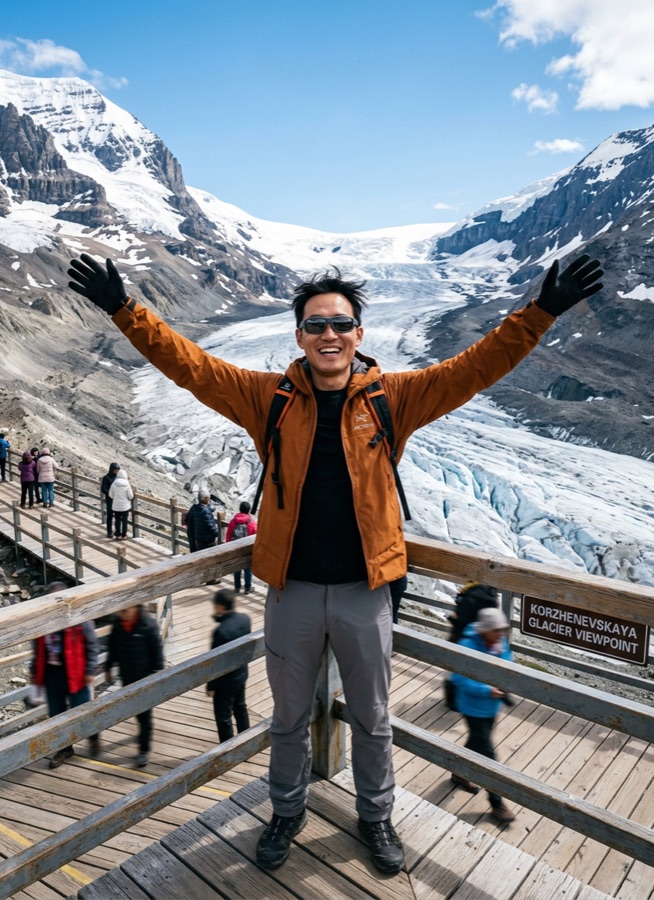}{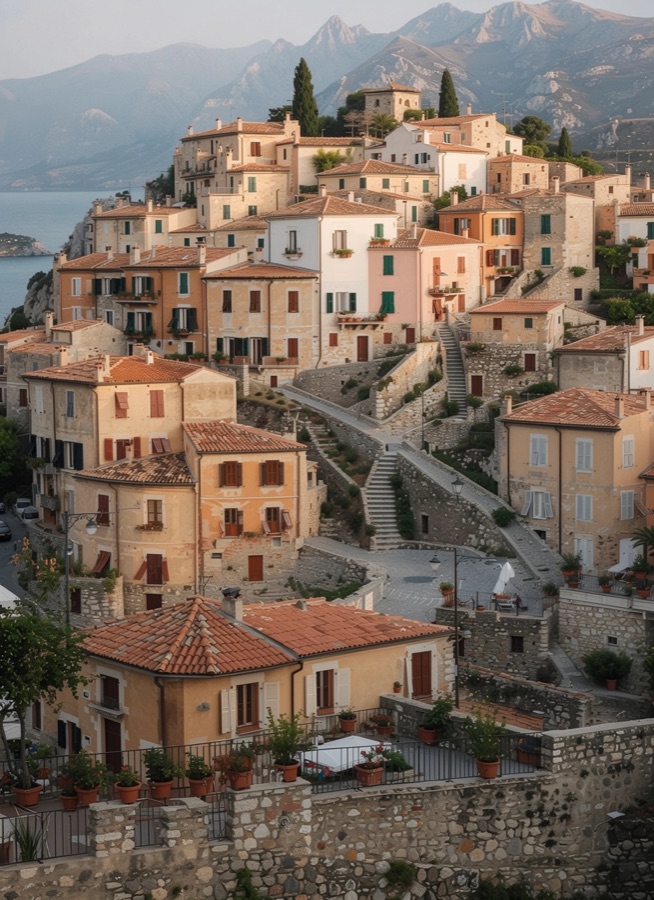}
\modelcomparisonpage{Additional Qualitative Comparison (Continued)}{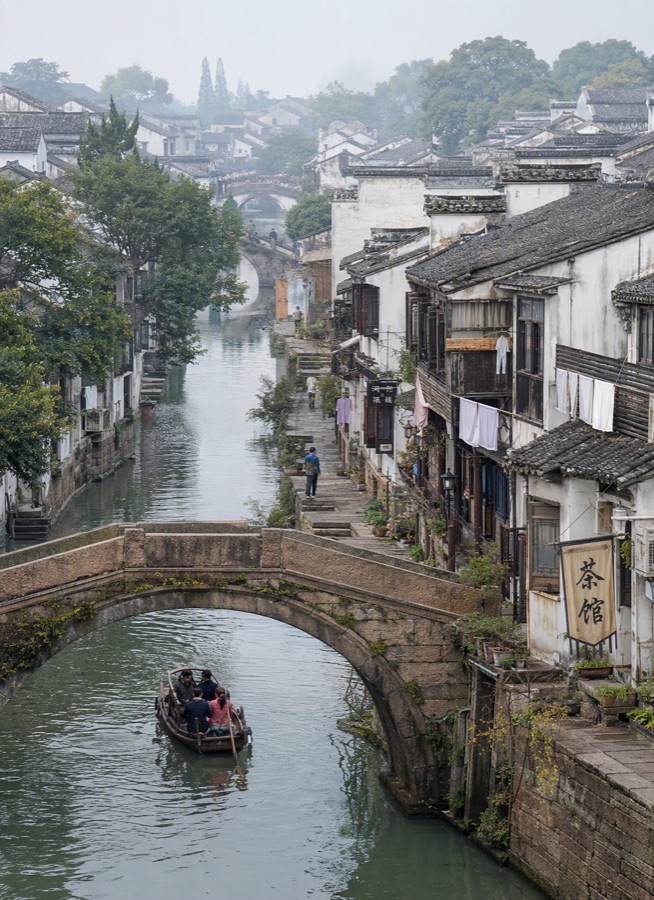}{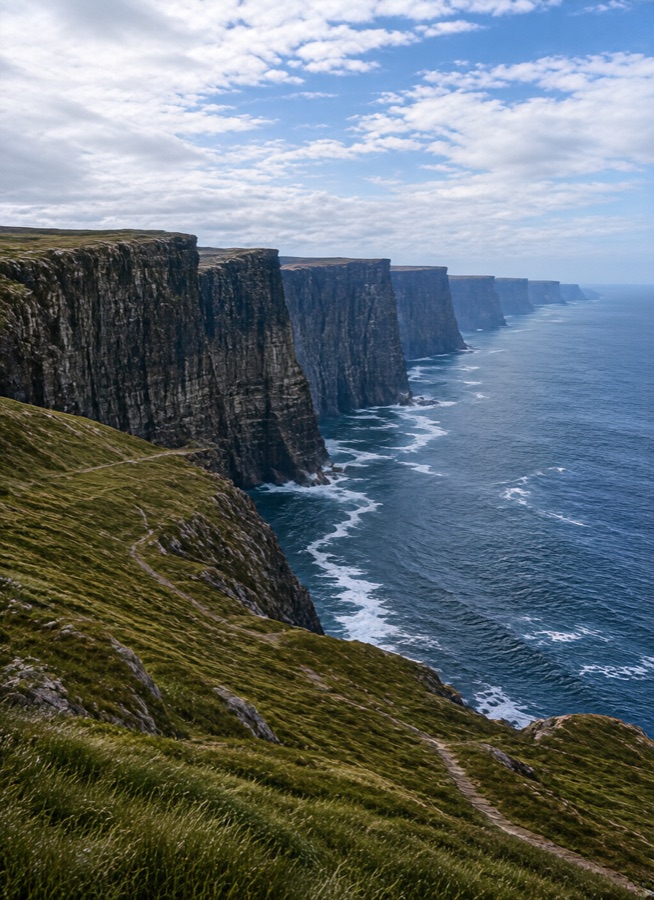}
\endgroup

\clearpage
\section{Image Captioning and Filtering}
\label{app:caption-generation-filtering}

For images selected for paired supervision, the captioner describes the main
subjects, attributes, actions, spatial relations, scene, and visual style based
only on the image.

After captioning, structural checks remove failed or incomplete outputs,
refusal responses, and degenerate repetitions. Qwen3.6-35B-A3B then assesses
both image quality and caption faithfulness from the image, caption, and
visible-text transcription. We remove samples containing watermark signals or
private information. We also reject captions that hallucinate visual content
or inaccurately describe or transcribe visible text.

\paragraph{Sample of Captioning Prompt.}
\begin{quote}\small
\begin{CJK*}{UTF8}{gbsn}
\textbf{Chinese Captioning Prompt.}

你是一位文本生成图片（T2I）模型的数据标注专家。你的目标是为图片生成忠实、具体、信息密度高、贴近真实用户输入习惯的中文训练标注。仅根据图片内容进行标注。

\textbf{detailed\_prompt 规则}

生成一段用于 T2I 训练的中文描述。描述应忠实、具体、信息密度高。不要写成鉴赏、总结或新闻稿。

覆盖以下要素（不要求固定顺序，自然组织语言即可）：
\begin{itemize}[leftmargin=16pt,itemsep=1pt,topsep=2pt]
  \item 主体：人物/物体/动物/食物/建筑/商品等，包括数量（仅在明确可辨时）、外观、颜色、材质、服装、姿态、动作
  \item 场景：室内/室外、地点类型、背景元素
  \item 构图与视角：特写、半身、全身、俯拍、仰拍、平视、近景、远景、居中、对称、留白、拼贴、分栏等，只在明显时描述
  \item 风格与媒介：实拍照片、产品摄影、插画、漫画、3D渲染、截图、UI设计、海报、信息图、表情包等，只在明显时描述
  \item 光线、色彩、材质：如暖黄灯光、冷色调、木质、金属质感、磨砂、透明等
  \item 可见文字（重要）：图中所有需要渲染的文字必须保留原文，不翻译、不改写，并用中文双引号“...”包裹，并标注位置和视觉属性（字号、颜色、字体粗细）。多行文字应逐行描述。如：顶部居中大号红色粗体”春节快乐”，底部右对齐小号白色细体”2026”。render\_text 字段另外做精确逐字转写用于校验
\end{itemize}

写作要求：
\begin{itemize}[leftmargin=16pt,itemsep=1pt,topsep=2pt]
  \item 中文，直接描述图中可见内容，开头方式应多样，可以从主体名词、视角描述、场景环境、媒介类型等不同角度切入
  \item 像是直接告诉生成模型"要画什么"的指令，而不是对图片的解说或鉴赏
  \item 只写有视觉证据支撑的内容，不编造
  \item 长度应随画面信息量变化：简单主体图可短至 150 字，复杂场景/密集文字图应长至 500-800 字，不要所有图都写成相近长度
  \item detailed\_prompt 和 short\_prompt 作为描述性文本，不使用 emoji、markdown 格式或换行符
  \item render\_text 是精确转写字段，完全如实还原图上文字的原始内容（包括其中的 emoji、符号、格式标记等）
\end{itemize}

\textbf{short\_prompt 规则}

模拟一个真实用户在文生图工具中会输入的简短提示词：
\begin{itemize}[leftmargin=16pt,itemsep=1pt,topsep=2pt]
  \item 只关注画面中最主要的几个要素，不需要覆盖全部内容
  \item 像真实用户会输入的提示词，简洁直接
  \item 大致10-50 字，不加句号
  \item 例："穿红裙子的女生在海边""白底产品图 蓝色运动鞋""赛博朋克城市夜景 霓虹灯"
\end{itemize}

\textbf{render\_text 规则}
\begin{itemize}[leftmargin=16pt,itemsep=1pt,topsep=2pt]
  \item 逐字转写画面中清晰可见的文字，保留原始语言
  \item 看不清或模糊的不写，无文字给空列表
\end{itemize}

\textbf{输出格式（仅输出 JSON，无任何额外文字）}

\{
\quad \texttt{"tag"}: [\texttt{"1-3个类别标签，优先从：fashion/portrait/food/scenery/product/beauty/home/pet/car/fitness/parenting/poster/ui/infographic/text\_doc/meme/tutorial/logo/typography/packaging/other 中选择"}],\\
\quad \texttt{"render\_text"}: [\texttt{"逐字转写的文字片段"}],\\
\quad \texttt{"short\_prompt"}: \texttt{"模拟用户输入的简短提示"},\\
\quad \texttt{"detailed\_prompt"}: \texttt{"客观信息密集的中文长描述"}
\}

\textbf{本批次额外硬性要求}
\begin{itemize}[leftmargin=16pt,itemsep=1pt,topsep=2pt]
  \item 本条必须输出中文。
  \item 图中可见文字必须保留原文，不翻译、不改写、不归一化。
  \item 在 detailed\_prompt 中提到任何图中文字时，必须使用中文双引号“...”包裹。
  \item render\_text 字段必须逐字转写图中文字原文；不要翻译。
  \item 仅输出 JSON，不要输出解释文字。
\end{itemize}
\end{CJK*}

\medskip
\textbf{English Captioning Prompt.}

You are a data annotation expert for text-to-image (T2I) models. Your task is
to produce faithful, specific, information-dense English training annotations
that resemble prompts written by real users. Base every annotation only on
the image content.

\textbf{detailed\_prompt rules}

Write an English description for T2I training. The description must be
faithful, specific, and information-dense. Do not write an aesthetic review,
summary, or news report.

Cover the following elements when they are visible. Use a natural order rather
than a fixed template.
\begin{itemize}[leftmargin=16pt,itemsep=1pt,topsep=2pt]
  \item Subjects: people, objects, animals, food, buildings, products, and
        other main elements, including count when clearly discernible,
        appearance, color, material, clothing, pose, and action.
  \item Scene: indoor or outdoor setting, location type, and background
        elements.
  \item Composition and viewpoint: close-up, half-body or full-body framing,
        overhead or low-angle view, eye-level view, near or distant view,
        centered or symmetric composition, negative space, collage, or column
        layout, but only when visually evident.
  \item Style and medium: photograph, product photography, illustration,
        comic, 3D rendering, screenshot, UI design, poster, infographic, meme,
        or another clearly visible medium.
  \item Lighting, color, and material: for example, warm lighting, a cool
        palette, wood, metal, matte surfaces, or transparency.
  \item Visible text: preserve all text that must be rendered in its original
        wording without translation or rewriting. Enclose it in double
        quotation marks and describe its position and visual properties,
        including size, color, and weight. Describe multiple lines separately.
        For example: large red bold text ``SUMMER SALE'' at the top center and
        small white light-weight text ``2026'' at the bottom right. The
        render\_text field separately records an exact transcription for
        verification.
\end{itemize}

Writing requirements:
\begin{itemize}[leftmargin=16pt,itemsep=1pt,topsep=2pt]
  \item Write in English and directly describe visible content. Vary the
        opening by starting from the main subject, viewpoint, setting, or
        medium as appropriate.
  \item Write like a direct instruction telling a generation model what to
        draw, rather than commentary or an aesthetic review.
  \item Include only details supported by visual evidence. Do not invent
        content.
  \item Let the length follow the amount of visible information. Use a shorter
        description for a simple subject and a longer one for a complex scene
        or text-dense image. Do not force all captions to similar lengths.
  \item detailed\_prompt and short\_prompt are descriptive text. Do not use
        emoji, Markdown formatting, or line breaks in either field.
  \item render\_text is a verbatim transcription field. Preserve the original
        visible text, including emoji, symbols, and formatting marks.
\end{itemize}

\textbf{short\_prompt rules}

Simulate a short prompt that a real user might enter into a T2I system.
\begin{itemize}[leftmargin=16pt,itemsep=1pt,topsep=2pt]
  \item Focus only on the most important elements rather than covering the
        entire image.
  \item Keep it concise, direct, and natural for a user prompt.
  \item Use roughly 5--30 words and do not add a final period.
  \item Examples: ``a woman in a red dress by the sea'', ``blue running shoe
        product photo on a white background'', ``cyberpunk city at night with
        neon lights''.
\end{itemize}

\textbf{render\_text rules}
\begin{itemize}[leftmargin=16pt,itemsep=1pt,topsep=2pt]
  \item Transcribe clearly legible text verbatim and preserve its original
        language.
  \item Omit illegible or blurred text. Return an empty list if no text is
        visible.
\end{itemize}

\textbf{Output format (JSON only, with no additional text)}

\{
\quad \texttt{"tag"}: [\texttt{"1--3 category labels, preferably selected from:
fashion/portrait/food/scenery/product/beauty/home/pet/car/fitness/parenting/
poster/ui/infographic/text\_doc/meme/tutorial/logo/typography/packaging/other"}],\\
\quad \texttt{"render\_text"}: [\texttt{"verbatim visible-text fragment"}],\\
\quad \texttt{"short\_prompt"}: \texttt{"short prompt resembling a user request"},\\
\quad \texttt{"detailed\_prompt"}: \texttt{"objective, information-dense English description"}
\}

\textbf{Additional mandatory requirements for this batch}
\begin{itemize}[leftmargin=16pt,itemsep=1pt,topsep=2pt]
  \item The output for this item must be in English.
  \item Preserve visible text in its original wording. Do not translate,
        rewrite, or normalize it.
  \item When detailed\_prompt mentions visible text, enclose it in double
        quotation marks.
  \item The render\_text field must transcribe visible text verbatim. Do not
        translate it.
  \item Output JSON only, with no explanatory text.
\end{itemize}
\end{quote}

\paragraph{Unified Filtering Prompt.}
\begin{quote}\small
\begin{CJK*}{UTF8}{gbsn}
你是 T2I 训练数据质量审核员。根据图片内容，判断以下四个字段并输出 JSON。

1. \texttt{has\_private\_info} (bool)

判断图中是否包含个人私密信息。
\begin{itemize}[leftmargin=16pt,itemsep=1pt,topsep=2pt]
  \item true：身份证、银行卡、证件照（含真人面部的证件）、私人聊天中的手机号或微信号、快递单上的个人姓名+地址+电话。
  \item false：餐厅地址、普通人像照片。
\end{itemize}
判断标准：是否为个人非公开的可识别信息。

2. \texttt{has\_watermark} (bool)

看图判断：图片上是否有社交媒体平台叠加的水印（如角落的平台logo、半透明的账号ID等）。
\begin{itemize}[leftmargin=16pt,itemsep=1pt,topsep=2pt]
  \item true：图上可见平台水印
  \item false：图上没有平台水印
\end{itemize}

3. \texttt{rt\_usable} (enum: yes / partial / no / na)

对比图片中实际可见的文字与 render\_text 的转写内容，判断转写质量。
\begin{itemize}[leftmargin=16pt,itemsep=1pt,topsep=2pt]
  \item yes：转写与图上实际文字一致，可直接用作渲染目标。
  \item partial：大部分正确但有少量错字或遗漏。
  \item no：存在严重错字或编造了图上不存在的文字。
  \item na：图上本身没有文字。
\end{itemize}

4. \texttt{caption\_faithfulness} (enum: faithful / minor\_issue / hallucinated)

对比图片内容与 detailed caption，判断描述是否忠实。
\begin{itemize}[leftmargin=16pt,itemsep=1pt,topsep=2pt]
  \item faithful：描述与画面一致。
  \item minor\_issue：存在小偏差（颜色、数量、位置略有出入）但整体正确，不影响训练。
  \item hallucinated：描述中包含图中明确不存在的物体、人物或文字。
\end{itemize}

只输出 JSON，不输出任何其他文字。

\textbf{当前样本}

short: \texttt{\{short\}}\\
detailed: \texttt{\{detailed\}}\\
render\_text: \texttt{\{render\_text\}}
\end{CJK*}
\end{quote}

\clearpage
\section{Prompts Used in The Report}
\label{app:prompts-used-in-report}

This section provides the prompts for the text-to-image examples presented in
the report. We retain their original wording and ordering for reproducibility.

\begingroup
\small
\begin{CJK*}{UTF8}{gbsn}
  \begin{enumerate}[leftmargin=*,label=\textbf{Prompt \#\arabic*},itemsep=0.8em]
    \item 一张城市公园中的自然街拍式人像摄影，拍摄于阴天之后刚刚放晴的下午。一位年轻东亚女性沿着树木茂密的步道缓慢行走，人物约占画面高度的一半，身体略微侧向镜头，刚好在经过一棵大树时回头看向身后，动作自然随意。她留着黑色中长发，头发整体柔顺，但被潮湿空气影响后有轻微蓬松和细碎毛躁，耳边有几缕头发贴近脸颊。脸型柔和偏圆，下颌收束自然，并非尖细瓜子脸。眉毛颜色较深，毛流完整，眉峰很轻。眼睛偏细长，眼睑厚度真实，瞳孔深棕色，视线直接落向镜头附近，但并不是刻意凝视，带有一点自然的好奇感。下眼睑有轻微阴影，鼻翼附近能看到非常淡的泛红。鼻梁柔和自然，鼻尖略圆。嘴唇厚薄适中，嘴角自然放松，唇纹细腻。皮肤并不追求无瑕，面颊有轻微肤色变化，额头和鼻梁有真实的柔和高光。她穿一件长度到膝盖附近的深绿色棉质风衣，衣服没有完全扣起，里面是一件灰白色薄针织上衣，下身搭配深蓝色牛仔裤。风衣面料稍厚，肩部、手肘和腰部形成明显自然褶皱。背景是公园内层层叠叠的树木、长椅、潮湿的石板路和低矮灌木，部分叶片仍挂着雨珠。远处有几个散步的人影，但经过景深处理后十分柔和。雨后阳光从树木之间穿过，形成局部明暗变化，在人物头发和肩膀上留下不规则亮边。整体像一张真实城市生活摄影。
\par\smallskip\noindent\textit{English translation.} A natural, candid street-style portrait photograph in a city park, taken on an afternoon when the sky has just cleared after overcast weather. A young East Asian woman walks slowly along a densely wooded path. She occupies about half the height of the frame, with her body angled slightly toward the camera, and turns to look back just as she passes a large tree; her movement is natural and casual. She has medium-length black hair that is generally smooth, but the humid air has made it slightly voluminous with fine flyaways; a few strands near her ears cling close to her cheeks. Her face is soft and somewhat round, with a naturally tapering jaw rather than a sharply pointed, V-shaped face. Her eyebrows are dark, with complete, clearly visible hair growth and only a very subtle arch. Her eyes are relatively long and narrow, with realistically thick eyelids and dark-brown irises. Her gaze falls directly near the camera, but she is not staring deliberately; it carries a hint of natural curiosity. There are slight shadows beneath her lower eyelids and extremely faint redness near the sides of her nose. Her nasal bridge is soft and natural, and the tip of her nose is slightly rounded. Her lips are of medium fullness, the corners of her mouth are naturally relaxed, and the fine lip lines are visible. Her skin is not intended to look flawless: there are slight variations in complexion across her cheeks and realistic, soft highlights on her forehead and nasal bridge. She wears a dark-green cotton trench coat reaching to around her knees. It is not fully buttoned, revealing a thin grayish-white knitted top underneath, paired with dark-blue jeans. The trench-coat fabric is somewhat thick, forming clearly visible, natural folds at the shoulders, elbows, and waist. The background contains layered park trees, benches, a damp flagstone path, and low shrubs, with raindrops still hanging from some leaves. Several distant walkers are visible, but are rendered very softly by the depth of field. Post-rain sunlight filters through the trees, creating localized changes between light and shade and leaving irregular rim highlights on the woman's hair and shoulders. The overall image should resemble an authentic photograph of everyday urban life.

\item 一张写实风格、像是手机后置镜头拍摄的、光线柔和的室内人像照片，捕捉一位年轻白人女性的温馨瞬间。画面主体位于中央，女性留着浓密波浪长发，发色呈现温暖的赤褐色，显得通透且富有光泽。女性面部特征清晰，皮肤白皙，鼻梁和脸颊上分布着明显的雀斑。女性面带灿烂的微笑，露出洁白整齐的牙齿，眼神温柔而自然地注视着镜头。她身穿一件白色长袖衬衫，外搭一件浅灰色针织背心，下身穿着一条淡粉色长裙。她左手手腕上佩戴一只银色金属表带手表，站在桌旁或坐在桌后，姿态自然端庄，用一只手指向相框，起到展示相框内容的作用，动作自然、清晰可见。 桌子上放置着一个巨大的长方形相框，相框位于画面视觉中心，尺寸非常醒目，正面朝向镜头展示。相框风格精美雅致，边框具有细腻的装饰感与真实材质质感。相框中的内容清晰可见，内部文字为超大号粗体字体，醒目、居中、完整展示。相框内清晰写着： “LLaDA-Image is Released!”。文字占据相框内部主要视觉区域，字形工整清楚，边缘锐利，清晰易读，不模糊、不变形、不被遮挡。 画面左侧前景放置一个透明玻璃花瓶，里面插着一束干枯的粉色玫瑰花，花茎细长，叶片呈灰绿色，为画面增添复古而自然的质感。背景是一面纯白色墙壁，没有任何装饰，以突出人物主体和桌上的相框。桌面呈现出深色且有光泽的质感，为深色大理石表面，并反射微弱柔和的光线。 整体色调以柔和的灰色、粉色和白色为主，营造出温暖、宁静、自然的氛围。光线均匀柔和，没有强烈阴影，属于典型的室内自然光或柔光摄影风格。整体画面呈现真实、细腻、温馨的人像摄影质感，人物清晰，对焦准确，浅景深适中，主体突出，构图重点明确落在女性与桌上巨大相框的互动展示上。
\par\smallskip\noindent\textit{English translation.} A realistic, softly lit indoor portrait photograph that looks as though it was taken with a phone's rear camera, capturing a warm moment with a young white woman. The subject is centered in the frame. The woman has thick, long, wavy hair in a warm auburn shade that appears translucent and lustrous. Her facial features are clear; she has fair skin and prominent freckles across the bridge of her nose and cheeks. She wears a radiant smile that reveals neat, white teeth, and looks gently and naturally at the camera. She is dressed in a white long-sleeved shirt under a light-gray knitted vest, with a pale-pink long skirt. She wears a watch with a silver metal band on her left wrist. Standing beside the table or sitting behind it in a natural, poised posture, she points toward the picture frame with one hand to present its contents; this gesture should look natural and be clearly visible. A huge rectangular picture frame sits on the table at the visual center of the image. Its conspicuous size immediately attracts attention, and its front faces the camera. The frame is refined and elegant, with delicately decorative detailing and a realistic material texture. Its contents are clearly visible, with exceptionally large, bold text displayed prominently, centrally, and in full. The frame clearly reads: “LLaDA-Image is Released!” The text occupies most of the visual area inside the frame; the lettering is neat and clear, with sharp edges, easy to read, neither blurred nor distorted, and not obscured. In the left foreground, a transparent glass vase holds a bunch of dried pink roses with slender stems and gray-green leaves, adding a vintage yet natural texture to the scene. The background is a completely plain white wall with no decoration, emphasizing the woman and the picture frame on the table. The tabletop is dark and glossy, made of dark marble, and reflects faint, soft light. The overall palette is dominated by soft gray, pink, and white, creating a warm, tranquil, and natural atmosphere. The illumination is even and gentle, with no harsh shadows, characteristic of indoor natural-light or soft-light photography. The overall image should have the realistic, delicate, and warm quality of portrait photography, with the woman clearly rendered, accurate focus, a moderate shallow depth of field, a prominent subject, and the composition explicitly centered on the woman's interaction with and presentation of the huge picture frame on the table.

\item 一张带有轻复古感的小红书风格街拍人像摄影作品，拍摄于老城区的咖啡馆外墙边。主体是一位二十多岁的年轻东亚女生，靠着一面浅米色旧墙站立，一只手拿着拍立得相机，另一只手提着纸袋。她的脸型偏柔和长椭圆，额头和下颌过渡自然，颧骨位置不高。黑色中长发自然散落，发尾有轻微弧度，头顶发丝细软，耳边有两缕碎发。眉毛自然偏浓，毛流清楚。眼睛偏细长，双眼皮较浅，瞳孔看向手里的相机，神态带一点安静和专注。眼下阴影很轻，右侧眼角附近有一颗很淡的小痣。鼻梁顺滑，鼻尖略圆，鼻翼附近保留自然毛孔。嘴唇厚薄适中，唇色接近自然豆沙色，唇纹清晰。皮肤偏哑光，有轻微肤色不均和真实质感，不显油润。她穿一件奶白色针织短袖开衫，下身搭配浅咖色高腰半裙，脚穿棕色乐福鞋和白色短袜，整体穿搭年轻又有一点文艺复古感。背景是旧墙、木窗、咖啡馆小圆桌和门口黑板，黑板上面写着“今日特调”四个字，下面写着“冰美式”三个字。傍晚柔和天光与店内暖光混合，整体画面轻松、时髦、青春，具有很强的社交平台分享感。
\par\smallskip\noindent\textit{English translation.} A Xiaohongshu-style street portrait with a lightly vintage character, photographed beside the exterior wall of a cafe in an old city district. The subject is a young East Asian woman in her twenties, standing against an aged, light-beige wall. She holds an instant camera in one hand and carries a paper bag in the other. Her face has a soft, elongated oval shape, with natural transitions across the forehead and jaw and cheekbones that are not set high. Her medium-length black hair falls naturally, with a slight curve at the ends, fine and soft strands at the crown, and two wispy locks beside her ears. Her eyebrows are naturally rather full, with individual hairs clearly visible. Her eyes are relatively long and narrow, with shallow double-eyelid creases, and her pupils look toward the camera in her hand; her expression is quiet and focused. There are only very faint shadows beneath her eyes and a subtle small mole near the outer corner of her right eye. Her nasal bridge is smooth, the tip of her nose is slightly rounded, and natural pores remain visible around the sides of her nose. Her lips are of medium fullness, with a natural muted-rose color and clearly visible lip lines. Her skin has a mostly matte appearance, with slight unevenness in tone and a realistic texture, rather than looking oily. She wears a cream-colored, short-sleeved knitted cardigan with a light-brown high-waisted midi skirt, brown loafers, and short white socks. The overall outfit feels youthful with a slightly artistic, vintage quality. The background contains the old wall, a wooden window, a small round cafe table, and a blackboard by the entrance. The blackboard must show the four Chinese characters “今日特调” on the upper part and the three Chinese characters “冰美式” beneath them. Soft evening daylight mixes with warm light from inside the cafe. The overall image is relaxed, fashionable, and youthful, with a strong sense of being made for sharing on social media.

\item 一张垂直构图的街头纪实摄影作品，拍摄于清晨刚开始营业的菜市场。画面主体是一位中年男性摊主，皮肤被阳光晒得略深，头发短而微乱，穿着浅灰色短袖T恤和深蓝色防水围裙，正低头整理摊位上的蔬菜。摊位上整齐摆放着绿色的黄瓜、红色番茄、紫色茄子、带泥的胡萝卜和成捆的小葱，部分蔬菜表面还带有细小水珠，显得新鲜。前景左下角是一只装着生菜的塑料筐，边缘略微虚化。画面右侧可以看到一位戴草帽的顾客正在挑选青椒，只露出半个身影。背景中是一排排简易金属支架和蓝绿色遮阳棚，棚下悬挂着白色灯泡和写有价格的手写纸牌，上面写着“小番茄5元/斤”几个字。清晨的自然光从市场入口处斜射进来，在蔬菜表面形成明亮高光，同时保留阴影区域的真实层次。整体画面色彩丰富但不过度饱和，呈现出普通城市早晨的烟火气息。
\par\smallskip\noindent\textit{English translation.} A vertically composed street-documentary photograph taken at a produce market just as it begins operating in the early morning. The main subject is a middle-aged male stallholder with slightly sun-darkened skin and short, mildly tousled hair. Wearing a light-gray short-sleeved T-shirt and a dark-blue waterproof apron, he looks down while arranging the vegetables on his stall. Neatly displayed on the stall are green cucumbers, red tomatoes, purple eggplants, mud-streaked carrots, and bundles of scallions. Tiny droplets of water remain on some of the vegetables, making them appear fresh. In the lower-left foreground is a plastic crate filled with lettuce, its edges slightly out of focus. On the right side of the frame, a customer in a straw hat is selecting green peppers, with only half of their figure visible. The background contains rows of simple metal supports and blue-green awnings. White light bulbs and handwritten paper price signs hang beneath the awnings; one sign must read “小番茄5元/斤”. Early-morning natural light enters diagonally from the market entrance, forming bright highlights on the vegetables while preserving realistic tonal depth in the shadowed areas. The overall image is colorful but not excessively saturated, conveying the lively, everyday atmosphere of an ordinary urban morning.

\item 一张傍晚海边栈道上的旅行合照摄影作品，主体是两位年轻东亚男性和一位年轻东亚女性站在木栈道护栏前拍合照。三人并排而立，身体朝向镜头，风把头发和衣角轻轻吹起，整体像傍晚散步时请旁人帮忙拍下的一张照片。左侧男生脸型偏长圆，黑色短发被海风吹得稍微凌乱，眉毛浓度中等，眼睛偏长，鼻尖略圆，嘴唇较薄，皮肤偏哑光。他穿浅卡其色风衣和深灰色长裤。中间女生脸型偏柔和鹅蛋形，黑色长发扎成低马尾，耳边有两缕细碎发丝，眼睛偏细长，鼻梁柔和，嘴唇略丰满，皮肤自然干净。她穿白色针织上衣和浅蓝色牛仔裤。右侧男生黑色短发较整齐，眉毛较粗，眼睛为内双，鼻梁中等高度，嘴角轻微上扬。他穿深蓝色夹克和米白色T恤。背景是灰蓝色海面、远处灯塔、礁石和被风吹动的低矮灌木。阴天傍晚的天光柔和均匀，整体安静、真实、带旅行感。
\par\smallskip\noindent\textit{English translation.} A travel group photograph taken on a seaside boardwalk in the evening. The subjects are two young East Asian men and one young East Asian woman posing for a group photo in front of the wooden boardwalk railing. The three stand side by side facing the camera as the wind gently lifts their hair and the edges of their clothing. The image should look like a photograph taken by a passerby for them during an evening stroll. The man on the left has a somewhat elongated, rounded face; short black hair made slightly messy by the sea breeze; eyebrows of medium density; relatively long eyes; a slightly rounded nose tip; thin lips; and mostly matte skin. He wears a light-khaki trench coat and dark-gray trousers. The woman in the center has a soft, oval face and long black hair tied in a low ponytail, with two fine, wispy strands beside her ears. Her eyes are relatively long and narrow, her nasal bridge is soft, her lips are slightly full, and her skin looks naturally clear. She wears a white knitted top and light-blue jeans. The man on the right has fairly neat short black hair, thick eyebrows, tapered double eyelids, a nasal bridge of medium height, and slightly upturned corners of the mouth. He wears a dark-blue jacket and an off-white T-shirt. The background consists of a gray-blue sea, a distant lighthouse, reefs, and low shrubs stirred by the wind. The overcast evening light is soft and even, and the overall image feels quiet, authentic, and evocative of travel.

\item 一张垂直构图的室内中景照片，捕捉了一对年轻东亚情侣在餐厅庆祝生日的温馨瞬间。画面中心，一位留着黑色长直发的年轻女子身穿深灰色毛呢大衣，内搭米白色高领针织衫，面带温柔微笑注视镜头，双手托举着一个精致的粉色圆形生日蛋糕。在她身后右侧，一位留着黑色短发的年轻男子身穿浅灰色连帽卫衣，左臂亲密地环绕在女子肩后，同样微笑着看向镜头。女子手中的蛋糕表面覆盖着粉色奶油，装饰有数只金边白翅的蝴蝶造型糖饰、大小不一的白色珍珠状装饰球，以及一支正在燃烧的细长生日蜡烛。前景的深色木质圆桌上放置着一盘意大利面，旁边摆放着两个透明玻璃水杯和一个插着燃烧白色粗蜡烛的金色烛台。背景中可见铺着深绿色桌布的餐桌、一瓶红酒、一瓶插着粉色花朵的小花瓶以及一盆绿植，墙面采用木质护墙板装饰。整体光线柔和温暖，营造出浪漫且充满生活气息的氛围。
\par\smallskip\noindent\textit{English translation.} A vertically composed, indoor medium shot capturing a warm moment as a young East Asian couple celebrates a birthday in a restaurant. At the center of the frame, a young woman with long, straight black hair wears a dark-gray wool coat over an off-white turtleneck sweater. Smiling gently and looking at the camera, she holds up an elegant, round pink birthday cake with both hands. Behind her on the right, a young man with short black hair wears a light-gray hoodie. His left arm is wrapped affectionately behind the woman's shoulders, and he likewise smiles toward the camera. The cake in the woman's hands is covered with pink frosting and decorated with several butterfly-shaped sugar ornaments featuring gold-edged white wings, white pearl-like decorative balls in varying sizes, and one slender birthday candle that is burning. On the dark wooden round table in the foreground is a plate of pasta, beside two transparent glass tumblers and a gold candlestick holding a lit, thick white candle. Visible in the background are a dining table covered with a dark-green tablecloth, a bottle of red wine, a small vase containing pink flowers, and a potted green plant; the walls are finished with wooden wainscoting. The overall lighting is soft and warm, creating a romantic atmosphere rich in the feeling of everyday life.

\item 一张图书馆儿童区里的自然人像摄影作品，主体是一位七岁左右的东亚小女孩，坐在靠窗的矮桌前翻看一本绘本。她的脸型偏长圆，额头平整，面颊仍然饱满，下颌线柔和，下巴略圆。黑色长发扎成松散低马尾，额前留着少量细碎刘海，耳边有两缕细发自然垂下。眉毛细而自然，眉峰很轻。眼睛偏细长，双眼皮较浅，深棕色瞳孔向下看着书页，睫毛自然，不浓密夸张。眼下有极轻的阴影和淡淡卧蚕。鼻梁低而顺滑，鼻尖圆润。嘴唇小巧，上唇略薄，下唇稍厚，嘴角自然放松。皮肤整体自然偏哑光，面颊和鼻侧有非常轻微的肤色变化。她穿浅粉色针织开衫，内搭白色圆领T恤，下身为深灰色百褶裙。桌面上放着几支彩色铅笔和一本打开的绘本，封面写着“森林朋友”。背景是儿童书架、软垫和窗外树影，上午自然光柔和明亮，整体安静、真实。
\par\smallskip\noindent\textit{English translation.} A natural portrait photograph in the children's area of a library. The subject is an East Asian girl about seven years old, seated at a low table by a window and leafing through a picture book. Her face is somewhat elongated and rounded, with a smooth forehead, still-full cheeks, a soft jawline, and a slightly rounded chin. Her long black hair is tied into a loose low ponytail, with a small amount of fine, wispy fringe across her forehead and two thin strands falling naturally beside her ears. Her eyebrows are fine and natural, with a very subtle arch. Her eyes are relatively long and narrow, with shallow double-eyelid creases. Her dark-brown pupils look downward at the pages, and her eyelashes are natural rather than excessively thick or exaggerated. Beneath her eyes are extremely faint shadows and a subtle fullness of the lower eyelids. Her nasal bridge is low and smooth, and her nose tip is rounded. Her lips are small, with a slightly thin upper lip and a somewhat fuller lower lip, and the corners of her mouth are naturally relaxed. Her skin has an overall natural, mostly matte appearance, with extremely slight variations in complexion across the cheeks and sides of the nose. She wears a light-pink knitted cardigan over a white crew-neck T-shirt and a dark-gray pleated skirt. Several colored pencils and an open picture book rest on the tabletop; the cover must read “森林朋友”. The background contains children's bookshelves, cushions, and tree shadows visible outside the window. Soft, bright morning daylight illuminates the scene, and the overall image is quiet and authentic.

\item 一张居家环境中的儿童生活摄影作品，主体是一位四岁左右的小女孩，盘腿坐在客厅地毯上搭积木。她拥有柔和偏圆的脸型，额头饱满，颧骨几乎不明显，面颊圆润，下巴短而圆。深棕黑色短发长度到耳下，发尾自然内扣，额前留着略微参差的薄刘海，头顶有几根细软发丝翘起。眉毛细而浅，眉头毛流自然。眼睛为偏圆的杏仁形，双眼皮褶皱较浅，深棕色瞳孔正看向手里准备叠上的红色积木，神情认真。鼻梁较低，鼻尖圆润，鼻翼自然。嘴唇小巧，上唇略薄，下唇饱满一点，嘴巴轻轻张开，呈现孩子专注时不自觉的表情。皮肤呈自然偏哑光的细腻质感，面颊有轻微自然红润，额头和鼻侧没有明显油光。她穿浅米色针织上衣和浅绿色棉质长裤，脚上只穿白色袜子。地毯上散落着木质积木、绘本和一只毛绒兔子，背景是浅色布艺沙发、落地窗和绿植，午后自然光从窗边照入，整体温暖、真实、生活化。
\par\smallskip\noindent\textit{English translation.} A slice-of-life photograph of a child in a home setting. The subject is a girl about four years old, sitting cross-legged on the living-room carpet and building with blocks. She has a soft, somewhat round face, a full forehead, nearly imperceptible cheekbones, rounded cheeks, and a short, round chin. Her short, dark brown-black hair reaches just below her ears and naturally curves inward at the ends. A thin, slightly uneven fringe crosses her forehead, and several fine, soft strands stand up at the crown. Her eyebrows are fine and light, with natural hair growth at their inner ends. Her eyes are relatively round and almond-shaped, with shallow double-eyelid creases. Her dark-brown pupils are focused on the red block in her hand that she is about to stack, and her expression is serious and attentive. Her nasal bridge is low, her nose tip is rounded, and the sides of her nose look natural. Her lips are small, with a slightly thin upper lip and a somewhat fuller lower lip. Her mouth is gently open, showing the unconscious expression of a child absorbed in concentration. Her skin has a delicate, natural, mostly matte texture, with a faint natural flush in her cheeks and no obvious shine on her forehead or the sides of her nose. She wears a light-beige knitted top and light-green cotton trousers, with only white socks on her feet. Wooden blocks, picture books, and a plush rabbit are scattered across the carpet. The background contains a light-colored fabric sofa, a floor-to-ceiling window, and green plants. Afternoon natural light enters from the window, and the overall image feels warm, authentic, and true to everyday life.

\item 一张冬日傍晚游乐场里的儿童旧照片风格摄影作品，画面主体是一位五岁左右的东亚小男孩，站在旋转木马入口旁，一只手摆出剪刀手，另一只自然下垂，抬头看向镜头。孩子脸型偏圆，额头较宽，面颊柔软饱满，下巴短而圆。黑色短发细软，从毛线帽边缘露出来，额前有几缕细碎头发被冷风轻轻吹乱，但整体没有油亮感。眉毛颜色较浅，整体偏直。眼睛偏圆，双眼皮浅，深棕色瞳孔正对镜头，眼神里带有期待和一点兴奋，下眼睑有轻微卧蚕。鼻梁低而自然，鼻尖圆润，鼻翼略宽，因为天气寒冷，鼻尖和双颊带有一点淡淡的自然红润。嘴唇偏小，下唇稍厚，嘴巴轻轻张开，露出几颗儿童牙齿。皮肤呈自然偏哑光质感，面部受冷空气影响有柔和的红润变化。孩子穿一件米白色羽绒外套，内搭浅灰色高领针织衫，下身是浅蓝色牛仔背带长裤，脚上穿红白相间的运动鞋，头上戴着一顶深红色针织毛线帽，脖子上围着一条浅卡其色围巾。背景是冬日傍晚的旋转木马、暖黄色灯泡、略微褪色的彩灯装饰、铁栏杆和远处穿着厚外套、手里拿着气球的游客，入口灯牌上写着“儿童乐园”。地面是略显冰冷的深灰色水泥地，边缘还能看到几片被风吹到角落里的枯叶，空气中带有明显冬季傍晚的清冷感。整张画面像家庭相册中保存多年的冬日游乐场旧照片，暖黄色胶片偏色明显但不过度，灯光有轻微晕开效果，整体饱和度偏低，带有细腻颗粒、轻微扫描灰尘和一点边缘柔化，呈现真实自然的年代感。右下角显示橙红色日期戳“02 12 02”。
\par\smallskip\noindent\textit{English translation.} An old-photo-style image of a child at an amusement park on a winter evening. The main subject is an East Asian boy about five years old, standing beside the entrance to a carousel. He raises one hand in a V sign while the other hangs naturally, and looks up toward the camera. The child has a somewhat round face, a broad forehead, soft full cheeks, and a short, round chin. His fine, soft black hair peeks out from the edge of his knitted hat. Several wispy strands across his forehead are lightly tousled by the cold wind, but his hair has no overall oily sheen. His eyebrows are light in color and mostly straight. His eyes are relatively round, with shallow double-eyelid creases. His dark-brown pupils face the camera, and his gaze conveys anticipation and a little excitement; there is a slight fullness beneath his lower eyelids. His nasal bridge is low and natural, his nose tip is rounded, and the wings of his nose are slightly wide. Because of the cold weather, the tip of his nose and both cheeks show a faint, natural flush. His lips are small, with a somewhat thicker lower lip, and his mouth is gently open, revealing several baby teeth. His skin has a natural, mostly matte texture, with soft changes in facial redness caused by the cold air. The child wears an off-white down jacket over a light-gray turtleneck sweater, light-blue denim overalls, red-and-white sneakers, a dark-red knitted hat, and a light-khaki scarf around his neck. The background shows a carousel on a winter evening, warm-yellow light bulbs, slightly faded colored-light decorations, iron railings, and distant visitors wearing thick coats and holding balloons. The illuminated entrance sign must read “儿童乐园”. The ground is cold-looking dark-gray concrete, with several dead leaves visible along the edges where the wind has blown them into corners, and the air has the distinct chill of a winter evening. The entire image should resemble an old winter-amusement-park photograph preserved for many years in a family album. It has a pronounced but not excessive warm-yellow film color cast, slight halation around the lights, generally low saturation, fine grain, a little scanned dust, and slight edge softness, producing an authentic and natural period quality. An orange-red date stamp reading “02 12 02” appears in the lower-right corner.

\item 一张写实自然风光摄影作品，表现一片由雪山融水补给的钙华湖群。远处背景是一列雪山，山顶积雪明亮，靠近山脊的地方可见被风吹出的雪纹，山腰为灰黑色裸岩和零星深绿色针叶林。中景主体是多个彼此相连的浅水钙华池，池壁由乳白色和浅米色矿物沉积形成，轮廓弯曲自然。水体颜色非常丰富，从浅蓝、青绿、碧绿到接近透明的浅色过渡明显，部分水池底部可见金黄色、浅褐色的矿物沉积纹路。几道细流从高处雪山方向汇入池群，在池间形成小型跌水和白色泡沫。前景是一片湿润的浅色石灰质地面，上面有不规则水道、低矮苔藓、几簇黄绿色草丛以及被矿物染成浅褐色的细小枝干。画面左侧有一片松林从山脚延伸至湖边，右侧则是较开阔的山坡，分布着灌木和浅灰岩石。水面倒映天空与远山，但因为流水和风的影响并不完全平整。天空为高原地区常见的浅蓝色与大片白云组合，云影投射在部分池面和山坡上，形成清晰的明暗变化。整体画面包含雪山、钙华池、流水、林地、草坡和矿物地表，细节充分，真实而富有层次。
\par\smallskip\noindent\textit{English translation.} A realistic natural-landscape photograph depicting a group of travertine lakes fed by meltwater from snow-covered mountains. In the distant background is a range of snowy mountains with bright snow on their peaks. Wind-sculpted patterns in the snow are visible near the ridgelines, while the mountainsides consist of gray-black exposed rock and scattered dark-green coniferous forest. The principal midground subject is a series of interconnected, shallow travertine pools. Their walls are formed from milky-white and light-beige mineral deposits and have naturally curving outlines. The water displays a rich range of colors, with clearly visible transitions from light blue through cyan-green and emerald green to pale, nearly transparent tones. Golden-yellow and light-brown patterns of mineral deposits can be seen on the bottoms of some pools. Several narrow streams flow toward the pool system from the higher snowy mountains, creating small cascades and white foam between the pools. The foreground consists of damp, pale limestone terrain crossed by irregular channels, with low moss, several clumps of yellow-green grass, and fine twigs stained light brown by minerals. On the left side of the image, a pine forest extends from the mountain foot to the lakeshore, while the right side contains a more open slope scattered with shrubs and light-gray rocks. The water reflects the sky and distant mountains but is not perfectly smooth because of flowing water and wind. The sky combines the pale blue typical of high-altitude regions with broad white clouds. Cloud shadows fall across some pool surfaces and slopes, creating clearly defined variations between light and dark. The overall image includes snowy mountains, travertine pools, flowing water, woodland, grassy slopes, and mineral terrain, with abundant detail, realism, and rich spatial depth.

\item 一张写实风光摄影作品，展示一处喀斯特峡谷中的地下河出口。画面主体是一座巨大的石灰岩崖壁，岩壁中央开有一个半圆形巨型洞口，洞口内部昏暗，地下河从洞中流出，形成一条青绿色的河流向画面前方延伸。洞口周围岩壁呈灰白色与深灰色相间，表面布满溶蚀形成的凹凸纹理、裂隙和垂直水痕，局部附着深绿色苔藓和蕨类植物。中景河岸两侧生长着浓密植被，包括灌木、藤蔓、小乔木和大片阔叶植物，层次非常丰富。前景是一片湿润的卵石河滩，散布着浅灰色、褐色和青灰色圆石，石头之间积着浅浅水洼，反射着天空和周围绿植。河水表面在靠近洞口处较为平静，向前流动时逐渐出现细小波纹和局部白色浪花。画面右侧还能看到一股小瀑布从崖壁裂缝中落下，汇入主河道。天空只在峡谷上方露出一条狭窄的浅蓝色区域，其余大部分被高耸崖壁和树冠遮挡。散射光从峡谷上方洒入，使岩石、水流和植物都保有真实质感。整体氛围清凉、潮湿、结构复杂，充分体现喀斯特峡谷和地下河系统的地貌特征。
\par\smallskip\noindent\textit{English translation.} A realistic landscape photograph showing the outlet of an underground river in a karst canyon. The main subject is an immense limestone cliff with a gigantic semicircular cave opening at its center. The interior of the cave is dim, and an underground river flows out from it, forming a blue-green river that extends toward the foreground. Around the cave opening, the cliff alternates between gray-white and dark-gray rock. Its surface is covered with uneven dissolution textures, fissures, and vertical water stains, with dark-green moss and ferns attached in places. Dense vegetation grows along both midground riverbanks, including shrubs, vines, small trees, and broad expanses of broadleaf plants, creating exceptionally rich layering. The foreground is a damp pebble beach scattered with rounded stones in light gray, brown, and blue-gray. Shallow puddles collect between the stones, reflecting the sky and surrounding greenery. The river surface is relatively calm near the cave opening; as the water flows forward, fine ripples and localized white splashes gradually appear. On the right side of the image, a small waterfall can also be seen dropping from a fissure in the cliff and joining the main river channel. Only a narrow strip of pale-blue sky is visible above the canyon; most of it is obscured by the towering cliffs and tree canopies. Diffuse light enters from above the canyon, allowing the rocks, flowing water, and plants to retain realistic textures. The overall atmosphere is cool, humid, and structurally complex, fully expressing the geomorphological characteristics of a karst canyon and underground-river system.

\item 一张高清风景摄影作品，展现了中国江南水乡的宁静景色，画面呈现出典型的“小桥流水人家”风貌。前景左下角是一丛茂密的绿色灌木，叶片间点缀着几片枯黄的叶子，增加了画面的层次感和自然气息。中景是一条宽阔平缓的河流，河水呈灰绿色，水面有细微的波纹，倒映着岸边的绿树和天空。河流左侧岸边错落分布着几座白墙黛瓦的传统中式民居，房屋依水而建，具有典型的江南建筑特色，屋檐下隐约可见红色的灯笼装饰。河流右侧是一条沿河铺设的蜿蜒步道，步道旁种植着高大茂盛的垂柳和其他阔叶树木，翠绿的枝叶繁茂，部分枝条垂向水面。在河流的中远景处，一座灰色的石拱桥横跨两岸，连接着左右两边的景色。背景是连绵起伏的青山，山上覆盖着浓密的森林，呈现出深浅不一的绿色，山峦在远处与灰白色的天空相接。整体光线柔和均匀，为阴天漫射光，没有强烈的阴影，营造出一种清幽、静谧且略带湿润的氛围。
\par\smallskip\noindent\textit{English translation.} A high-definition landscape photograph presenting the tranquil scenery of a Jiangnan water town in China and the archetypal character of “small bridges, flowing water, and waterside homes.” In the lower-left foreground is a dense cluster of green shrubs, with several withered yellow leaves interspersed among the foliage to add visual depth and a natural quality. The midground contains a broad, gently flowing river. Its water is gray-green, with fine ripples across the surface reflecting the green trees along the banks and the sky. Several traditional Chinese residences with white walls and dark-gray tiled roofs are irregularly distributed along the left bank. Built beside the water, the houses display typical Jiangnan architectural features, and red lantern decorations are faintly visible beneath their eaves. On the right side of the river, a winding path follows the bank. Tall, lush weeping willows and other broadleaf trees grow beside it, with abundant bright-green foliage and some branches hanging down toward the water. In the middle-to-far distance, a gray stone arch bridge spans the river, connecting the scenery on the two banks. The background consists of rolling green mountains covered in dense forest and rendered in varied shades of green; in the distance, the mountains meet a gray-white sky. The overall lighting is soft and even, consisting of diffuse overcast light with no harsh shadows, creating a secluded, serene, and slightly humid atmosphere.

        \item 一张垂直构图的写实摄影作品，捕捉了威尼斯狭窄运河中的经典景象。画面主体是一艘黑色的贡多拉（Gondola），正沿着水道向画面深处驶去。船尾站立着一位男性船夫，背对镜头，身穿经典的蓝白横条纹水手衫和深色长裤，头戴深色帽子，正手持长桨划船。贡多拉内部可见红色的软垫座椅和金色的金属装饰细节。运河水面呈深绿色，波光粼粼，清晰地倒映着两侧建筑的色彩和天空的光线，尤其是中央黄色建筑的金黄色倒影在水面上拉得很长。画面左侧是一排红褐色的多层建筑，墙面斑驳，显露出岁月的痕迹，窗户配有深绿色的百叶窗，二楼有一个带有白色栏杆的小阳台，墙根处可见裸露的红砖。画面右侧是另一栋红褐色建筑，墙面同样有剥落的痕迹，设有拱形的门洞和窗户。视线向远处延伸，可以看到一座白色的石拱桥横跨在运河上，连接着两岸。背景中央矗立着一栋醒目的黄色建筑，拥有白色的窗框和阳台，与周围暖色调的建筑形成对比。光线柔和，呈现出黄昏或清晨的暖色调，阳光从侧面照射在建筑上，营造出宁静、浪漫且充满历史感的氛围。
    \par\smallskip\noindent\textit{English translation.} A vertically composed, photorealistic photograph capturing a classic scene along a narrow canal in Venice. The main subject is a black gondola traveling along the waterway into the depth of the frame. A male gondolier stands at the stern with his back to the camera, wearing a classic blue-and-white horizontally striped sailor shirt, dark trousers, and a dark hat, and propelling the boat with a long oar. Inside the gondola, red upholstered seats and gold-colored metal decorative details are visible. The canal water is deep green and glitters with reflected light, clearly mirroring the colors of the buildings on both sides and the light from the sky; in particular, the golden reflection of the yellow building in the center stretches far across the water. On the left is a row of reddish-brown multistory buildings whose mottled walls show the passage of time. Their windows have dark-green shutters; a small second-floor balcony has white railings, and exposed red brick can be seen near the base of the wall. On the right is another reddish-brown building with similarly peeling walls, arched doorways, and arched windows. Looking farther into the distance, a white stone arch bridge spans the canal and connects the two banks. An eye-catching yellow building with white window frames and balconies rises in the center of the background, contrasting with the surrounding warm-toned architecture. The light is soft and warmly toned, suggesting dusk or early morning. Side lighting falls across the buildings, creating a tranquil, romantic atmosphere rich in historical character.

    \item 一张写实城市风光摄影作品，拍摄于雨后刚刚放晴的傍晚。画面从一处较高的人行天桥向下俯视，前景是被雨水打湿的金属栏杆和几片粘在地面的梧桐叶，栏杆表面残留细小水珠。中景是一条宽阔城市主干道，深灰色柏油路面仍然湿润，车道上的白色标线、公交车和私家车都在积水中形成拉长的倒影，路口处有骑自行车和撑伞的行人正在等待信号灯。道路两侧是高度不同的写字楼、住宅楼和沿街商铺，玻璃幕墙反射着雨后浅蓝灰色天空，部分较老建筑的外墙能看到空调机位、晾衣架和褪色招牌。远处一座高架桥横跨画面，桥下亮起第一批暖黄色路灯，更远处是密集城市天际线。云层正在从西侧散开，一束偏暖的夕阳从建筑缝隙中照入，使湿润路面、玻璃幕墙和树冠出现细碎亮光。整体色彩自然克制，既有雨后空气的清透感，也保留交通、人群、建筑和街道设施构成的真实城市复杂度。
    \par\smallskip\noindent\textit{English translation.} A photorealistic urban landscape photograph taken on an evening when the sky has just cleared after rain. The scene looks downward from an elevated pedestrian overpass. In the foreground are rain-soaked metal railings and several plane-tree leaves stuck to the ground, with tiny water droplets still clinging to the railings. The middle ground contains a broad urban arterial road whose dark-gray asphalt remains wet. The white lane markings, buses, and private cars cast elongated reflections in standing water, while cyclists and umbrella-carrying pedestrians wait for the traffic light at the intersection. Office towers, residential buildings, and street-level shops of varying heights line both sides of the road. Glass curtain walls reflect the pale blue-gray post-rain sky; on the exteriors of some older buildings, air-conditioner mounting bays, clothes-drying racks, and faded signs are visible. In the distance, an elevated roadway crosses the frame, with the first warm-yellow streetlights glowing beneath it, and a dense city skyline farther beyond. Clouds are breaking apart in the west, and a warm shaft of sunset light enters through gaps between the buildings, creating scattered glints on the wet roadway, glass facades, and tree canopies. The overall colors are natural and restrained, conveying both the crystalline clarity of the air after rain and the authentic urban complexity formed by traffic, crowds, buildings, and street infrastructure.

    \item 一张清晨老城区屋顶视角的写实城市摄影作品。前景是几排灰黑色瓦屋顶和红褐色砖墙，屋檐上能看到雨槽、电视天线和成排空调外机，部分平屋顶上摆放着水箱、晾衣架和盆栽。狭窄街巷隐藏在建筑之间，只能从缝隙中看到早餐摊升起的淡淡蒸汽和几辆缓慢经过的电动车。中景逐渐过渡到八九十年代建设的浅灰和米黄色住宅楼，阳台密集，窗户样式不统一。更远处则是一片明显更现代的城市核心区，高层写字楼和住宅塔楼从旧城区后方升起，玻璃表面反射清晨浅金色光线。低处仍残留少量薄雾，使不同距离的建筑产生自然空气透视。天空为浅蓝色，几条淡薄云带横向延伸。画面重点不是宏伟天际线，而是通过瓦屋、生活设施、旧住宅和现代高楼同时出现，表现城市长期生长形成的丰富层次。
    \par\smallskip\noindent\textit{English translation.} A photorealistic urban photograph viewed across the rooftops of an old city district in the early morning. The foreground contains several rows of gray-black tiled roofs and reddish-brown brick walls. Gutters, television antennas, and rows of outdoor air-conditioning units can be seen along the eaves, while water tanks, clothes-drying racks, and potted plants sit on some flat roofs. Narrow lanes are hidden between the buildings; only through the gaps can one glimpse faint steam rising from breakfast stalls and several electric scooters passing slowly. The middle ground gradually transitions to pale-gray and beige residential blocks built in the 1980s and 1990s, with densely packed balconies and mismatched window styles. Farther away lies a distinctly more modern urban core, where high-rise office buildings and residential towers rise behind the old neighborhood, their glass surfaces reflecting the pale golden light of morning. A small amount of thin mist lingers at lower elevations, creating natural aerial perspective among buildings at different distances. The sky is pale blue, crossed horizontally by several faint, slender bands of cloud. The visual emphasis is not on a grand skyline, but on the rich layers created by tiled houses, everyday residential fixtures, older apartment buildings, and modern towers appearing together, expressing the city's long-term organic growth.

    \item 一张高清风景摄影作品，展现了中国黄山（Huangshan）冬季雪后的壮丽景色。画面前景被茂密的松树占据，深绿色的针叶上覆盖着厚实、蓬松的白雪，树枝向四周伸展，形成自然的画框。中景是陡峭险峻的灰白色花岗岩山峰，岩石纹理清晰，表面斑驳地覆盖着积雪，几株苍劲的松树顽强地生长在悬崖峭壁之上。在两座山峰之间的峡谷深处，云雾缭绕，白色的浓雾如丝绸般缠绕在山腰，营造出朦胧幽深的意境。在云雾掩映的山腰处，隐约可见一座依山而建的传统中式建筑，拥有深色的屋顶和木质结构，与周围险峻的自然环境融为一体。背景是连绵起伏的山峦，逐渐隐没在灰白色的阴沉天空和漫天的雾气中。整体色调以冷峻的灰白、墨绿和深褐为主，光线柔和且漫射，没有强烈的阴影，呈现出一种静谧、肃穆且充满东方水墨画意境的氛围。
    \par\smallskip\noindent\textit{English translation.} A high-definition landscape photograph presenting the magnificent scenery of Huangshan, China, after a winter snowfall. Dense pine trees occupy the foreground, their dark-green needles covered in thick, fluffy white snow. Their branches extend in all directions to form a natural frame. The middle ground features steep, precipitous gray-white granite peaks with clearly visible rock textures and irregular patches of snow across their surfaces. Several vigorous pine trees grow tenaciously on the cliffs. Deep in the gorge between two peaks, clouds and mist swirl, with dense white fog wrapping around the mountainsides like silk and creating an ethereal sense of secluded depth. Partly concealed by mist on the mountainside, a traditional Chinese building constructed against the slope can be faintly seen. Its dark roof and wooden structure blend into the surrounding rugged natural environment. In the background, undulating mountain ranges gradually disappear into a gloomy gray-white sky and pervasive fog. The overall palette is dominated by austere gray-white, deep ink green, and dark brown. The light is soft and diffuse, with no harsh shadows, producing a tranquil, solemn atmosphere imbued with the poetic character of an East Asian ink landscape painting.

    \item 一张写实风格的风光摄影作品，画面主体是一座从左向右横向贯穿整个画面的桥梁，桥体几乎与画面下边缘平行，形成非常明确的水平视觉结构。桥梁位于画面中部偏下的位置，两端分别连接左右两侧的河岸或山体，不做斜向延伸，也不采用强烈透视消失点，整体以正侧面视角完整展示桥梁的长度和结构。  桥体采用浅灰色混凝土与深色钢结构结合的现代设计，桥面平直修长，下方由数个间距均匀的桥墩支撑，桥墩垂直落入水中，结构比例真实。桥面上有少量汽车、公交车和行人经过，人物和车辆尺寸较小，用来体现桥梁的实际尺度。护栏、路灯、伸缩缝和桥面边缘都保留清楚的工程细节，但不要让车辆遮挡桥体主体。  桥下是一条宽阔而平缓的河流，水面呈灰蓝色，微风形成细密波纹，桥墩和桥体在水面留下断断续续的倒影。前景靠近岸边的位置可以看到浅色石滩、芦苇、低矮灌木和几块被水冲刷得较圆润的岩石，使画面下方不显得空旷。河岸两侧分布着树木、步道和零散建筑，远处则是一层较低的城市天际线或起伏山脉，让桥梁保持绝对视觉主体。  天空占据画面上半部分，采用阴天转晴后的浅蓝灰色调，云层有明显厚薄变化，局部阳光从云隙中照到桥面和水面，在桥体边缘形成柔和亮部。整体色彩自然克制，不使用夸张霞光，不追求奇幻感。画面强调桥梁横向展开的长度感、稳定感和空间秩序，桥体从画面左侧一直延伸到右侧，正侧面完整可见，构图平衡、开阔、真实。
    \par\smallskip\noindent\textit{English translation.} A photorealistic landscape photograph whose principal subject is a bridge running horizontally across the entire frame from left to right. The bridge is almost parallel to the lower edge of the image, creating a very clear horizontal visual structure. It is positioned slightly below the center of the frame, with its two ends connecting to the riverbanks or mountains on the left and right. It must not extend diagonally or use a strong perspective vanishing point; instead, a direct side view should present the bridge's full length and structure. The bridge has a modern design combining light-gray concrete with dark steel construction. Its deck is straight, long, and slender, supported below by several evenly spaced piers descending vertically into the water, with structurally realistic proportions. A small number of cars, buses, and pedestrians cross the bridge; the people and vehicles are small in scale to convey the bridge's true dimensions. Engineering details such as guardrails, streetlights, expansion joints, and the edges of the deck remain clearly visible, but the vehicles must not obscure the main bridge structure. Beneath the bridge flows a broad, gently moving river with gray-blue water. A light breeze produces fine ripples, and the bridge and its piers cast intermittent reflections on the surface. Near the bank in the foreground are a pale stony shore, reeds, low shrubs, and several rocks rounded by water erosion, preventing the lower portion of the image from feeling empty. Trees, walking paths, and scattered buildings line both riverbanks, while a low city skyline or undulating mountain range appears in the distance, allowing the bridge to remain the unequivocal visual focus. The sky occupies the upper half of the image and has a pale blue-gray tone after an overcast sky has begun to clear. The cloud layers vary noticeably in thickness, and patches of sunlight pass through gaps in the clouds onto the bridge deck and water, forming soft highlights along the bridge's edges. The overall colors are natural and restrained, without exaggerated sunset glow or a fantastical appearance. The image emphasizes the bridge's horizontally extended length, stability, and spatial order. The bridge spans continuously from the left edge of the frame to the right, fully visible in direct side profile, in a balanced, expansive, and realistic composition.

    \item 一张拍摄于中国秦岭森林中的野生动物摄影作品，主体是一只中国金丝猴停在粗壮树枝上，身体微微侧向镜头，双手抓住树干，头部转向前方，神态警觉而安静。它拥有标志性的金色至橙黄色毛发，脸部带有淡蓝色调的皮肤，眼睛明亮，鼻部轮廓清晰，面部表情真实自然。毛发非常丰富，脸周、肩部和背部的毛发更长，能看到蓬松且有层次的真实质感。背景是中国北方山林环境，树干、苔藓、树叶和柔和的远景森林层次清楚但不杂乱。早晨薄云天气下的自然散射光照亮金丝猴，让毛发和脸部结构清晰呈现。真实自然生态摄影，高分辨率，长焦镜头拍摄感，细节丰富，真实毛发纹理，画面安静而有野外气息。
    \par\smallskip\noindent\textit{English translation.} A wildlife photograph taken in the forests of China's Qinling Mountains. The subject is a golden snub-nosed monkey perched on a thick tree branch, its body turned slightly toward the camera, both hands gripping the tree trunk, and its head turned forward with an alert yet calm expression. It has the species' signature golden to orange-yellow fur, pale blue-toned facial skin, bright eyes, a clearly defined nose, and a realistic, natural facial expression. Its coat is exceptionally abundant, with longer hair around the face, shoulders, and back, revealing a fluffy, layered, lifelike texture. The background depicts a northern Chinese mountain-forest environment, where tree trunks, moss, leaves, and softly rendered distant forest layers are clearly distinguished without appearing cluttered. Natural diffuse light under thin morning clouds illuminates the monkey, clearly revealing its fur and facial structure. Authentic natural-history photography, high resolution, the visual feel of a telephoto-lens capture, rich detail, realistic fur texture, and a quiet image imbued with the atmosphere of the wild.

    \item 一张拍摄于中国四川山地竹林中的野生动物纪实摄影作品，主体是一只成年大熊猫，正安静地坐在一片湿润的竹林地面上，前爪抱着一根新鲜竹子低头啃食。它的黑白毛发厚实蓬松，脸部圆润，黑色眼圈边缘自然，鼻子湿润，嘴边毛发因为咀嚼竹叶而微微散开。毛发质感真实，能够看到不同区域毛发的长短和方向变化，白色毛发不是纯白，带有轻微灰调和自然阴影，黑色毛发有真实层次。环境是四川常见的山地竹林，背景有绿色竹叶、潮湿泥土、苔藓和少量岩石，空气略带湿润感。阴天自然光柔和均匀地落在大熊猫身上，没有夸张高光，整体光线安静自然。真实野生动物摄影，高分辨率，自然色彩，纪实风格，皮毛细节清晰，整体像国家公园里真实拍到的一张照片。
    \par\smallskip\noindent\textit{English translation.} A documentary wildlife photograph taken in a mountainous bamboo forest in Sichuan, China. The subject is an adult giant panda sitting quietly on the damp bamboo-forest floor, holding a fresh stalk of bamboo in its forepaws and lowering its head to eat. Its black-and-white fur is thick and fluffy; its face is round, the edges of its black eye patches are natural, its nose is moist, and the hair around its mouth is slightly splayed from chewing bamboo leaves. The fur texture is lifelike, showing variations in hair length and growth direction across different areas. The white fur is not pure white but carries a slight gray cast and natural shading, while the black fur has realistic tonal depth. The setting is a mountain bamboo forest typical of Sichuan, with green bamboo leaves, damp soil, moss, and a small number of rocks in the background, and a subtly humid quality to the air. Soft, even natural light from an overcast sky falls on the giant panda without exaggerated highlights, creating quiet and natural illumination throughout. Authentic wildlife photography, high resolution, natural colors, documentary style, clearly resolved fur details, and the overall appearance of a photograph genuinely captured in a national park.

    \item 一幅垂直构图的中国传统水墨山水画轴，画面描绘秋日山谷中的溪流与松林。近景是数株苍劲古松，树干以浓墨皴擦出粗糙树皮，枝叶层层展开。中景有一条浅溪从山石间穿过，水面以留白表现，岸边点缀几块灰褐色岩石和少量枯黄草木。远景群山层层递进，以淡墨晕染出薄雾笼罩的空间感。画面右上角以行楷竖排题写“空山新雨后，天气晚来秋”，在其左侧有小字写着“丙午年秋谢俊”，文字清晰完整，墨色略干，旁边钤有一方朱红色印章。画作绘制在略微泛黄的纸本上，边缘能够看到细小折痕和自然氧化痕迹。四周以米黄色绫边装裱，整体色彩以墨黑、灰褐、浅赭和少量青绿色为主，光线柔和均匀，呈现真实古画展陈的沉静质感。
    \par\smallskip\noindent\textit{English translation.} A vertically composed hanging scroll in the style of traditional Chinese ink landscape painting, depicting a stream and pine forest in an autumn valley. In the foreground stand several vigorous ancient pines, their rough bark rendered with dense-ink texture strokes and their branches and needles unfolding in successive layers. In the middle ground, a shallow stream passes between mountain rocks; the water surface is represented by reserved blank paper, and several gray-brown rocks and a small amount of withered yellow vegetation dot the banks. In the distance, mountain ranges recede layer by layer, with pale-ink washes creating a sense of space enveloped in thin mist. In the upper-right corner, the inscription “空山新雨后，天气晚来秋” is written vertically in running-standard script. To its left, smaller characters read “丙午年秋谢俊”. All characters are clear and complete, the ink is slightly dry, and a vermilion seal is stamped beside them. The painting is executed on slightly yellowed paper, with tiny creases and traces of natural oxidation visible along the edges. It is mounted on all sides with beige silk borders. The overall palette consists mainly of ink black, gray-brown, pale ocher, and a small amount of blue-green. Soft, even lighting conveys the tranquil material presence of an authentic antique painting on display.

    \item 一幅书法作品特写，采用平铺视角拍摄，画面完全被米黄色的、带有金色丝线的宣纸背景填满，没有任何边框或留白。纸张表面呈现出自然的纤维纹理和细微的颗粒感，色调温暖且均匀。画面主体为两列竖向排列的黑色行草书法汉字，墨色浓郁，笔触苍劲有力，墨迹在纸面上有自然的晕染和飞白效果，显示出毛笔书写的动态感。右侧一列从上至下书写着“功不唐捐”四个大字，左侧一列从上至下书写着“玉汝于成”四个大字。最左侧中下方是落款，从上到下用小字写着“政淋题于丙午年秋”，落款正下方红色小猫头像印章清晰可见。字体结构舒展，笔画间有牵丝连带，墨色浓淡变化丰富，既有饱满的墨块，也有干枯的笔锋。整体风格古朴典雅，光线柔和均匀，无明显阴影，突出了书法艺术的质感和纸张的肌理。
    \par\smallskip\noindent\textit{English translation.} A close-up photograph of a work of calligraphy, shot from a flat, directly overhead viewpoint. The frame is completely filled by a beige xuan-paper background threaded with gold-colored silk fibers, with no border or blank margin. The paper surface displays natural fibrous texture and a subtle grain, in a warm, even tone. The main subject consists of two vertical columns of black Chinese characters in running-cursive calligraphy. The ink is rich, the brushwork vigorous and forceful, and the ink naturally feathers across the paper with areas of dry-brush white streaking, conveying the dynamic motion of brush writing. The right-hand column reads, from top to bottom, the four large characters “功不唐捐”; the left-hand column reads, from top to bottom, the four large characters “玉汝于成”. At the middle-to-lower left is the signature, written vertically in smaller characters as “政淋题于丙午年秋”. Directly beneath the signature, a red seal depicting a small cat's head is clearly visible. The character structures are open and expansive, with fine linking strokes connecting the brushwork. The ink varies richly in density, including both full, saturated masses and dry, depleted brush tips. The overall style is classically simple and elegant. Soft, even light with no obvious shadows emphasizes the material quality of the calligraphy and the texture of the paper.

    \item 一张垂直构图的高清摄影作品，主体为米开朗基罗（Michelangelo）著名雕塑《大卫》（David）的白色大理石半身像。雕塑置于画面中央，安放在一个覆盖着深红色织物的圆形台座上，背景是纯色的深红色墙面，墙面上方有横向的装饰线条。大卫的头部微微向右侧转，目光凝视右前方，神情专注而严肃，卷曲的头发雕刻精细，颈部肌肉线条清晰可见，左肩部分残缺。在雕塑的头顶上，巧妙地插着几朵鲜花：左侧有两朵盛开的白色玫瑰，花瓣层层叠叠，右侧有三朵白色雏菊，花蕊呈黄色，绿色的花茎从发丝间伸出，与白色的大理石形成鲜明对比。光线柔和且均匀，从正面照射，在雕塑的右侧和下方投下淡淡的阴影，突显了大理石的质感和花朵的娇嫩。整体色调以红色和白色为主，营造出一种古典艺术与自然生机相结合的氛围。
    \par\smallskip\noindent\textit{English translation.} A vertically composed, high-definition photograph whose subject is a white marble bust of Michelangelo's renowned sculpture \textit{David}. The sculpture is centered in the frame on a round pedestal covered with deep-red fabric. The background is a solid deep-red wall with horizontal decorative molding near its top. David's head turns slightly to the right, his gaze directed toward the front right with a focused, solemn expression. His curly hair is finely carved, the musculature of his neck is clearly visible, and part of his left shoulder is missing. Several fresh flowers are cleverly tucked into the top of the sculpture's head: two fully bloomed white roses with layer upon layer of petals on the left, and three white daisies with yellow centers on the right. Their green stems rise from between the locks of hair, contrasting sharply with the white marble. The lighting is soft and even, falling from the front and casting faint shadows to the sculpture's right and below, emphasizing both the marble's texture and the flowers' delicacy. The overall palette is dominated by red and white, creating an atmosphere that combines classical art with the vitality of nature.

    \item 清康熙1662-1722年松花石龙凤{\CJKfamily{bsmi}硯}，清康熙松花石龙凤砚，包含绿色石砚与紫色石砚盒。左侧为椭圆形绿色石砚，砚面打磨光滑，中央留有一粗椭圆形墨堂，表面可见细微墨痕，砚面通体分布深浅不一的绿色天然横纹。砚池呈新月形，砚面周边浮雕仿古凤纹，凤首位于池左上方，双翼向左右覆盖环绕砚周，周棱因凤羽纹呈连弧状，池右下方浮雕凤爪。右侧为紫色椭圆形砚盒，盒身与盒盖借砚身扣合，盒体由紫绿两层色石琢制，盒身为紫色，盒盖表面为暗紫色层中夹灰绿色层。盒盖周边镂空雕刻仿古子母夔龙纹，大龙在右，幼龙居左，大龙之首雕于上边正中，幼龙之首雕于右下角，二者上下遥望，大龙龙身下垂，小龙龙身上翘，环绕盒盖周边。向内剔去暗紫色层露出灰绿色层，雕出仿古变形带状夔龙纹，呈椭圆形镂空绕接于暗紫色子母夔龙纹，龙身阴刻方回纹、勾云纹、龙鳞等纹饰。纹饰下嵌有玻璃，可透视砚面。盒背周缘起一圈宽棱，表面有一层褐黄色涂料。整体置于浅灰背景上，光线均匀。
    \par\smallskip\noindent\textit{English translation.} A Songhua-stone inkstone with dragon-and-phoenix decoration from the Kangxi reign of the Qing dynasty, 1662--1722; a Qing Kangxi Songhua-stone dragon-and-phoenix inkstone comprising a green stone inkstone and a purple stone inkstone case. On the left is an oval green stone inkstone with a smoothly polished surface. A broad oval grinding surface is reserved at the center, with faint traces of ink visible upon it, and natural horizontal green striations of varying lightness and darkness extend throughout the stone. The ink reservoir is crescent-shaped. Archaistic phoenix motifs are carved in relief around the edge of the inkstone: the phoenix's head is positioned at the upper left of the reservoir, its two wings spread left and right to encircle the inkstone's perimeter, the rim forming a sequence of connected arcs in response to the feather pattern, and a phoenix claw is carved in relief at the lower right of the reservoir. On the right is an oval purple inkstone case whose base and lid clasp around the body of the inkstone. The case is carved from stone with two layers of color, purple and green: the base is purple, while the lid's surface consists of a dark-purple layer interspersed with a gray-green layer. Around the lid, openwork carving forms an archaistic paired parent-and-young \textit{kui}-dragon design, with the large dragon on the right and the young dragon on the left. The large dragon's head is carved at the center of the upper edge, and the young dragon's head at the lower-right corner; they look toward each other from above and below. The large dragon's body curves downward while the smaller dragon's body turns upward, encircling the perimeter of the lid. Farther inward, the dark-purple layer has been cut away to reveal the gray-green layer, in which an archaistic, transformed ribbon-like \textit{kui}-dragon motif is carved. This oval openwork design winds around and connects with the dark-purple parent-and-young \textit{kui}-dragon motif. The dragons' bodies are incised with square-spiral patterns, hooked-cloud patterns, scales, and other ornament. Glass is inlaid beneath the decoration, allowing the inkstone surface to be seen through it. A broad raised ridge runs around the back edge of the case, whose surface bears a layer of brownish-yellow coating. The entire object is placed against a light-gray background under even lighting.

    \item 明{\CJKfamily{bsmi}錢}{\CJKfamily{bsmi}穀}张复合画水程图（二）册高{\CJKfamily{bsmi}郵}，一幅明代钱谷张复合绘制的《水程图（二）》册页，描绘了高邮的水乡风光。画面采用散点透视，前景为宽阔平静的水面，左下角有一处小岛，岛上生长着几株柳树，岸边停泊着几艘小船。中景展示了蜿蜒的河岸，岸边错落有致地分布着多座民居，屋顶多为黄色或灰色，房屋周围绿树环绕。一座石拱桥横跨在河道之上，连接着两岸。背景是连绵起伏的山丘，山脚下有一座宏伟的城楼，城楼上方可见城墙蜿蜒延伸，远处还有一座高耸的宝塔。画面右侧的山坡上也有房屋点缀。整幅画作以水墨淡彩为主，笔触细腻，色彩淡雅，营造出宁静悠远的意境。画面右侧中部有竖排汉字“高{\CJKfamily{bsmi}郵}”。左下角有一方红色印章。整体构图疏密有致，展现了古代水陆交通的繁忙景象与水乡的秀美风光。
    \par\smallskip\noindent\textit{English translation.} \textit{Water Route Map (II), Gaoyou}, an album leaf jointly painted by Qian Gu and Zhang Fu in the Ming dynasty, depicting the waterside scenery of Gaoyou. The image employs shifting perspective. In the foreground is a broad, tranquil expanse of water, with a small island in the lower-left corner. Several willow trees grow on the island, and several small boats are moored along its shore. The middle ground presents a winding riverbank, where numerous residences with mostly yellow or gray roofs are arranged in a varied yet orderly manner and surrounded by green trees. A stone arch bridge spans the waterway, connecting the two banks. The background consists of undulating hills. At the foot of the hills stands a magnificent gate tower, above which a city wall can be seen winding into the distance; farther away rises a tall pagoda. Houses also dot the hillside on the right side of the image. The work is executed primarily in ink and light colors, with delicate brushwork and an understated palette that creates a tranquil, far-reaching mood. At the middle right of the image, the vertical Chinese characters “高{\CJKfamily{bsmi}郵}” appear. A red seal is positioned in the lower-left corner. The overall composition balances dense and open areas, presenting both the bustle of ancient transportation by water and land and the graceful beauty of the waterside region.

        \item 一张垂直构图的极近距离美食摄影作品，聚焦于一块刚刚切开的厚切炸猪排。画面中央，一双深棕色木筷夹起一块猪排切片，炸衣呈均匀的金黄色，粗糙的面包糠颗粒清晰可见，边缘带有几处略深的焦黄色区域。切面中的猪肉厚实而柔嫩，颜色为自然的浅粉白色，纤维细密，内部保留着明显汁水，在切口边缘形成细小反光。下方是一只深色陶瓷餐盘，盘中整齐排列着剩余的猪排切片，旁边堆放着大量切得极细的浅绿色卷心菜丝，部分菜丝自然卷曲。盘子右侧放着一小勺亮黄色芥末和一滩浓稠的深棕色猪排酱，酱汁边缘带有自然流动痕迹。背景虚化，可见一碗白米饭和一只盛着味噌汤的黑色漆碗。光线从侧上方照射，使炸衣颗粒产生细密阴影，突出酥脆外层与湿润肉质之间的强烈质感对比。
    \par\smallskip\noindent\textit{English translation.} A vertically composed, extreme close-up food photograph focusing on a freshly sliced, thick-cut tonkatsu. At the center of the image, a pair of dark-brown wooden chopsticks lifts one slice of the pork cutlet. The breaded crust is an even golden yellow, with the coarse panko crumbs clearly visible and several slightly darker, toasted golden areas along the edges. The pork exposed by the cut is thick and tender, naturally pale pinkish white, with fine fibers and visibly retained juices that form tiny highlights along the cut edge. Below is a dark ceramic plate on which the remaining pork-cutlet slices are arranged neatly, beside a generous mound of extremely finely shredded pale-green cabbage, some strands curling naturally. On the right side of the plate are a small spoonful of bright-yellow mustard and a pool of thick, dark-brown tonkatsu sauce, whose edges show natural traces of flowing. The blurred background reveals a bowl of white rice and a black lacquer bowl filled with miso soup. Light enters from the upper side, casting fine shadows among the breading particles and emphasizing the strong textural contrast between the crisp exterior and the moist meat.

    \item 一张写实风格的美食摄影作品，采用中近距离、略高于桌面的拍摄视角，主体是一份正在烤制中的韩式烤肉。画面中央是一只嵌入餐桌的不锈钢圆形烤盘，烤盘表面铺着数片厚薄不一的五花肉和牛肉片，肉片边缘已经卷起并呈现金黄与焦褐色过渡，中间仍保留粉棕色与油脂透明感，表面能看到明显的烤纹、油脂渗出后的微亮光泽，以及高温烤制时形成的细小焦边。部分肉片上点缀着蒜片、洋葱块和青椒片，烤盘边缘还能看到几朵正在加热的杏鲍菇和金针菇。烤盘中央微微冒起热气，局部可见油脂滴落后产生的细小泡沫和焦化痕迹，营造出真实的现烤状态。  画面前景左侧摆放着一只白色小碗，里面是翠绿色生菜叶和紫苏叶，叶片新鲜舒展，表面带有轻微水珠。前景右侧是一只浅色蘸料碟，分别装有棕色麻油盐、红色韩式辣酱和切碎的蒜末。桌面上还分布着多只韩式小菜碟，包括橙红色泡菜、浅黄色腌萝卜片、凉拌豆芽、拌菠菜和几片白色洋葱丝，颜色丰富但不杂乱。画面右上角有一双金属筷子正夹起一片刚烤好的五花肉，肉片柔软下垂，边缘微焦，能够清楚看到肥瘦相间的层次。背景略微虚化，可以看到深色木纹桌面、不锈钢排烟管、一只装着米饭的不锈钢饭碗，以及一杯浅黄色啤酒。整体光线明亮而偏暖，重点突出烤肉表面的油脂光泽、热气、焦香质感和韩式烤肉店热闹真实的用餐氛围。
    \par\smallskip\noindent\textit{English translation.} A realistic food photograph taken from a medium-close viewpoint slightly above tabletop height, featuring Korean barbecue as it cooks. At the center is a round stainless-steel grill inset into the dining table. Several slices of pork belly and beef of varying thickness cover its surface; their edges have curled and transition from golden to charred brown, while their centers retain pinkish-brown tones and translucent fat. Distinct grill marks, the faint sheen of rendered fat, and tiny charred rims created by high-temperature grilling are visible on the meat. Some slices are garnished with sliced garlic, chunks of onion, and pieces of green pepper, while several king oyster mushrooms and enoki mushrooms can be seen heating around the edge of the grill. A small amount of steam rises from the center, and in places tiny bubbles and caramelized traces caused by dripping fat are visible, creating the authentic appearance of meat being grilled to order. In the left foreground sits a small white bowl filled with vivid-green lettuce and perilla leaves; the leaves are fresh and unfurled, with fine droplets of water on their surfaces. In the right foreground is a light-colored dipping-sauce dish holding, separately, brown sesame oil with salt, red Korean chili paste, and minced garlic. Several small dishes of Korean banchan are also distributed across the table, including orange-red kimchi, pale-yellow pickled radish slices, seasoned bean sprouts, seasoned spinach, and several shreds of white onion; the colors are rich without appearing cluttered. In the upper-right corner, a pair of metal chopsticks lifts a freshly grilled slice of pork belly. The meat hangs softly, its edge lightly charred, and its alternating layers of fat and lean meat are clearly visible. The slightly blurred background shows a dark wood-grain tabletop, a stainless-steel exhaust duct, a stainless-steel rice bowl filled with rice, and a glass of pale-yellow beer. The overall lighting is bright and warm, emphasizing the oily sheen, rising heat, and char-grilled texture of the meat, as well as the lively, authentic dining atmosphere of a Korean barbecue restaurant.

    \item 一张垂直构图、高清晰度的美食摄影照片，近距离俯拍展示了一桌热气腾腾的潮汕牛肉火锅。画面中心是一个置于嵌入式电磁炉上的圆形不锈钢火锅，锅内清澈的牛骨汤正在剧烈沸腾，表面翻滚着密集的气泡，锅中可见一片柠檬和白色的萝卜片，一把不锈钢漏勺斜靠在锅沿右侧，上方升腾着白色的蒸汽。前景横向排列着三个长方形浅色竹木托盘，左侧托盘上整齐码放着红白相间的肥牛卷，纹理清晰；中间托盘上铺着色泽鲜红、质地紧实的嫩牛肉片，正中央点缀着一株翠绿的香菜；右侧托盘上则摆放着厚切的粉褐色牛舌片，同样装饰有香菜。背景左侧的木质桌面上，叠放着两个带有蓝色边缘的黄色陶瓷小碗，后方立着一个红色的方形纸巾盒，盒身印有白色二维码和文字，旁边还有一个米黄色的圆柱形调料罐。背景右上角模糊可见一盘新鲜的黄芽白菜。整体光线明亮均匀，呈现出餐厅室内暖色调的氛围，焦点集中在前景的肉片纹理和锅内的沸腾细节上，景深自然，色彩鲜艳诱人。
    \par\smallskip\noindent\textit{English translation.} A vertically composed, high-definition food photograph showing, in a close overhead view, a steaming table of Chaoshan beef hot pot. At the center is a round stainless-steel hot pot set on an inset induction cooker. The clear beef-bone broth is boiling vigorously, with dense bubbles rolling across the surface; a slice of lemon and slices of white radish are visible in the pot. A stainless-steel skimmer rests diagonally against the right rim, while white steam rises above it. Three rectangular, light-colored bamboo trays are arranged horizontally across the foreground. On the left tray, red-and-white rolls of fatty beef are stacked neatly, with their marbling clearly visible. The center tray is covered with vividly red, firm-textured slices of tender beef, garnished at the exact center with a sprig of bright-green cilantro. The right tray holds thick-cut, pinkish-brown slices of beef tongue, likewise garnished with cilantro. On the wooden tabletop in the left background, two yellow ceramic bowls with blue rims are stacked together. Behind them stands a red square tissue box printed with a white QR code and text, next to a beige cylindrical condiment jar. A plate of fresh yellow-heart napa cabbage is faintly visible, out of focus, in the upper-right background. The lighting is bright and even, creating the warm-toned atmosphere of a restaurant interior. Focus is concentrated on the texture of the meat slices in the foreground and the boiling details inside the pot, with natural depth of field and vivid, appetizing colors.

    \item 竖版的摄影海报，真实质感的自然摄影风格，一只白色家鹅在深蓝色湖面上高速向镜头游来，主体占据画面左下至中央区域，鹅身体微微倾斜，白色羽毛湿润蓬松、纹理清晰，橙黄色宽大鹅嘴正对镜头并微微张开，神态生动、有点滑稽又充满冲劲；鹅快速划水，在身体周围激起大量透明水花和飞溅水珠，水珠被高速快门凝固在空中，细小水滴清晰锐利，形成强烈的动态感。湖水呈浓郁的宝石蓝、深海蓝与藏青色渐变，水面布满柔和波纹、反射和高光，背景只有纯净开阔的水面，不出现岸边、建筑或其他动物。 采用近距离低机位抓拍视角，镜头几乎贴近水面，与鹅保持平视甚至略微俯视的感觉，鹅头位于画面中央偏左，身体从左下方延伸进入画面，尾部靠近左侧边缘，形成斜向前冲的视觉动势；画面右侧和右上方保留大面积平静的深蓝色负空间，使主体与背景形成明显的不对称构图。前景鹅与水花清晰锐利，远处湖面略微柔化，浅景深但整体环境仍具有真实摄影质感。自然日光照明，偏冷色调，鹅的白色羽毛与橙色鹅嘴在蓝色湖水中形成鲜明的冷暖色彩对比，高光自然，阴影柔和，曝光真实，具有专业野生动物摄影、抓拍摄影、杂志封面摄影的视觉品质。 在画面右上方负空间加入极简主义英文排版，使用粗体白色无衬线字体，大号文字分成三行排列：“Stop explaining yourself.”，文字左对齐，行距紧凑，视觉重量明显；主标题下方加入两行较小的白色英文副标题：“They heard you. They just liked the drama.”，再下方加入一行更小的中文“停止解释自己”。所有文字保持干净、现代、克制，与蓝色水面形成高对比，同时不遮挡鹅与飞溅水花。
    \par\smallskip\noindent\textit{English translation.} A vertical photographic poster in a natural-photography style with realistic texture. A white domestic goose swims rapidly toward the camera across a deep-blue lake, occupying the area from the lower left to the center of the image. Its body tilts slightly; its white feathers are wet, fluffy, and clearly textured. Its broad orange-yellow bill points straight at the camera and is slightly open, giving the goose a vivid expression that is a little comical yet full of momentum. As it paddles quickly, the goose throws up abundant transparent splashes and airborne droplets around its body. A high-speed shutter freezes the water in midair, rendering even the tiny droplets crisp and sharp and producing a powerful sense of motion. The lake transitions through rich jewel blue, deep-ocean blue, and navy, with soft ripples, reflections, and highlights across its surface. The background contains only clean, open water, with no shore, buildings, or other animals. Use a close-range, low-angle candid viewpoint, with the lens almost touching the water and viewing the goose at eye level or from very slightly above. Place the goose's head just left of center; its body enters the image from the lower left, with its tail near the left edge, creating a diagonal, forward-surging visual movement. Preserve a large expanse of calm, deep-blue negative space on the right and upper-right, creating a distinctly asymmetric composition between subject and background. Keep the foreground goose and splashes crisp and sharp while softening the distant lake slightly. Use shallow depth of field, but retain the realistic photographic quality of the overall environment. Illuminate the scene with natural daylight in a cool palette. The goose's white feathers and orange bill should form a vivid warm--cool color contrast against the blue water, with natural highlights, soft shadows, and realistic exposure, achieving the visual quality of professional wildlife photography, candid action photography, and magazine-cover photography. In the negative space in the upper right, add minimalist English typography in bold white sans-serif type. Set the large text over three lines: “Stop explaining yourself.” Align it left, use tight leading, and give it clear visual weight. Beneath the main title, add the smaller white English subtitle over two lines: “They heard you. They just liked the drama.” Below that, add an even smaller line of Chinese text, “停止解释自己”. Keep all text clean, modern, and restrained, with high contrast against the blue water, while ensuring that it does not obscure the goose or the splashing water.

    \item 一张手机拍摄的博物馆展品说明牌照片，展品旁的标准说明牌格式，包含文物名称年代材质尺寸和简要描述，设计简约底色为深灰或黑色配白色文字。 顶部居中的画面顶部文字层以中等大小字清晰可见："青花瓷盘（明代 1403-1424）" 中央的字幕条以小字居中清晰显示："材质：瓷器 产地：景德镇 尺寸：口径45.2cm 足径28.5cm 高8.3cm 馆藏编号：故瓷001256"。 底部注释栏以左对齐小字呈现："故宫文物博物院 陶瓷馆  请勿触摸展品 展厅温度：20$\pm$2$^\circ$C  湿度：45\%-55\% 来源：清宫旧藏  入藏时间：1925年" 展签底部的导览信息区以小号字体标注着："语音讲解编号：TC-0256  时长：3分28秒"
    \par\smallskip\noindent\textit{English translation.} A smartphone photograph of a museum exhibit label, using the standard format of an information plaque placed beside an artifact. It includes the artifact's name, period, material, dimensions, and a brief description. The minimalist design uses white text on a dark-gray or black background. Centered at the top, a medium-sized top text layer is clearly visible and reads: "青花瓷盘（明代 1403-1424）". At the center, a caption strip displays the following clearly in small, centered type: "材质：瓷器 产地：景德镇 尺寸：口径45.2cm 足径28.5cm 高8.3cm 馆藏编号：故瓷001256". In the bottom notes area, the following appears in small, left-aligned type: "故宫文物博物院 陶瓷馆  请勿触摸展品 展厅温度：20$\pm$2$^\circ$C  湿度：45\%-55\% 来源：清宫旧藏  入藏时间：1925年". In the visitor-guide information area at the bottom of the label, small type reads: "语音讲解编号：TC-0256  时长：3分28秒".

    \item 竖版东方美学艺术海报，融合传统中国工笔画、古典宫廷建筑细密插画与当代平面设计语言。整体背景为大面积温润的米白色、象牙白宣纸质感，带有细腻自然的纸张纤维与轻微颗粒纹理，画面克制，大量留白。  画面中央由左下向右上贯穿一条巨大而厚重的黑色墨刷笔触，形成强烈的斜向对角线构图。笔触边缘带有明显的干笔飞白、断裂墨迹、毛笔拖拽感和破碎缺口，局部墨块向外飞散，富有书法般的力量感。大面积米白留白与浓黑笔触形成极强视觉反差。  在黑色笔触内部嵌入一组沿中轴层层展开的故宫建筑群，整体呈现紫禁城的秩序感与皇家气象。包括朱红宫墙、金黄色琉璃瓦、汉白玉台基、宽阔石阶、宫门、殿宇、回廊与庭院，建筑由下向上逐层展开，形成清晰的纵深与节奏。建筑细节精密，具有传统界画与工笔重彩插画风格。  建筑之间点缀整齐而繁茂的古树，树冠色彩以深青蓝、孔雀蓝、青绿与金黄色为主，象征秋日宫苑景观。局部穿插乳白色传统祥云纹样，卷云造型细腻典雅。庭院、台阶与宫门前分布少量古代人物，人物极小，身着传统服饰，在宫殿之间行走、驻足、交谈，为恢弘宫阙注入人文气息。  整体色彩以象牙白、墨黑、朱砂红、琉璃黄、孔雀蓝、青绿为主，层次高级，兼具传统中国审美与现代设计海报的精致感。主体景观严格被限制在黑色笔触内部，边缘局部被飞白切断，强化“从墨中浮现”的视觉效果。构图极端不对称，主体沿斜向聚集，四周保持大面积干净留白。  左上角加入极小而克制的中文排版，使用纤细金色高雅字体，两行排列： “故宫博物院” “紫禁城”  旁边搭配一个极简几何圆形徽记，低调精致，不喧宾夺主。  底部中央加入极小字号、字距疏朗的中文文字： “宫阙秋深” 两侧加入细线分隔，右下角加入四个极简轮廓几何符号。  整体呈现高端文化展览海报、博物馆视觉识别、现代东方平面设计风格，中国风编辑海报，故宫建筑工笔插画，宣纸纹理，飞白墨刷构图，大面积负空间，极精细细节，竖版构图
    \par\smallskip\noindent\textit{English translation.} A vertical art poster with an Eastern aesthetic, blending traditional Chinese meticulous gongbi painting, finely detailed illustration of classical imperial architecture, and contemporary graphic-design language. The overall background consists of a large expanse of warm off-white and ivory-white xuan-paper texture, with delicate natural paper fibers and a subtle grain. Keep the image restrained and leave abundant negative space. A huge, heavy black ink-brush stroke runs through the center from the lower left to the upper right, creating a powerful diagonal composition. Its edges show pronounced dry-brush reserve white, broken ink marks, the dragging texture of the brush, and fractured gaps; localized chunks of ink scatter outward, conveying calligraphic force. The broad off-white negative space and dense black stroke create an extremely strong visual contrast. Embed within the black stroke a group of Forbidden City buildings unfolding layer by layer along the central axis, conveying the order of the Forbidden City and the grandeur of imperial authority. Include vermilion palace walls, golden-yellow glazed roof tiles, white-marble terraces, broad stone stairways, palace gates, halls, corridors, and courtyards. Arrange the architecture in successive tiers rising from bottom to top to create clear depth and rhythm. Render the architectural details with precision, in the style of traditional ruled-line architectural painting and richly colored gongbi illustration. Between the buildings, place orderly, luxuriant ancient trees whose crowns are predominantly deep blue-green, peacock blue, turquoise green, and golden yellow, evoking an imperial garden in autumn. Interweave localized creamy-white traditional auspicious-cloud motifs, with delicately elegant scrolling-cloud forms. Distribute a small number of ancient figures in the courtyards, on the steps, and before palace gates. The figures should be tiny, dressed in traditional clothing, walking, pausing, and conversing among the palaces, adding human presence to the magnificent architecture. Use a principal palette of ivory white, ink black, cinnabar red, glazed-tile yellow, peacock blue, and blue-green. The result should feel sophisticated and layered, combining traditional Chinese aesthetics with the refinement of a modern design poster. Strictly confine the main landscape inside the black brushstroke, with parts of its edges interrupted by dry-brush white gaps, reinforcing the visual effect of the scene “emerging from the ink.” Make the composition extremely asymmetric, with the subject clustered along the diagonal and large areas of clean negative space maintained on all sides. In the upper left, add extremely small, restrained Chinese typography in a slender, elegant gold typeface, arranged in two lines: “故宫博物院” and “紫禁城”. Pair it with a minimalist geometric circular emblem that is understated and refined and does not compete with the subject. At the bottom center, add very small Chinese text with generous tracking: “宫阙秋深”. Place fine dividing lines on both sides, and add four minimalist outlined geometric symbols in the lower right. The overall result should resemble a high-end cultural-exhibition poster, museum visual identity, and modern Eastern graphic design: a Chinese-style editorial poster, detailed gongbi illustration of Forbidden City architecture, xuan-paper texture, dry-brush ink composition, abundant negative space, extremely fine detail, and vertical composition.

    \item 横版城市文旅插画海报，现代极简排版结合东方国风风景插画，整体画面为高端、干净、雅致的旅游城市宣传海报风格。背景为大面积温润的米白色、浅米灰色纯净底色，带有轻微纸张质感，整体留白充足，版式整洁，视觉高级。  画面主体由超大英文单词 “HANGZHOU” 构成，字母占据画面中央大部分区域，采用粗壮、狭长、规整的无衬线大写字体，字母内部不是纯色，而是作为“镂空取景窗”，每个字母中都填充精致的杭州风景插画。字母内部呈现连续但又各自独立的江南景观，包括西湖水面、古典亭台楼阁、石拱桥、宝塔、荷花荷叶、柳枝、园林水榭、曲折步道、湖边树木、远山、茶园梯田，以及现代高楼建筑群。每个字母都像一个微缩景观切片，内部构图精致丰富，形成“城市风景被装进文字里”的视觉效果。  在“HANGZHOU”各字母内部，重点体现杭州代表性元素：  古典宝塔高耸，立于湖山之间； 石拱桥横跨水面，桥上有细小行人； 西湖边有亭台、水榭、小船、曲桥； 湖面上点缀荷叶与淡粉色荷花； 局部出现茶园梯田与采茶人物； 一侧融入现代杭州城市天际线与高楼群，体现传统与现代并存； 柳枝从部分字母顶部自然垂下，增强江南柔美气质； 个别字母中出现圆门、庭院、松树、湖岸步道等中式园林元素。  画面最上方横向排布一条细长的杭州全景插画带，像一条城市长卷。长卷中从左到右依次出现：远山、古塔、拱桥、湖面、亭台、现代城市天际线、湖上列车或轻轨、另一侧古塔与垂柳，形成传统与现代并置的城市轮廓线。天空极简，仅保留淡淡的云朵、几只飞鸟，以及左上方一轮低饱和暖红色圆日，营造安静、诗意、东方审美的氛围。  整体插画风格细腻、平面化、装饰性强，介于国风绘本、旅游宣传插画、文创海报与高级信息图形设计之间。色彩柔和克制，采用低饱和配色，以浅青绿、灰蓝、湖蓝、豆绿色、浅卡其、淡粉、米白、青灰为主，避免高对比和浓烈艳色。水面清透平静，远山朦胧，建筑精巧，植物枝叶柔和细密，整体呈现温润、轻盈、安静、诗意的江南气质。  构图上强调字体海报设计感：  “HANGZHOU”是绝对视觉中心； 顶部的横向城市长卷作为辅助信息层； 背景大面积留白，画面呼吸感强； 整体对称中带有轻微变化，版式平衡而现代； 文字与插画高度融合，兼具平面设计感与叙事性。  画面风格要求：高端城市宣传海报、杭州文旅视觉、国风城市插画、现代排版设计、文字景观融合、极简留白、雅致清新的东方配色、精致细节、平面插画质感、横版海报。
    \par\smallskip\noindent\textit{English translation.} A horizontal illustrated urban-culture-and-tourism poster combining modern minimalist typography with an Eastern, Chinese-style landscape illustration. The overall image should have the high-end, clean, and elegant look of a promotional poster for a tourist city. Use a large expanse of warm off-white and pale beige-gray as a pristine background, with a subtle paper texture, ample negative space, a tidy layout, and a sophisticated visual finish. Construct the main subject from the enormous English word “HANGZHOU,” occupying most of the center of the image in bold, condensed, regular sans-serif capitals. The interiors of the letters should not be solid color; instead, treat them as “cutout viewing windows,” each filled with a refined illustration of Hangzhou scenery. Within the letters, depict Jiangnan landscapes that are continuous yet individually distinct, including the surface of West Lake, classical pavilions and towers, stone arch bridges, pagodas, lotus flowers and leaves, willow branches, waterside garden pavilions, winding paths, lakeside trees, distant mountains, terraced tea plantations, and clusters of modern high-rises. Each letter should resemble a miniature landscape cross-section with an intricate, rich internal composition, creating the visual effect of “city scenery placed inside the text.” Within the letters of “HANGZHOU,” prominently feature representative Hangzhou elements: a tall classical pagoda standing amid lake and mountains; a stone arch bridge spanning the water, with tiny pedestrians crossing it; pavilions, waterside halls, small boats, and curved bridges beside West Lake; lotus leaves and pale-pink lotus blossoms scattered across the lake; terraced tea fields and tea pickers in selected areas; a modern Hangzhou skyline and clusters of high-rises integrated on one side to express the coexistence of tradition and modernity; willow branches hanging naturally from the tops of some letters to enhance the gentle Jiangnan character; and Chinese-garden elements such as moon gates, courtyards, pine trees, and lakeside paths inside individual letters. Across the very top of the image, arrange a slender horizontal band of panoramic Hangzhou illustration like a long city handscroll. From left to right, this scroll should show distant mountains, an ancient pagoda, an arch bridge, the lake, pavilions, a modern urban skyline, a train or light rail over the lake, and another ancient pagoda with weeping willows, forming a city silhouette in which the traditional and the modern are juxtaposed. Keep the sky extremely minimal, retaining only faint clouds, a few flying birds, and a low-saturation warm-red circular sun in the upper left, creating a quiet, poetic atmosphere with an Eastern aesthetic. The overall illustration style should be delicate, flat, and highly decorative, positioned between a Chinese-style picture book, tourism-promotion illustration, cultural-creative poster, and high-end infographic design. Use a soft, restrained, low-saturation palette dominated by pale blue-green, gray-blue, lake blue, muted bean green, light khaki, pale pink, off-white, and blue-gray, avoiding high contrast and intense colors. Make the water clear and calm, the distant mountains hazy, the architecture finely rendered, and the foliage soft and detailed, giving the whole image the warm, light, quiet, poetic character of Jiangnan. Emphasize the feel of typographic poster design in the composition: “HANGZHOU” must be the absolute visual center; the horizontal city handscroll at the top serves as a secondary information layer; a large amount of background negative space gives the image room to breathe; the composition should be broadly symmetric with slight variations, creating a balanced, modern layout; and text and illustration should be deeply integrated, combining graphic-design clarity with narrative richness. Required visual style: high-end city-promotion poster, Hangzhou cultural-tourism identity, Chinese-style urban illustration, modern typographic design, integration of text and landscape, minimalist negative space, elegant and fresh Eastern colors, refined details, flat-illustration texture, and a horizontal poster format.

    \item 竖版现代东方文化海报，温暖象牙白纸张背景。右上角一支真实毛笔正在绘制一道浅青绿色与湖蓝色混合的长弧形水墨笔触。  整条笔触既是毛笔留下的颜料，也是苏州河道与水巷。  沿蓝绿色笔触从左下向右上布置精致微缩苏州景观。  左下前景是一片白墙黛瓦江南民居，建筑紧邻水巷，两岸有石板路和垂柳，一座拱形石桥跨过水面，小木船从桥洞中穿过。  向中央逐渐出现苏州园林，月洞门、太湖石、亭台、水榭、回廊和松树组成精致的园林微缩景观。  中段以虎丘塔作为主要城市文化地标，塔身从青绿色树木中升起。  随后传统园林逐渐过渡至现代苏州城市轮廓，可以克制加入现代高楼和金鸡湖城市意象，但现代建筑不能压过古城气质。  最后所有景观逐渐缩小，并在右上角融入毛笔笔尖。  整体为低饱和湖蓝、竹青、黛绿、暖白、淡灰和少量砖红，气氛细腻、安静、文雅。  指定文字  顶部：  SUZHOU CHINA  副标题：  A City Between Gardens and Water  中部：  Canal Whispers  小字：  Bridges, gardens and quiet lanes unfold beside the water.  底部：  Gardens of Time  下一行：  White walls, dark roofs and ancient canals preserve the rhythm of Jiangnan.  最后：  Poetry in Every Turn  下一行：  Cross a bridge, enter a garden, and discover another view.  只允许出现以上指定文字，全部准确清晰渲染。
    \par\smallskip\noindent\textit{English translation.} A vertical modern Eastern cultural poster on a warm ivory-white paper background. In the upper right, a realistic calligraphy brush is painting a long, arcing ink stroke that blends pale blue-green with lake blue. The entire stroke is both the pigment left by the brush and the waterways and canal lanes of Suzhou. Arrange exquisite miniature Suzhou scenery along the blue-green stroke from the lower left to the upper right. In the lower-left foreground, show a group of Jiangnan residences with white walls and dark-tiled roofs standing immediately beside a canal lane. Include stone-paved paths and weeping willows on both banks, an arched stone bridge spanning the water, and a small wooden boat passing through the bridge opening. Moving toward the center, gradually introduce Suzhou gardens: moon gates, Taihu rocks, pavilions, waterside halls, covered corridors, and pine trees forming an intricate miniature garden landscape. In the middle section, use Tiger Hill Pagoda as the principal urban-cultural landmark, its tower rising from blue-green trees. Then gradually transition from the traditional gardens to the modern Suzhou skyline. Modern high-rises and visual references to Jinji Lake may be added with restraint, but the modern architecture must not overpower the character of the old city. Finally, make all scenery gradually diminish in scale and merge into the brush tip in the upper right. Use an overall low-saturation palette of lake blue, bamboo green, dark green, warm white, pale gray, and small amounts of brick red, creating a delicate, quiet, and refined atmosphere. Use only the following specified text, rendered accurately and clearly. At the top: SUZHOU CHINA. Subtitle: A City Between Gardens and Water. In the middle: Canal Whispers. Small text: Bridges, gardens and quiet lanes unfold beside the water. At the bottom: Gardens of Time. On the next line: White walls, dark roofs and ancient canals preserve the rhythm of Jiangnan. Finally: Poetry in Every Turn. On the next line: Cross a bridge, enter a garden, and discover another view. No text other than the specified text may appear.

    \item 竖版潮流编辑海报，夜晚街头摄影与高饱和实验平面设计结合。  背景是一条深蓝色夜间街道，一台略显老旧的白色自动售货机独自立在街角，机器内部发出冷白与浅青色灯光，周围环境整体被处理为强烈蓝色调。  地面略微湿润，可以看到冷蓝反光。一罐橙黄色汽水刚刚从机器出口滚出来，成为画面唯一明显暖色视觉焦点。  自动售货机后方叠加一个巨大亮黄色圆形几何图形。  顶部使用巨大黄色高对比衬线标题：  "Small Joys"  下方中文：  "小小快乐"  底部中文：  "快乐不必有正当理由"  左下：  "今天也值得" "奖励自己"  右侧：  "随时供应" "快乐不断"  底部细小英文：  "A LITTLE JOY IS STILL JOY"  外围使用深海军蓝宽边框  下方加入白色细线和三个向右白色三角形。  整体使用深蓝、亮黄、橙色强烈撞色，颗粒胶片、粗网点、丝网印刷和潮流独立杂志视觉。  只允许出现以上双引号内的中文和英文。
    \par\smallskip\noindent\textit{English translation.} A vertical trend-focused editorial poster combining nighttime street photography with highly saturated experimental graphic design. The background is a deep-blue street at night, where a slightly old white vending machine stands alone on a street corner. Its interior emits cool-white and pale-cyan light, and the surrounding environment is treated in an intense blue palette. The ground is slightly wet, showing cool-blue reflections. A can of orange-yellow soda has just rolled out of the machine's dispensing slot, becoming the image's only clearly warm-colored visual focal point. Overlay a huge, bright-yellow circular geometric shape behind the vending machine. At the top, use an enormous yellow high-contrast serif title: "Small Joys". Below it, place the Chinese text: "小小快乐". At the bottom, place the Chinese text: "快乐不必有正当理由". In the lower left: "今天也值得" and "奖励自己". On the right: "随时供应" and "快乐不断". At the bottom, in very small English type: "A LITTLE JOY IS STILL JOY". Surround the image with a wide, deep-navy border. At the bottom, add a fine white line and three right-pointing white triangles. Use a forceful color clash of deep blue, bright yellow, and orange, together with grainy film texture, coarse halftone dots, screen printing, and the visual language of a fashionable independent magazine. Only the Chinese and English text inside the quotation marks above may appear.

    \item 横版极简硬科幻电影概念海报，16:9宽幅构图，融合复古科幻小说封面、现代主义电影海报和粗颗粒丝网印刷质感。整体仅使用深黑、温暖奶油白、暗金黄色以及极少量冷蓝色。  整个画面采用明显的左右不对称构图。左侧约55\%的区域完全由超大字体占据，右侧约45\%用于恒星、宇航员与外星文明视觉。  左侧使用巨大、粗壮、狭窄的奶油米白色无衬线字体，三行排列：  "SAVE" "THE" "SUN"  文字几乎顶到画面上下边缘，字号极大、字距紧密，形成如建筑结构般的巨大文字墙。部分字母被右侧恒星圆盘侵入和遮挡，使字体与图像不是简单并排，而是彼此咬合。  右侧放置一颗占据几乎整个画面高度的巨大恒星。恒星并非写实太阳照片，而是一枚由暗金、米白和粗颗粒印刷纹理构成的巨大圆盘，右侧边缘仍然明亮，左侧部分逐渐被黑色侵蚀，形成恒星正在失去能量的视觉隐喻。  巨大恒星与左侧文字局部重叠，例如太阳边缘切入"SUN"最后几个字母。  恒星前方下部放置一名极小的孤独宇航员，人物背对镜头，站在狭窄的飞船观察平台上，仰望巨大恒星。宇航员尺寸非常小，用尺度差异强化宇宙的压迫感。  宇航员旁边加入一个完全不同于人类科技的非对称外星机械结构，由金属几何体、多面体、支架和奇异接口组成，不设计成恐怖怪物，而是让两个文明的科技同时面对同一个问题。  在右上角留出少量黑色负空间，准确排版：  "PROJECT HAIL MARY"  其下：  "ONE STAR. TWO WORLDS. ONE CHANCE."  右下角较小字体：  "THE SUN IS DYING. THE ANSWER IS LIGHT-YEARS AWAY."  最下方：  "SCIENCE IS THE ONLY WAY HOME."  所有文字必须完整、锐利、拼写准确。禁止伪文字、乱码、额外标志和水印。  整体强调巨大字体、宇宙尺度、孤独感、科学合作、复古硬科幻印刷质感。
    \par\smallskip\noindent\textit{English translation.} A horizontal minimalist hard-science-fiction film concept poster in a 16:9 widescreen composition, blending the look of a vintage science-fiction novel cover, a modernist film poster, and coarse-grained screen printing. Use only deep black, warm cream white, dark golden yellow, and a very small amount of cool blue. Give the entire image a clearly asymmetric left--right composition. Approximately 55\% of the left side is occupied completely by oversized typography, while approximately 45\% on the right is reserved for imagery of a star, an astronaut, and an alien civilization. On the left, use enormous, bold, condensed, cream-beige sans-serif letters arranged over three lines: "SAVE", "THE", and "SUN". The text should nearly touch the upper and lower edges of the image, with an extremely large font size and tight tracking, forming an immense typographic wall like an architectural structure. Allow the stellar disk on the right to intrude upon and obscure parts of some letters so that text and image do not simply sit side by side but interlock. On the right, place a gigantic star occupying almost the full height of the image. The star must not look like a realistic photograph of the Sun; instead, make it a huge disk composed of dark gold, cream white, and coarse-grained print textures. Its right edge remains bright, while its left side is gradually eroded by black, creating a visual metaphor for a star losing its energy. Let the giant star partially overlap the text on the left; for example, its edge may cut into the last few letters of "SUN". In the lower area before the star, place a tiny, solitary astronaut with their back to the camera, standing on a narrow spacecraft observation platform and gazing up at the immense star. Keep the astronaut extremely small, using the disparity in scale to intensify the oppressive immensity of space. Beside the astronaut, add an asymmetric alien mechanical structure entirely unlike human technology, composed of metallic geometric solids, polyhedra, supports, and unfamiliar interfaces. Do not design it as a frightening monster; instead, show the technologies of two civilizations confronting the same problem together. Leave a small area of black negative space in the upper right and typeset the following accurately: "PROJECT HAIL MARY". Beneath it: "ONE STAR. TWO WORLDS. ONE CHANCE." In smaller type in the lower right: "THE SUN IS DYING. THE ANSWER IS LIGHT-YEARS AWAY." At the very bottom: "SCIENCE IS THE ONLY WAY HOME." All text must be complete, sharp, and spelled correctly. Prohibit pseudo-text, garbled characters, additional logos, and watermarks. Overall, emphasize monumental typography, cosmic scale, solitude, scientific cooperation, and the print texture of vintage hard science fiction.

    \item 横版荒诞现代编辑海报，16:9 构图，复古办公室摄影、幽默动物摄影与独立杂志拼贴设计结合。  浅灰白色办公室背景，画面右侧偏中位置站着一匹真实的成年马，体型健壮，毛色自然，身体完整地进入画面。马以一种平静而略显荒诞的方式出现在办公室中，四蹄稳稳站在地面上，身体微微侧向镜头，头部自然转向前方，目光看向镜头，神态自信、倔强、理直气壮，带有一种“我今天就是不想工作”的冷静态度。  马保持真实的身体比例、骨骼结构与肌肉质感，站姿自然，不做夸张拟人化动作。它的脖子上挂着一张办公室工牌，工牌通过挂绳套在颈部，尺寸相对偏小，形成轻微幽默感。工牌清晰可见，但不过分抢眼，可以写上简单英文"STAFF" 。  背景极简，仅保留一张办公桌和一台关闭的笔记本电脑，办公室空间略显空旷，进一步强化“马居然是这里的员工”这种荒诞反差。  左侧上方使用巨大黑色高对比衬线字体：  "Not Today"  下面小型中文：  "今天不行"  中部加入橙色斜体英文：  "Maybe Tomorrow?"  左下排布中文：  "不是没能力" "只是今天不想发挥"  下方小号英文：  "REST IS ALSO PART OF THE PLAN"  最后一行：  "先坐一会"  整体主要使用米白、钴蓝、橙黄和黑色，构图为左侧文字、右侧主体，具有复古半色调印刷、杂志扫描颗粒、荒诞高级编辑视觉。  只能出现以上双引号内指定的中文和英文，不得生成任何其他语言、伪文字、乱码、随机字符、水印或品牌标志。
    \par\smallskip\noindent\textit{English translation.} A horizontal absurdist modern editorial poster in a 16:9 composition, combining vintage office photography, humorous animal photography, and independent-magazine collage design. Against a pale gray-white office background, position a real adult horse slightly right of center. It has a sturdy build and natural coat color, and its entire body is visible within the frame. The horse appears in the office in a calm yet mildly absurd manner: all four hooves stand firmly on the floor, its body is angled slightly toward the camera, its head turns naturally forward, and its gaze meets the camera. Its expression is confident, stubborn, and unapologetically self-assured, conveying the calm attitude of “I simply do not feel like working today.” Preserve realistic body proportions, skeletal structure, and muscular texture. Give it a natural standing pose without exaggerated anthropomorphic gestures. Hang an office ID badge from its neck on a lanyard. The badge should be relatively small, creating subtle humor; it must be clearly visible without becoming overly prominent and may carry the simple English word "STAFF". Keep the background minimal, retaining only a desk and a closed laptop. Make the office feel somewhat empty, further reinforcing the absurd contrast that “the horse is actually an employee here.” In the upper left, use enormous black high-contrast serif type: "Not Today". Beneath it, in small Chinese text: "今天不行". In the middle, add orange italic English text: "Maybe Tomorrow?" In the lower left, arrange the Chinese text: "不是没能力" and "只是今天不想发挥". Below that, add small English text: "REST IS ALSO PART OF THE PLAN". On the final line: "先坐一会". Use primarily off-white, cobalt blue, orange-yellow, and black. Compose the image with text on the left and the subject on the right, and give it vintage halftone printing, scanned-magazine grain, and a sophisticated absurdist editorial look. Only the specified Chinese and English text inside the quotation marks above may appear; do not generate any other language, pseudo-text, garbled text, random characters, watermarks, or brand logos.

    \item 竖版中国文学主题艺术海报，围绕李清照一剪梅中的“月满西楼”进行视觉创作。整体融合中国古典楼阁建筑、秋夜意境、旧式木刻印刷、碑拓质感、丝网印刷与现代编辑设计，呈现安静、怀旧、含蓄而庄重的东方文学气质。 背景是一张具有明显纤维质感的浅奶油色旧宣纸，纸面温润、略微泛黄，可以看到自然的纸纤维、灰色细小斑点、旧纸颗粒与轻微磨损痕迹。画面不是写实摄影，而是具有明显的平面印刷感和旧版画质感。 画面上半部分中央偏右悬挂一轮极其巨大的金黄色满月，月亮占据显著视觉面积，呈完整圆形，颜色为温暖而浓郁的金黄色。月面具有轻微斑驳、颜料不均、旧印刷颗粒与自然色块变化，不做逼真的月球坑洞，也不产生真实摄影式照明，而是一枚具有象征意味的巨大暖色圆盘。 画面中央略偏左位置矗立一座高而狭窄的中国传统多层楼阁式古塔，采用低机位仰视角度。古塔明显窄于背后的满月，使金色圆月从塔身两侧大面积露出。 古塔采用深靛青紫色印刷剪影表现，但保留丰富建筑细节：多层飞檐、向上翘起的檐角、斗拱结构、窗格、栏杆、层层收窄的塔身、顶部细长塔刹，以及少量细长结构线。建筑边缘并非完全锐利，而带有木版印刷般的细小缺口、干墨和颗粒。 画面下半部分由一组斜向延伸的传统瓦屋顶构成主要前景。黛瓦屋檐从左下向右侧斜向穿过画面，层层瓦片、屋脊和飞檐结构清晰，但全部被处理为靛青、灰紫与淡紫色的粗糙印刷剪影。 屋檐周围生长大量秋季树木和枝叶。树冠与枝条具有细密自然的植物结构，同时被压缩成类似旧版画的平面色块。部分枝叶从画面左右边缘伸入，与楼阁和满月形成遮挡关系。画面右侧增加一座较小的传统亭阁或楼阁，隐藏在浓密树木之间，用于丰富层次，但不能抢夺中央古塔的视觉中心。 整体构图形成非常明确的视觉关系： 巨大金色满月 $\rightarrow$ 狭长靛青古塔 $\rightarrow$ 斜向古建筑屋顶 $\rightarrow$ 秋季枝叶。 顶部和两侧保留较多奶油色负空间，下半部分则逐渐被淡紫、灰紫与深靛青建筑和植物填满，形成上轻下重的传统海报结构。 主色严格控制为复古低饱和东方配色： 大面积浅奶油宣纸色，接近 \#E7E0B8 下半部分大面积淡薰衣草紫，接近 \#AFAAD0 建筑、枝叶与主要文字使用深靛青紫，接近 \#575991 次级建筑使用柔和灰紫色，接近 \#928FC0 局部使用中等紫罗兰色，接近 \#7473A8 满月使用明显集中的暖金黄色 少量文字可以使用与满月呼应的金黄色作为强调 整体不要艳丽紫色，也不要现代霓虹感，强调旧印刷中的紫、蓝、奶油色与金黄色关系。 印刷与材质 采用粗粝的双色或多色丝网印刷、Risograph 孔版印刷、传统木刻版画和碑拓质感。 纸张上具有： 轻微套色错位、半色调颗粒、颜料堆积、干墨断裂、墨色深浅不均、细小漏印点、灰尘颗粒和自然老化痕迹。 建筑瓦片、树皮、树叶和斗拱仍然保留精细结构，不要因为复古印刷效果而完全变成模糊色块。 大号中文字体尤其重要：字体应该像旧式木刻标题字或碑刻拓印字，字形方正、紧凑、直立，笔画厚重、横平竖直，末端钝而有力，仅有克制的干墨侵蚀。 不要草书，不要行书，不要夸张毛笔飞白，不要长笔锋，不要书法飘带感。 文字排版 所有需要真正出现在海报中的文字，都必须严格按照双引号中的内容渲染。 左上角 放置一个尺寸很大的单独汉字： "月" 使用深靛青色木刻式方正字体，字形紧凑、直立、厚重，表面有少量旧印刷磨损。 上方中央 使用小号宋体或明体横向排版： "雁字回时，月满西楼" 文字纤细、规整，与巨大标题形成明显字号反差。 在右上方奶油色负空间内预留一块宽约画面 38\% 的横向标题区域。 严格使用一整行： "满西楼" 三个字必须横向并列在同一条基线上，绝对不能换行，不能上下堆叠。 字体采用巨大、厚重、方正的木刻印刷字体，深靛青紫色。笔画横平竖直，字形紧凑，带有轻微干墨颗粒与旧版画磨损，但主体结构必须清晰稳定。 不要使用草书或自由毛笔字。 画面中央偏右 使用单列纤细、规整的竖排宋体文字： "轻解罗裳，独上兰舟。" 文字不要过大，与建筑和月亮保持明显的视觉层级。 左下方 使用一行较小的金黄色中文作为局部视觉强调： "云中谁寄锦书来" 与巨大金色满月形成色彩呼应。 整体效果是一张李清照文学主题的新中式复古编辑海报：巨大的金色满月压在画面上方，深靛青古塔从淡紫色屋顶与秋树之间向月亮升起，建筑与枝叶像旧时代木刻印刷在泛黄宣纸上；文字克制而具有碑拓感，在怀旧印刷质地中呈现“雁字回时，月满西楼”的清冷、思念与秋夜意境。
    \par\smallskip\noindent\textit{English translation.} A vertical art poster themed around Chinese literature, visually interpreting “月满西楼” from Li Qingzhao's poem “一剪梅”. Blend classical Chinese pavilion and tower architecture, an autumn-night mood, old-fashioned woodblock printing, the texture of stone-rubbing impressions, screen printing, and modern editorial design to create a quiet, nostalgic, restrained, and dignified Eastern literary sensibility. The background is a sheet of aged, pale-cream xuan paper with a pronounced fibrous texture. Its surface is warm and slightly yellowed, showing natural paper fibers, tiny gray specks, old-paper grain, and slight signs of wear. The image must not be realistic photography; it should have a clear flat-print character and the texture of an antique print. In the upper half, slightly right of center, suspend an extremely large golden-yellow full moon. The moon occupies a prominent visual area, forms a complete circle, and has a warm, saturated golden-yellow color. Its surface shows subtle mottling, uneven pigment, aged-print grain, and natural variations in blocks of color. Do not depict realistic lunar craters or photographic lighting; instead, render it as a huge, symbolic warm-colored disk. Slightly left of center, erect a tall, narrow, traditional Chinese multi-story pavilion-style pagoda, viewed from a low upward-looking angle. The pagoda is visibly narrower than the full moon behind it, allowing broad areas of the golden disk to remain exposed on both sides of the tower. Render the pagoda as a deep indigo-purple printed silhouette while preserving rich architectural detail: multiple tiers of projecting eaves, upturned corners, dougong bracket sets, lattice windows, railings, a body narrowing tier by tier, a slender finial at the top, and a few fine structural lines. Its edges should not be perfectly sharp, but should show tiny gaps, dry ink, and grain like a woodblock print. Form the main foreground in the lower half from a group of traditional tiled roofs extending diagonally. Dark-tiled eaves run diagonally from the lower left toward the right side of the image. The layered tiles, ridgelines, and projecting-eave structures remain clear, but all are treated as rough printed silhouettes in indigo, gray-purple, and pale purple. Surround the eaves with abundant autumn trees, branches, and foliage. The crowns and branches should retain delicate, natural botanical structure while being compressed into flat blocks of color resembling an antique print. Let some branches and leaves enter from the left and right edges and overlap the pavilion and full moon. Add a smaller traditional pavilion or tower on the right, hidden among dense trees, to enrich the depth, but do not let it compete with the central pagoda as the visual focus. Establish a very explicit compositional relationship: enormous golden full moon $\rightarrow$ slender indigo pagoda $\rightarrow$ diagonally arranged ancient tiled roofs $\rightarrow$ autumn branches and leaves. Preserve ample cream negative space at the top and on both sides, while gradually filling the lower half with pale-purple, gray-purple, and deep-indigo architecture and vegetation, producing a traditional poster structure that is light above and visually heavy below. Strictly control the principal palette as a vintage, low-saturation Eastern color scheme: a large expanse of pale-cream xuan-paper color, approximately \#E7E0B8; a broad area of pale lavender in the lower half, approximately \#AFAAD0; deep indigo-purple for the architecture, branches, foliage, and main typography, approximately \#575991; soft gray-purple for secondary architecture, approximately \#928FC0; medium violet in selected areas, approximately \#7473A8; and a distinctly concentrated warm golden yellow for the full moon. A small amount of text may use golden yellow as an accent echoing the moon. Avoid vivid purple and any modern neon feeling; emphasize the relationship among purple, blue, cream, and gold found in aged printing. For printing and material qualities, use rough two-color or multicolor screen printing, Risograph stencil printing, traditional woodblock-print texture, and the texture of stone-rubbing impressions. The paper should show slight color-registration misalignment, halftone grain, accumulated pigment, broken dry ink, uneven ink density, tiny unprinted pinholes, dust particles, and natural aging marks. Preserve fine structures in roof tiles, tree bark, leaves, and dougong; do not allow the vintage-print effect to reduce everything to blurred blocks of color. The large Chinese typography is especially important: it should resemble old woodblock title type or carved inscription characters reproduced by rubbing, with square, compact, upright forms; heavy strokes; level horizontals and vertical uprights; blunt, forceful terminals; and only restrained dry-ink erosion. Do not use cursive script, semi-cursive script, exaggerated flying-white brushwork, long brush tips, or flowing calligraphic ribbons. For typography, every piece of text that actually appears on the poster must be rendered strictly as written inside the quotation marks. In the upper left, place one very large standalone Chinese character: "月". Use a deep-indigo, square woodblock-style typeface with a compact, upright, heavy form and a small amount of aged-print wear on its surface. At the upper center, horizontally typeset in small Songti or Mingti type: "雁字回时，月满西楼". Keep the lettering slender and regular, creating a clear contrast in scale with the enormous title. Reserve a horizontal title area approximately 38\% of the image width within the cream negative space in the upper right. Use strictly one unbroken line: "满西楼". The three characters must be arranged horizontally side by side on the same baseline and must never wrap or stack vertically. Use enormous, heavy, square woodblock-print lettering in deep indigo-purple, with level horizontal and vertical strokes, compact character forms, slight dry-ink grain, and worn antique-print texture, while keeping the primary structures clear and stable. Do not use cursive or freely brushed calligraphy. Slightly right of center, use a single vertical column of slender, orderly Songti text: "轻解罗裳，独上兰舟。" Keep it modest in size and clearly subordinate to the architecture and moon. At the lower left, use one line of smaller golden-yellow Chinese text as a localized visual accent: "云中谁寄锦书来". This should echo the color of the enormous golden moon. The overall result is a new-Chinese-style vintage editorial poster themed around Li Qingzhao's literature: an immense golden full moon presses over the upper image; a deep-indigo pagoda rises toward it from pale-purple roofs and autumn trees; architecture and foliage appear woodblock-printed on yellowed xuan paper; and restrained typography with the texture of a stone rubbing conveys the cool solitude, longing, and autumn-night mood of “雁字回时，月满西楼” through nostalgic print textures.

    \item 偏方形食物科普海报，1:1 构图，复古植物学图鉴、自然史插画与现代食品编辑设计结合。 背景为温暖象牙白色旧纸张，具有轻微纸纤维、泛黄和细颗粒印刷质感。 画面中央放置最大主体：一串成熟的小番茄连着绿色藤蔓，果实大小略有不同，表皮呈自然鲜红和橙红色，高光柔和，使用精致植物学水彩方式描绘，不是摄影，不是卡通。 主体周围不规则分布四幅小型科学插画： 左上是一株带叶、花朵和未成熟绿色果实的小番茄植株； 上方中央是几颗不同成熟程度的小番茄，从绿色、橙黄色逐渐变化到鲜红色； 右侧是一颗从中间切开的番茄横截面，清楚表现果肉、籽腔、种子与汁液结构； 左下是一小组不同形状的小番茄，包括圆形、椭圆形、梨形果实。 右上使用巨大黑色中文标题： "小番茄" 下方黑色衬线英文： "Cherry Tomato" 左侧放置简洁信息： "口味：" "酸甜 · 清爽 · 多汁" 右下放置： "主要营养：" "维生素 C" "番茄红素" "膳食纤维" 底部加入一句很小的文字： "一颗小小的红色果实" 整体颜色以番茄红、橙红、叶片绿、嫩黄和米白为主，复古自然博物学海报，科学食物插画，精细水彩，旧植物图谱，少量文字，大面积留白。
    \par\smallskip\noindent\textit{English translation.} A nearly square educational food poster in a 1:1 composition, combining a vintage botanical field guide, natural-history illustration, and modern food-editorial design. Use warm ivory-white aged paper as the background, with subtle paper fibers, yellowing, and fine-grained print texture. Place the largest subject at the center: a cluster of ripe cherry tomatoes attached to a green vine. The fruits vary slightly in size, their skins naturally bright red and orange-red with soft highlights. Render them as refined botanical watercolor, not as photography and not as cartoons. Distribute four small scientific illustrations irregularly around the main subject. In the upper left, show a cherry-tomato plant with leaves, flowers, and unripe green fruit. At the upper center, show several cherry tomatoes at different stages of ripeness, transitioning from green through orange-yellow to vivid red. On the right, show a cross-section of a tomato cut through the middle, clearly depicting the flesh, seed cavities, seeds, and juice structure. In the lower left, show a small group of cherry tomatoes of different shapes, including round, oval, and pear-shaped fruit. In the upper right, use an enormous black Chinese title: "小番茄". Beneath it, set the black serif English text: "Cherry Tomato". On the left, place concise information: "口味：" and "酸甜 · 清爽 · 多汁". In the lower right, place: "主要营养：", "维生素 C", "番茄红素", and "膳食纤维". At the bottom, add one very small line of text: "一颗小小的红色果实". Use primarily tomato red, orange-red, leaf green, tender yellow, and off-white. The result should resemble a vintage natural-history poster, scientific food illustration, fine watercolor, and an old botanical plate, with little text and abundant negative space.

  \end{enumerate}
\end{CJK*}
\endgroup

\end{document}